\documentclass{article}
\usepackage[utf8]{inputenc}
\usepackage{graphicx}
\usepackage{caption}
\usepackage{subcaption}
\usepackage{wrapfig}
\usepackage{tikz}
\usepackage{microtype}
\usepackage{etoolbox}
\usepackage{hyperref}

\usepackage{amsmath, amssymb}
\usepackage{bm}
\usepackage{mathtools}
\usepackage{chemformula}
\usepackage{siunitx}
\usepackage{enumitem}
\usepackage{breqn}
\usepackage{algorithm}
\usepackage{algpseudocode}

\usepackage{times}
\usepackage{setspace}
\usepackage[margin=1.25in]{geometry}
\usetikzlibrary{positioning}
\usetikzlibrary{fit}

\newcommand\numberthis{\addtocounter{equation}{1}\tag{\theequation}}

\usepackage{xcolor}

\DeclareMathOperator*{\Extr}{Extr}
\DeclareMathOperator*{\argmax}{argmax}

\DeclareMathOperator{\Prob}{P}

\title{Saddle Hierarchy in Dense Associative Memory}
\author{Robin Thériault$^1$\thanks{Corresponding author. E-mail address: \href{mailto:robin.theriault@sns.it}{robin.theriault@sns.it}} \ and Daniele Tantari$^2$ \\
\small{$^1$Scuola Normale Superiore di Pisa, Piazza dei Cavalieri 7, 56126, Pisa (PI), Italy} \\
\small{$^2$Department of Mathematics, University of Bologna}, \\
\small{Piazza di Porta San Donato 5, 40126, Bologna (BO), Italy}}
\date{November 2025}

\begin{document}

\maketitle

\begin{abstract}
    Dense associative memory (DAM) models have been attracting renewed attention since they were shown to be robust to adversarial examples and closely related to cutting edge machine learning paradigms, such as the attention mechanism and generative diffusion. We study a DAM built upon a three-layer Boltzmann machine with Potts hidden units, which represent data clusters and classes. Through a statistical mechanics analysis, we derive saddle-point equations that characterize both the stationary points of DAMs trained on real data and the fixed points of DAMs trained on synthetic data within a teacher-student framework. Based on these results, we propose a novel regularization scheme that makes training significantly more stable. Moreover, we show empirically that our DAM learns interpretable solutions to both supervised and unsupervised classification problems. Pushing our theoretical analysis further, we find that the weights learned by relatively small DAMs correspond to unstable saddle points in larger DAMs. We implement a network-growing algorithm that leverages this saddle-point hierarchy to drastically reduce the computational cost of training dense associative memory.
\end{abstract}

\section{Introduction}
\label{sec:intro}
Studying the stationary points of machine learning algorithms is crucial to understand how they work. For example, \cite{auffinger2013random, choromanska2015loss} demonstrated that local minima in the loss landscape of large artificial Neural Networks (NNs) are relatively close to the global minimum, explaining why they generalize well in practice. Moreover, \cite{razvan2014saddle, dauphin2014identifying} showed that saddle points are much more numerous than local minima in large NNs. These breakthroughs were made by establishing deep connections between the loss landscape of NNs and the energy landscape of disordered systems studied in statistical mechanics. Beyond the broad insights provided by these studies, and despite the progress made by \cite{FUKUMIZU2000local, fukuzumi2019semi-flat, simsek2021geometry, zhang2021embedding}, the classification of stationary points in machine learning algorithms remains an open problem. 

Energy-based models have been playing a central role in studies of NNs and their theoretical properties.
The Hopfield network, one of the most historically important energy-based models, was originally introduced
as a paradigmatic model of biological associative memory \cite{hopfield1982neural}.
Generalized Hopfield networks \cite{chen1986high, psaltis1986nonlinear} were then developed to improve upon the limited storage capacity of the original \cite{hopfield1982neural, amit1985storing}. The scale of the improvement was determined rigorously in following studies \cite{baldi1987number, gardner1987multiconnected, abbott1987storage}. A few years ago, these generalized networks, commonly referred to as Dense Associative Memory (DAM) or modern Hopfield networks, were made into trainable machine learning models capable of accurate pattern classification by Krotov and Hopfield (K \& H) \cite{krotov2016dense}. In a nutshell, K \& H's DAM learns prototypes of patterns in a trainable weight matrix. Each prototype casts a vote for a class, and the patterns awaiting classification are assigned based on the votes of the prototypes that most closely resemble them. The resulting classification scheme is considerably more adversarially robust \cite{krotov2018dense, theriault2024dense} and interpretable \cite{krotov2016dense} than that of feedforward NNs with ReLU activation functions. Since their debut as trainable machine learning architectures, deep connections have been made between modern Hopfield networks and other well-known and established machine learning paradigms, such as attention  \cite{ramsauer2020hopfield} and  generative diffusion \cite{hoover2023memory, ambrogioni2024search}. In particular, modern Hopfield networks were used to implement the attention mechanism of transformers \cite{ramsauer2020hopfield}, which has been attracting a lot of interest in fundamental and applied research. See \cite{krotov2023new, krotov2025modern} for a review of recent advances on the topic. Recently, it was observed that the trainable weights of K \& H's DAM are channeled toward minima by a low-dimensional network of valleys in the loss landscape \cite{boukacem2024waddington}. Moreover, the points where valleys branch out from one another were identified as saddles in the simple case where the DAM has two patterns to learn. In general, it is not straightforward to classify the stationary points of machine learning algorithms \cite{zhang2021embedding}. However, the results of \cite{boukacem2024waddington} and the interpretability of DAMs suggest that their stationary points are both fundamental to their learning dynamics and easier to characterize than that of generic NNs. With this goal in mind, we revisit dense associative memory for pattern classification \cite{krotov2016dense} using the framework of Boltzmann Machines (BMs) \cite{ackley1985learning, smolensky1986information, freund1991unsupervised, hinton2002training} and statistical mechanics.

The body of this paper is structured as follows. In Section \ref{sec:model}, we introduce the DAM model that we study and the analytical tools that we use throughout our work. In particular, in Section \ref{sec:DAM_model}, we derive the DAM in question from a BM template, and, in Section \ref{sec:teacher-student}, we explain the setting that we use to analyze DAM stationary points.

Section \ref{sec:theoretical_results} presents our theoretical contributions.  In Section \ref{sec:saddle-point}, we formulate saddle-point equations that characterize DAM stationary points, and, in Section \ref{sec:hierarchy}, we leverage these equations to establish a saddle-point hierarchy principle stating that weights learned by DAMs of a given width are embedded as saddle points in wider DAMs.

Section \ref{sec:empirical_results} explains how the theoretical insights of Section \ref{sec:theoretical_results} can be used to enhance training and explores how DAMs represent data once trained.
In Section \ref{sec:learning_eff_loss}, we introduce a regularization method that facilitates supervised learning. Next, in Section \ref{sec:interpretability}, we show that our DAM, although designed for supervised tasks, can discover interpretable solutions in both supervised and unsupervised classification settings. Finally, in Section \ref{sec:splitting}, we link our findings to the learning dynamics shaped by valleys and saddles studied in \cite{boukacem2024waddington}, and we implement a network-growing algorithm \cite{wu2019splitting} that exploits the saddle-point hierarchy to significantly reduce DAM training costs.

The code and hyperparameter configurations used in our experiments are available in the following public repository \cite{theriault2025saddlesoftware}.

\section{Model}
\label{sec:model}
A Boltzmann Machine is a canonical graphical model of correlations in discrete data \cite{ackley1985learning}. It is customary to partition BMs into a visible layer {$\mathbf{v} = \left\{ v_i \right\}_{i = 1}^N \in \mathbb{R}^N$} and a hidden layer {$\mathbf{h} = \left\{ h_\mu \right\}_{\mu = 1}^P \in \mathbb{R}^P$} such that connections between the two layers are allowed, but connections within them are prohibited \cite{smolensky1986information}. In this case, the visible layer represents concrete features of the data, whose mutual correlations are encoded in connections with the hidden layer. The Restricted Boltzmann Machine (RBM) obtained using this partition is much easier to train than a generic BM \cite{freund1991unsupervised, hinton2002training} and still has considerable generating power \cite{freund1991unsupervised, leroux2008representational}, making it more practical in machine learning applications \cite{Salakhutdinov2007restricted, kiviken2012multiple, srivastava2013modelling}. The visible and hidden units of an RBM follow the Gibbs distribution
\begin{align*}
    \Prob_\beta \left( \mathbf{v}, \mathbf{h} \big| \mathbf{J} \right) &= Z_\beta \left( \mathbf{J} \right)^{-1} \Prob_0 \left( \mathbf{v} \right) \Prob_0 \left( \mathbf{h} \right) \exp \left( -\beta H \left[ \mathbf{v}, \mathbf{h} ; \mathbf{J} \right] \right),
\end{align*}
where $\beta \geq 0$ is known as the inverse temperature, $\Prob_0 \left( \mathbf{v} \right)$ and $\Prob_0 \left( \mathbf{h} \right)$ are priors on $\mathbf{v}$ and $\mathbf{h}$, $H \left[ \mathbf{v}, \mathbf{h} ; \mathbf{J} \right] = -\sum_{i = 1}^N  \sum_{\mu = 1}^P J^{\mu}_i v_i  h_\mu$ is called the energy function or Hamiltonian, $\mathbf{J} = \left\{ J^\mu_i \right\}_{1 \leq i \leq N}^{1 \leq \mu \leq P}$ are trainable weights, and $Z_\beta \left( \mathbf{J} \right)$ is a normalization constant called the partition function. The inverse temperature $\beta$ represents the absolute strength of the RBM connections, or equivalently controls the amount of noise $T = 1/\beta$ in the RBM. In this regard, $\Prob_0 \left( \mathbf{v} \right)$ and $\Prob_0 \left( \mathbf{h} \right)$, which restrict the form of the Gibbs distribution to help the RBM represent the data, are the marginal laws of $\mathbf{v}$ and $\mathbf{h}$ when there are no connections, i.e. $\beta = 0$. Their contribution to the Gibbs distribution can be tuned with $\beta$, which can therefore be interpreted as a regularization parameter.

\subsection{A Dense Associative Memory (DAM) model}
\label{sec:DAM_model}

As mentioned in the Introduction, we will now derive a DAM model for classification from a BM template. We will explain why our model is a DAM at the end of this Section, once we have clearly defined it. We make three basic assumptions on the distribution of data to be classified:
\begin{enumerate}
    \item \label{assum:invariance} the data is scale invariant, i.e. for all positive scalars $c$, data points $\mathbf{x}$ and $c \mathbf{x}$ are equivalent;
    \item \label{assum:clustering} the data can be partitioned in disjoint clusters;
    \item \label{assum:labeling} the clusters can be grouped into mutually exclusive classes. 
\end{enumerate}
In order to exploit these three assumptions, we study a BM partitioned into three layers with different roles: the data layer $\mathbf{x}$, the hidden layer $\mathbf{h}$ and the class layer $\mathbf{q}$, which represent data, cluster membership and class membership, respectively. The corresponding energy is 
\begin{align*}
    \label{eq:RBM_energy}
    - H \left[ \mathbf{x}, \mathbf{q}, \mathbf{h} ; \mathbf{J} \right] =\sum_{i = 1}^N \sum_{\mu = 1}^P  w^\mu_i x_i h_\mu + \sum_{y = 1}^C \sum_{\mu = 1}^P   u^\mu_y q_y h_\mu + \sum_{\mu = 1}^P h_\mu b^\mu, \numberthis
\end{align*}
where $\mathbf{J} = \left\{ \mathbf{w}, \mathbf{u}, \mathbf{b} \right\}$ is the set of the trainable weights $\mathbf{w} = \left\{ w^\mu_i \right\}_{1 \leq i \leq N}^{1 \leq \mu \leq P}$, $\mathbf{u} = \left\{ u^\mu_i \right\}_{1 \leq i \leq C}^{1 \leq \mu \leq P}$ and $\mathbf{b} = \left\{ b^\mu \right\}_{\mu = 1}^P$.
There are no direct interactions between the visible layer and the class layer. In other words, conditional on the cluster layer, the visible layer and the class layer are independent. Therefore, this BM is a Deep Boltzmann machine (DBM) with 3 layers \cite{salakhutdinov2009deep}, which can also be thought of as an RBM whose visible layer $\mathbf{v}$ is further divided into $\mathbf{x}$ and $\mathbf{q}$.  

Since the data is scale invariant (Assumption \ref{assum:invariance}), the scale of individual data points contains no information about the classification task, so we normalize them by their (Euclidean) norm in the data layer. In other terms, we take the data units $x_i$ to be continuous variables with unit norm $\sqrt{\sum_{i = 1}^N \left( x_i \right)^2} = 1$. We assume no further knowledge about $\mathbf{x}$, so we take the prior $\Prob_0 \left( \mathbf{x} \right)$ to be the uniform distribution on the $N-1$ dimensional unit hypersphere $S^{N-1}$, {i.e. is the set of all $\mathbf{x}$ with $\sqrt{\sum_{i = 1}^N \left( x_i \right)^2} = 1$}. Data normalization is a very common practice in machine learning. For example, normalization by the Euclidean norm has been popular in text document clustering even since its introduction in the 1980s \cite{salton1983introduction}. Various types of normalization also occur in the brain and retina \cite{Carandini2012normalization}.

Since the hidden layer and the class layer aim to represent disjoint clusters and classes, respectively (Assumptions \ref{assum:clustering} and \ref{assum:labeling}), we take their respective units to be mutually exclusive binary variables, i.e. $\mathbf{h}\in \{0,1\}^P$ with $\sum_{\mu=1}^P h_\mu\in\{0,1\}$ and $\mathbf{q}\in \{0,1\}^C$ with $\sum_{y=1}^C q_y\in\{0,1\}$. At any given time, at most one unit per layer can take the value $1$, representing the fact that the clusters and classes are disjoint.
In other words, we take each of the two layers to be the vector representation of a single categorical (or Potts \cite{potts1952some,wu1982potts}) variable with $P+1$ and $C+1$ categories, respectively. As such, $\Prob_0 \left( \mathbf{h} \right)$ and $\Prob_0 \left( \mathbf{q} \right)$ simplify to probability mass functions $\Prob_0 \left( \mathbf{h} = \mathbf{e}_{\gamma} \right)$ and $\Prob_0 \left( \mathbf{q} = \mathbf{e}_{y} \right)$, where we introduce $\mathbf{e}_\gamma = \left\{ \delta_{\gamma \mu} \right\}_{\mu = 1}^P$ for $\gamma \in \left\{ 0,\ldots, P \right\}$ and define $\mathbf{e}_y\in\{0,1\}^C$ analogously for $y\in\{0,\ldots , C\}$. In particular $\mathbf{e}_0=\mathbf{0}$ represents a state outside the $P$ clusters or the $C$ classes.

These priors on the hidden layer and the class layer can also be obtained by introducing fixed inhibitory connections within the hidden layer and the class layer, respectively \cite{kappen1993using, KAPPEN1995deterministic}.
Since at most one hidden unit $h_\mu$ can be activated at once, the hidden layer is a very sparse representation of the visible layer. In machine learning, sparsity can improve interpretability \cite{mozer1988skeletonization}, generalization, computational efficiency \cite{hoefler2021sparsity}, and adversarial robustness \cite{guo2018sparse}. The sparsity of the brain suggests that it is also beneficial for biological neural networks \cite{friston2008hierarchical}.

Given these priors $\Prob_0 \left( \mathbf{x} \right)$, $\Prob_0 \left( \mathbf{h} \right)$ and $\Prob_0 \left( \mathbf{q} \right)$, we derive the marginal distribution of the visible layer $\left( \mathbf{x}, \mathbf{q} \right)$ (see Appendix \ref{app:model} for details). We start our derivation by showing that the conditional distribution of the data layer given the hidden layer has the form
\begin{align*}
    \label{eq:vmf_distribution}
    \Prob_{\beta} \left( \mathbf{x} | \mu, \mathbf{J} \right) &:= \Prob_{\beta} \left( \mathbf{x} | \mathbf{h} = \mathbf{e}_\mu, \mathbf{J} \right) \numberthis \\
    &\propto \exp \left( \beta \sum_{i = 1}^N w^\mu_i x_i \right) \quad \forall \ \mathbf{x}\in S^{N-1} \ \text{and} \ \mu \in \left\{ 1, ..., P \right\}.
\end{align*}
In other words, the probability density $\Prob_{\beta} \left( \mathbf{x} | \mu, \mathbf{J} \right)$ corresponding to each cluster $\mu > 0$ is a von Mises-Fisher (vMF) distribution centered on the direction $\mathbf{w}^\mu = \left\{ w^\mu_i \right\}_{i = 1}^N$ (see Appendix \ref{app:vmf_integration}). In order to interpret the $\mathbf{w}^\mu$ as centroids for their respective clusters, we assume that they belong to $S^{N - 1}$ like the data layer. Under this assumption, we find the normalization constant of $\Prob_{\beta} \left( \mathbf{x} | \mu, \mathbf{J} \right)$ to be $\Omega_N \left( \beta \right) = \frac{\left( 2\pi \right)^{N/2} I_{N/2 - 1} \left( \beta \right)}{\beta^{N/2 - 1}}$, where $I_n \left( x \right)$ is the modified Bessel function of the first kind of order $n$ (see Appendix \ref{app:vmf_integration}).

After slightly more work, we find the marginal distribution of the visible layer to be
\begin{align*}
    \label{eq:direct_distribution}
    \Prob_\beta \left( \mathbf{x}, y \big| \mathbf{w}, \mathbf{p} \right) &:= \Prob_\beta \left( \mathbf{x}, \mathbf{q}=\mathbf{e}_y \big| \mathbf{J} \right) \\
    &= \sum_{\mu = 1}^P p^\mu_y \frac{\exp \left( \beta \sum_{i = 1}^N w^\mu_i x_i \right)}{ \Omega_N \left( \beta \right)} + p^0_y \frac{1}{\Omega_N \left( 0 \right)} \quad \forall \ \mathbf{x}\in S^{N-1} \ \text{and} \ y \in \left\{ 0, ..., C \right\},\numberthis
\end{align*}
where $\mathbf{p} = \left\{ p^\gamma_y \right\}_{0 \leq y \leq C}^{0 \leq \gamma \leq P} = \left\{ \Prob_\beta\left(\mathbf{q}=\mathbf{e}_y,\mathbf{h}=\mathbf{e}_\gamma|\mathbf{J}\right) \right\}_{0 \leq y \leq C}^{0 \leq \gamma \leq P}$. $\Omega_N \left( 0 \right) = \frac{2 \pi^{N/2}}{\Gamma \left( N/2 \right)}$ is the surface area of $S^{N-1}$, where $\Gamma \left( x \right)$ is the Gamma function, so $\frac{1}{\Omega_N \left( 0 \right)}$ is the uniform distribution on $S^{N - 1}$. The uniform distribution term of Eq. (\ref{eq:direct_distribution}) encourages the model to ignore very noisy data during training, which may or may not be desirable depending on the application.

The detailed derivation of Eq. (\ref{eq:direct_distribution}) (in Appendix \ref{app:model}) is inspired by the derivation of Gaussian mixtures from RBMs presented in \cite{Decelle2021restricted} based on the framework of \cite{kappen1993using, KAPPEN1995deterministic}. In our case, the marginal distribution $\Prob_\beta \left( \mathbf{x} | \mathbf{w}, \mathbf{p} \right) = \sum_{y = 0}^C \Prob_\beta \left( \mathbf{x}, y | \mathbf{w}, \mathbf{p} \right)$ and conditional distributions $\Prob_\beta \left( \mathbf{x} | y ; \mathbf{w}, \mathbf{p} \right) = \frac{\Prob_\beta \left( \mathbf{x}, y | \mathbf{w}, \mathbf{p} \right)}{\sum_{\gamma = 0}^P \Prob_\beta \left( y, \gamma | \mathbf{w}, \mathbf{p} \right)}$ are convex mixtures between the uniform distribution on $S^{N - 1}$ and the vMF distributions $\Prob_\beta \left( \mathbf{x} | \mu, \mathbf{J} \right)$ (see Eq. \ref{eq:vmf_distribution}). Probabilistic models similar to $\Prob_\beta \left( \mathbf{x} | \mathbf{w}, \mathbf{p} \right)$ are notably used in text document clustering \cite{banerjee2005clustering}. Mixture distributions that have class-dependent weights like $\Prob_\beta \left( \mathbf{x} | y ; \mathbf{w}, \mathbf{p} \right)$ are also used in Gaussian mixture discriminant analysis \cite{hastie1996discriminant}.

The class weights $\mathbf{p}$ depend on the trainable parameters $\mathbf{u}$ and $\mathbf{b}$ of Eq. (\ref{eq:RBM_energy}) (see Appendix \ref{app:model}). Without loss of generality, we choose to directly study (and train) $\mathbf{p}$ instead of $\mathbf{u}$ and $\mathbf{b}$. Recall that $p^\gamma_y$ is a probability distribution, and in particular $p^\gamma_y \geq 0$ (see Eq. \ref{eq:direct_distribution}). We constrain the marginal $\sum_{y = 0}^C p^\gamma_y$, i.e. the fraction of data in each cluster, to a fixed distribution $p_{\mathbf{h}} \left( \gamma \right)$ and the marginal $\sum_{\gamma = 0}^P p^\gamma_y$, i.e. the proportion of data in each class $y$, to another fixed distribution $p_{\mathbf{q}} \left( y \right)$.
In sum, we end up with the constraints $p^\gamma_y \geq 0$, $\sum_{\gamma = 0}^P p^\gamma_y = p_{\mathbf{q}} \left( y \right)$ and $\sum_{y = 0}^C p^\gamma_y = p_{\mathbf{h}} \left( \gamma \right)$. Since each cluster belongs to a single class (Assumption \ref{assum:labeling}), we expect the $p^\gamma_y$ of a trained model to be close to $\sum_{y^\prime = 0}^C p^\gamma_{y^\prime}$ for a given $y$ and close to $0$ otherwise. 

Given a dataset of $P^*$ patterns $\left\{ \mathbf{x}^{* \mu} \right\}_{\mu = 1}^{P^*}$ with soft labels $q^{* \mu}_y$ \cite{Szegedy2016rethinking}, we can train the weights $\mathbf{w}$ and $\mathbf{p}$ by minimizing the negative log-likelihood loss
\begin{equation}
    \label{eq:loss}
    L \left( \mathbf{w}, \mathbf{p} \right) = -\frac{1}{P^*} \sum_{\mu = 1}^{P^*} \sum_{y = 0}^C q^{* \mu}_y \log \Prob_\beta \left( \mathbf{x}^{* \mu}, y | \mathbf{w}, \mathbf{p} \right),
\end{equation}
which is a form of maximum likelihood estimation. We do so using constrained Stochastic Gradient Descent with momentum (simply called SGD in this paper). We explain how the constraints on $\mathbf{w}$ and $\mathbf{p}$ are enforced in Appendix \ref{app:weight_normalization}, and we briefly discuss the initial conditions and the learning rate in Appendix \ref{app:initialization}.
\begin{figure}
    \centering
    \includegraphics[width=0.495\linewidth]{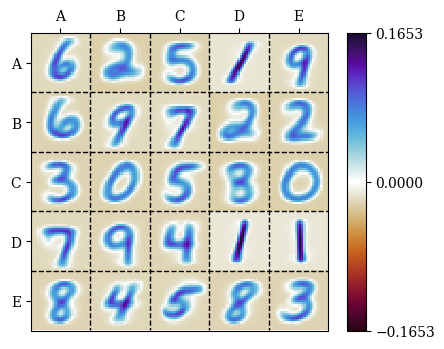}
    \caption{All of the $P = 25$ memories $\left\{ \mathbf{w}^\mu \right\}_{\mu = 1}^{25}$ learned by an instance of our model with $\beta = 16$ when it is trained on the MNIST dataset of handwritten digits \cite{lecun1998gradient} using constrained stochastic gradient descent (SGD) of the negative log-likelihood loss (Eq. \ref{eq:loss}). The hidden units are indexed using pairs of letters from A to E.}
    \label{fig:prelim_DAM_memories}
\end{figure}

The weights $\mathbf{w}^\mu$ learned by SGD of the loss (Eq. \ref{eq:loss}) \cite{lecun1998gradient} are interpretable prototypes, or memories, of the data (see Fig. \ref{fig:prelim_DAM_memories} in the case of the MNIST dataset of handwritten digits \cite{lecun1998gradient}), which is consistent with their role as cluster centroids. Once the model is trained, we can use its conditionals $\Prob_\beta \left( y | \mathbf{x}; \mathbf{w}, \mathbf{p} \right)$ and $\Prob_\beta \left( \mathbf{x} | y; \mathbf{w}, \mathbf{p} \right)$ to efficiently reconstruct $y$ from $\mathbf{x}$ or $\mathbf{x}$ from $y$, respectively. In particular, we can classify unseen patterns using the Bayes classification rule $y = \argmax_{y^\prime} \left\{ \log \Prob_\beta \left( y^\prime | \mathbf{x} ; \mathbf{w}, \mathbf{p} \right) \right\}$ and reconstruct patterns of a given class by finding the local minima of the effective energy $-\log \Prob_\beta \left( \mathbf{x} | y ; \mathbf{w}, \mathbf{p} \right)$ as a function of $\mathbf{x} \in S^{N - 1}$.

High-dimensional probabilistic models that store or learn prototypes, such as our model, can typically reconstruct a limited number of patterns with nontrivial accuracy. In other words, they have limited storage capacity. For instance, Hopfield's model of associative memory, the Hopfield network \cite{hopfield1982neural}, has a capacity of up to $\mathcal{O} \left( N \right)$ patterns \cite{hopfield1982neural, amit1985storing}. Dense Associative Memory (DAM) models, which are inspired by the Hopfield network, are a class of models with an asymptotically much higher capacity \cite{gardner1987multiconnected, krotov2016dense}. The effective energy $-\log \Prob_\beta \left( \mathbf{x} | y ; \mathbf{w}, \mathbf{p} \right)$ that we can minimize to reconstruct patterns with our model (see Eq. \ref{eq:direct_distribution}) is very similar to that of the DAM with exponential capacity introduced in \cite{ramsauer2020hopfield}, which uses energy minimization to implement the attention mechanism of transformers. In fact, according to \cite{lucibello2024exponential}, our model also belongs to the class of DAMs with exponential capacity.

\subsection{Teacher-student setting}
\label{sec:teacher-student}
Among the extensive research on the properties of artificial Neural Networks (NNs) from the perspective of statistical mechanics (see \cite{charbonneau2023spin} for a review), there have been many studies of simple RBMs trained by maximum likelihood estimation (a line of work pioneered in \cite{decelle2017spectral, decelle2018thermodynamics, Decelle2020gaussian}) or by averaging samples from the posterior distribution of the weights $\mathbf{J}$ given some observed data $\mathcal{D}$.
The latter approach has notably been used to characterize the fundamental limits of learning for many types of RBMs \cite{huang2016unsupervised, huang2017statistical, barra2017phase}, with different priors \cite{barra2018phase,huang2018role, manzan2025effect} and architectures \cite{hou2019minimal,theriault2025modeling}, in the \textit{teacher-student setting} \cite{gardner1989unfinished, decelle2021inverse, ALEMANNO2023hopfield} where the data used to train $\mathbf{J}$ is sampled from another RBM with planted weights $\mathbf{J}^*$. This teacher-student setting can also be used to study our DAM. In this scenario, a \textit{teacher} DAM with weights $\mathbf{w}^*$ and $\mathbf{p}^*$ generates a large amount $M = \alpha N$ of noisy data $\mathcal{D} = \left\{ \mathbf{x}^c, y^c \right\}_{c = 1}^M$ and feeds them to a \textit{student} DAM, which then trains its weights $\mathbf{w}$ and $\mathbf{p}$ by averaging samples from the posterior distribution
\begin{align*}
    \label{eq:posterior}
    \Prob_\beta \left( \mathbf{w}, \mathbf{p} | \mathcal{D} \right) &= \mathcal{Z}_\beta \left( \mathcal{D} \right)^{-1} \Prob_0 \left( \mathbf{w}, \mathbf{p} \right) \prod_{c = 1}^M \Prob_\beta \left( \mathbf{x}^c, y^c | \mathbf{w}, \mathbf{p} \right), \numberthis
\end{align*}
where $\mathcal{Z}_\beta \left(\mathcal{D} \right) = \mathbb{E}_{\mathbf{w}, \mathbf{p}} \left[ \prod_{c = 1}^M \Prob_\beta \left( \mathbf{x}^c, y^c | \mathbf{w}, \mathbf{p} \right) \right]$ is the posterior partition function and $\Prob_0 \left( \mathbf{w}, \mathbf{p} \right)$ is the prior on $\mathbf{w}$ and $\mathbf{p}$, which for simplicity we choose as uniform over the sets in which $\mathbf{w}$ and $\mathbf{p}$ are constrained.

We use a statistical mechanics approach to derive a single set of \textit{saddle-point} equations that simultaneously characterize the weights that are stationary points of maximum likelihood estimation (Eq. \ref{eq:loss}) for generic data and the typical weight configurations obtained by averaging samples from Eq. (\ref{eq:posterior}) in the teacher-student setting. We assume that the student does not know the number of hidden units $P^*$ and the inverse temperature $\beta^*$ of the teacher, so it cannot match them with its own. In particular, we consider the case where the noise injected by the teacher in the data is relatively small, i.e. $\beta^* / N > 0$ as $N \rightarrow \infty$, while the student chooses a conservative inverse temperature $\beta \ll N$ to avoid overfitting. Moreover, we fix $\sum_{y = 0}^C p^{* 0}_y = p_{\mathbf{h}}^* \left( 0 \right) = 0$ and $\sum_{y = 0}^C p^{* \mu}_y = p_{\mathbf{h}}^* \left( \mu \right) = 1/P^*$ for all $\mu > 0$ so that each $\mathbf{g}^{* \mu} := \Prob_\beta \left( y | \mu ; \mathbf{w}, \mathbf{p} \right) = \mathbf{p}^{* \mu} / p_{\mathbf{h}}^* \left( \mu \right) = P^* \mathbf{p}^{* \mu}$ is a soft label for the corresponding $\mathbf{w}^{* \mu}$. On the contrary, we do not give $\sum_{\gamma = 0}^{P^*} p^{* \gamma}_y = p_{\mathbf{q}}^* \left( y \right)$ a restrictive form. In other words, $p_{\mathbf{q}}^* \left( y \right)$ is free to be any given probability mass function.

Table (\ref{tab:notation_table}) of Appendix \ref{app:model_summary} summarizes, for quick reference, the most important symbols introduced in this Section and their meaning.

\section{Theoretical results}
\label{sec:theoretical_results}

\subsection{Saddle-point equations}
\label{sec:saddle-point}
In this Section, we introduce a set of equations for the stationary points of the loss (Eq. \ref{eq:loss}), which we then relate to the saddle-point equations emerging from the statistical mechanics analysis of posterior sampling (see Eq. \ref{eq:posterior}) in the teacher-student setting. 

Let us first establish a few definitions that we will use frequently throughout the Section. Given two matrices $\mathbf{w}^*\in\mathbb{R}^{P^*\times N}$ and $\mathbf{w}\in\mathbb{R}^{P\times N}$, we define the overlap matrix $\mathbf{m}(\mathbf{w}^*,\mathbf{w})=\mathbf{w}^*\mathbf{w}^T \in\mathbb{R}^{P^*\times P}$. We write its entries as $m^{\mu_* \mu} \left( \mathbf{w}^*, \mathbf{w} \right) = \sum_{i = 1}^N w^{* \mu_*}_i w^\mu_i$ and its row vectors as $m^{\mu_*}\left( \mathbf{w}^*, \mathbf{w} \right)$, where $1 \leq \mu_* \leq P^*$ and $1 \leq \mu \leq P$. Moreover, for any matrix $\mathbf{m}\in \mathbb{R}^{P^*\times (P+1)}$ with entries $m^{\mu_* \gamma}$ and row vectors $m^{\mu_*}$, we use 
$$\sigma_\gamma(m^{\mu_*})= \frac{\exp \left( m^{\mu_* \gamma} \right)}{\sum_{\nu = 0}^P \exp \left( m^{\mu_* \nu} \right)}$$
to represent the entry number $\gamma \in \left\{ 0, 1, ..., P \right\}$ of the softmax function applied to the row vector $m^{\mu_*}$. In this context, $m^{\mu_*}$ has a zeroth component $m^{\mu_*0}$, and so does its softmax.

In Appendix \ref{app:stationarity}, we show that the stationary points of the negative log-likelihood loss (Eq. \ref{eq:loss}) satisfy
\begin{align*}
    \label{eq:stationarity}
    w^\mu_i &= \frac{\bar{w}^\mu_i}{\sqrt{\sum_{j = 1}^N \left[\bar{w}^\mu_j \right]^2}} \\
    p^\gamma_y &= \frac{\bar{p}^\gamma_y}{\zeta^\gamma_y \left( \mathbf{\bar{p}} ; p_{\mathbf{h}} \right)} \numberthis \\
    \text{with} \quad \bar{w}^{ \mu}_i &= \sum_{\mu_* = 1}^{P^*} x^{* \mu_*}_i \sum_{y = 0}^C q^{* \mu_*}_y \sigma_\mu \left( \beta m^{\mu_*} \left( \mathbf{x}^*, \mathbf{w} \right) + \log \left[ \mathbf{p}_y \right] \right) \\
    \bar{p}^\gamma_y &= \sum_{\mu_* = 1}^{P^*} q^{* \mu_*}_y \sigma_\gamma \left( \beta m^{\mu_*} \left( \mathbf{x}^*, \mathbf{w} \right) + \log \left[ \mathbf{p}_y \right] \right)
\end{align*}
for all $1 \leq \mu \leq P$ and $0 \leq \gamma \leq P$, where $m^{\mu_* 0} \left( \mathbf{x}^*, \mathbf{w} \right) = \frac{1}{\beta} \log \left[ \Omega_N \left( \beta \right) / \Omega_N \left( 0 \right) \right]$ and the normalization constant $\zeta^\gamma_y \left( \mathbf{\bar{p}} ; p_{\mathbf{h}} \right)$ is defined in Appendix \ref{app:weight_normalization}.

Similar equations arise naturally in our statistical mechanics analysis, which amounts to using the replica method \cite{nishimori2001statistical, charbonneau2022replica} to compute the limiting free entropy
\begin{align}
    \label{eq:free_entropy}
    f \left( \varrho, \upsilon, \beta, P^*, P \right) = \lim_{\beta^*, M,N\to\infty} \frac{1}{N} \mathbb{E}_{\mathbf{w}^*, \mathbf{p}^*, \mathcal{D}} \log \left[ \mathcal{Z}_\beta \left( \mathcal{D} \right) \right] ,
\end{align}
where $\upsilon = \beta^* / N$, $\varrho = \frac{M}{P^* N}$ and $\mathbb{E}_{\mathbf{w}^*, \mathbf{p}^*, \mathcal{D}}$ is the joint expectation over the distribution of examples $\mathcal{D} = \left\{ \mathbf{x}^c, y^c \right\}_{c = 1}^M$ generated by the teacher and the priors on the teacher weights. To be more precise, we show that the free entropy can be computed with a variational principle of the form
\begin{align}
  f \left( \varrho, \upsilon, \beta, P^*, P \right) = \operatorname{Extr}_{\mathbf{m},\hat{\mathbf{m}},\mathbf{p}} f(\mathbf{m},\hat{\mathbf{m}},\mathbf{p}),
\end{align}
whose extremizer $\mathbf{m} = \left\{ m^{\mu_* \mu} \right\}_{1 \leq \mu \leq P}^{1 \leq \mu_* \leq P^*}$ can be interpreted as the $N \rightarrow \infty$ limit of the expected value of the teacher-student overlaps $\mathbf{m}(\mathbf{w}^*,\mathbf{w})$. 
We show the derivation of the variational principle in Appendix \ref{app:partition_function} together with an explicit expression for the trial function $f(\mathbf{m},\hat{\mathbf{m}},\mathbf{p})$. If we assume that there are $P^* \ll N$ teacher memories and that the priors on $\mathbf{w}^{* \mu_*}$ and $\mathbf{g}^*$ are uniform like those of the student, we find that the expected teacher-student overlaps must satisfy the saddle-point equations
\begin{align*}
    \label{eq:uniform_saddle-point}
    m^{\mu_* \mu} &= \varsigma \left( 2 \beta_{\text{eff}} \varrho \sqrt{\sum_{\nu_* = 1}^{P^*} \left[ \hat{m}^{\nu_* \mu} \right]^2} \right) \frac{\hat{m}^{\mu_* \mu}}{\sqrt{\sum_{\nu_* = 1}^{P^*} \left[ \hat{m}^{\nu_* \mu} \right]^2}} \\
    p^\gamma_y &= \frac{\bar{p}^\gamma_y}{\zeta^\gamma_y \left( \mathbf{\bar{p}} ; p_{\mathbf{h}} \right)} \numberthis \\
    \text{with} \quad \hat{m}^{\mu_* \mu} &= \sum_{y = 0}^C p_{\mathbf{q}}^* \left( y \right) \sigma_\mu \left( \beta_{\text{eff}} m^{\mu_*} + \log \left[ \mathbf{p}_y \right] \right) \\
    \bar{p}^\gamma_y &= p_{\mathbf{q}}^* \left( y \right) \sum_{\mu_* = 1}^{P^*} \sigma_\gamma \left( \beta_{\text{eff}} m^{\mu_*} + \log \left[ \mathbf{p}_y \right] \right)
\end{align*}
for all $1 \leq \mu \leq P$ and $0 \leq \gamma \leq P$, where $\varsigma \left( x \right) = \frac{x}{\sqrt{x^2 + 1} + 1}$, $\beta_{\text{eff}} = \varsigma \left( 2 \upsilon \right) \beta$ and $m^{\mu_* 0} = \frac{1}{\beta_{\text{eff}}} \log \left[ \Omega_N \left( \beta \right) / \Omega_N \left( 0 \right) \right]$.

If we instead clamp the teacher weights $\mathbf{w}^{* \mu_*}$ and $\mathbf{g}^{* \mu_*}$ to fixed patterns $\mathbf{x}^{* \mu_*}$ and their corresponding soft labels $\mathbf{q}^{* \mu_*}$ \cite{Szegedy2016rethinking}, respectively, to mimic a more general distribution for the data, then Eqs. (\ref{eq:uniform_saddle-point}) become (see Appendix \ref{app:saddle-point})
\begin{align*}
    \label{eq:saddle-point}
    m^{\mu_* \mu} &= \varsigma \left( 2 \beta_{\text{eff}} \varrho \sqrt{\sum_{i = 1}^N \left[ \bar{x}^\mu_i \right]^2} \right) \frac{\sum_{i = 1}^N x^{* \mu_*}_i \bar{x}^\mu_i}{\sqrt{\sum_{i = 1}^N \left[ \bar{x}^\mu_i \right]^2}} \\
    p^\gamma_y &= \frac{\bar{p}^\gamma_y}{\zeta^\gamma_y \left( \mathbf{\bar{p}} ; p_{\mathbf{h}} \right)} \numberthis \\
    \text{with} \quad \bar{x}^\mu_i &= \sum_{\mu_* = 1}^{P^*} x^{* \mu_*}_i \sum_{y = 0}^C q^{* \mu_*}_y \sigma_{\mu} \left( \beta_{\text{eff}} m^{\mu_*} + \log \left[ \mathbf{p}_y \right] \right) \\
    \bar{p}^\gamma_y &= \sum_{\mu_* = 1}^{P^*} q^{* \mu_*}_y \sigma_\gamma \left( \beta_{\text{eff}} m^{\mu_*} + \log \left[ \mathbf{p}_y \right] \right). 
\end{align*}
for all $1 \leq \mu \leq P$ and $0 \leq \gamma \leq P$, where we recall that $m^{\mu_* 0} = \frac{1}{\beta_{\text{eff}}} \log \left[ \Omega_N \left( \beta \right) / \Omega_N \left( 0 \right) \right]$.

In the limit of $\varrho, \upsilon \rightarrow \infty$, which indicates a large number of examples and a low level of teacher noise, Eqs. (\ref{eq:saddle-point}) become equivalent to Eqs. (\ref{eq:stationarity}) if we make the identification $\bar{x}^\mu_i=\bar{w}^\mu_i$, i.e.
\begin{equation}\label{eq:identification}
w^\mu_i = \frac{\bar{x}^\mu_i}{\sqrt{\sum_{j = 1}^N \left[ \bar{x}^\mu_j \right]^2}}.
\end{equation}
Let us explain this limit step by step. For any $\upsilon$ and $\varrho$, the $M$ examples $\{\mathbf{x}^c\}_{c=1}^M$ given to the student are corrupted versions of the teacher patterns $\mathbf{x}^{*}$ with a noise level of $1/\upsilon$. Therefore, in the limit of $\upsilon \rightarrow \infty$ with finite $\varrho$, {each example $\mathbf{x}^c$} is one of the original patterns $\mathbf{x}^*$, as in Eqs. (\ref{eq:stationarity}). However, even in this case, the empirical distribution of the examples deviates from that of $\mathbf{x}^*$, of which it is merely a bootstrap sample. At fixed $P^*$ and $P$, this mismatch disturbs the accurate learning of $\mathbf{x}^*$ when $\varrho = \frac{M}{P^* N}$ is finite, or equivalently the expected number of repetitions $M / P^*$ of each pattern is not sufficiently large compared to $N$ (see Eq. \ref{eq:saddle-point}). This influence progressively weakens as $\varrho$ grows and the $\varsigma$ function approaches $1$, reflecting the convergence of the empirical distribution of the examples to that of the teacher patterns.

In the limit of $\varrho \rightarrow \infty$ with finite $\upsilon$, Eq. (\ref{eq:saddle-point}) is still very similar to Eq. (\ref{eq:stationarity}). In that case, the only difference between them is that Eq. (\ref{eq:saddle-point}) has $\beta_{\text{eff}}$ instead of $\beta$ in the argument of $\sigma$ and in the denominator of the definition of $m^{\mu_* 0}$. Using $\varsigma$ as a shorthand for $\varsigma \left( 2 \upsilon \right)$, we find that the fixed points of Eqs. (\ref{eq:saddle-point}) with $\varrho \rightarrow \infty$ are related, through the identification made in Eq. (\ref{eq:identification}), to the stationary points of the effective loss
\begin{gather*}
    \label{eq:effective_loss}
    \mathcal{L} \left( \mathbf{w}, \mathbf{p} \right) = -\frac{1}{P^*} \sum_{\mu = 1}^{P^*} \sum_{y = 0}^C q^{* \mu}_y \log \mathcal{P}_{\beta, \varsigma} \left( \mathbf{x}^{* \mu}, y | \mathbf{w}, \mathbf{p} \right), \numberthis \\
    \text{where} \quad \mathcal{P}_{\beta, \varsigma} \left( \mathbf{x}, y \big| \mathbf{w}, \mathbf{p} \right) = \sum_{\mu = 1}^P p^\mu_y \frac{\exp \left( \varsigma \beta \sum_{i = 1}^N w^\mu_i x_i \right)}{ \Omega_N \left( \beta \right)} + p^0_y \frac{1}{\Omega_N \left( 0 \right)}.
\end{gather*}
What distinguishes this equation from the standard loss (Eq. \ref{eq:loss}) is that $\mathcal{P}_{\beta, \varsigma} \left( \mathbf{x}, y \big| \mathbf{w}, \mathbf{p} \right)$ has $\beta_{\text{eff}} = \varsigma \beta$ in the argument of the exponential function instead of $\beta$. As a consequence, $\mathcal{P}_{\beta, \varsigma} \left( \mathbf{x}, y \big| \mathbf{w}, \mathbf{p} \right)$ is not a probability distribution unless $\varsigma = 1$, in the limit of $\upsilon \rightarrow \infty$ (see Appendix \ref{app:vmf_integration}). A value of $\varsigma$ less than one in the effective loss (Eq. \ref{eq:effective_loss}) is reminiscent of the presence of noise in the data generation process, so we propose to use it as a regularizer for the weights. We discuss this point in more detail in Section \ref{sec:learning_eff_loss}.

\subsection{Saddle-point hierarchy}
\label{sec:hierarchy}
\begin{figure}
    \centering
    \includegraphics[width=0.9\linewidth]{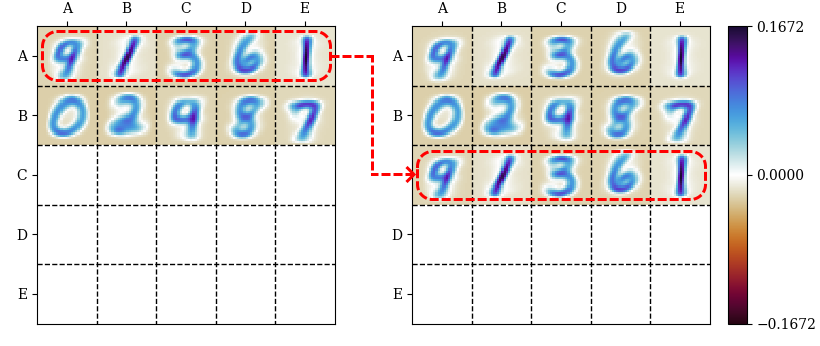}
    
    \hspace{-20pt} $\bar{x}^{\text{fixed}, \mu}_i$ \hspace{146pt} $\bar{x}^{\text{dupli}, \mu}_i$
    
    \caption{Illustration of the relationship  between $\bar{x}^{\text{fixed}, \mu}_i$ and $\bar{x}^{\text{dupli}, \mu}_i$ stated in Eq. (\ref{eq:fixed_point}). The left panel represents the fixed point parameters $\bar{x}^{\text{fixed}, \mu}_i$ of Eq. (\ref{eq:saddle-point}) with $P = 10$, and the right panel represents the fixed point parameters $\bar{x}^{\text{dupli}, \mu}_i$ of Eq. (\ref{eq:saddle-point}) with $P = 15$. In this example, $R = 5$, which means that the first $5$ entries of $\bar{x}^{\text{fixed}, \mu}_i$ are repeated twice in $\bar{x}^{\text{dupli}, \mu}_i$, while the remaining ones are repeated only once. The first $10$ entries of $\bar{x}^{\text{dupli}, \mu}_i$ are identical to $\bar{x}^{\text{fixed}, \mu}_i$, and the dashed red lines highlights that the first $5$ entries of $\bar{x}^{\text{fixed}, \mu}_i$ are repeated a second time at the end of $\bar{x}^{\text{dupli}, \mu}_i$.}
    \label{fig:hierarchy_diagram}
\end{figure}

As shown in \cite{zhang2021embedding}, the loss landscape of any NN with unconstrained weights contains the stationary points of narrower NNs with the same architecture. In Appendix \ref{app:fixed-point}, we show that this result also applies to the teacher-student setting with $\varrho \rightarrow \infty$ and any nonzero $\upsilon$. To be more precise, we show that, if the parameters $\bar{x}^{\text{fixed}, \mu}_i$, $\bar{p}^{\text{fixed}, \gamma}_y$, $m^{\text{fixed}, \mu_* \gamma}$, $p^{\text{fixed}, \gamma}_y$ with hidden unit prior $p_{\mathbf{h}}^{\text{given}} \left( \gamma \right)$ are a fixed point of Eqs. (\ref{eq:saddle-point}) with $P$ hidden units, then the duplicated parameters
{\allowdisplaybreaks
\begin{align*}
    \label{eq:fixed_point}
    \bar{x}^{\text{dupli}, \mu}_i
    &= \begin{cases}
        \bar{x}^{\text{fixed}, \mu}_i &\quad 0 < \mu \leq P \\
        \bar{x}^{\text{fixed}, \mu - P}_i &\quad P < \mu \leq P + R \\
    \end{cases} \\
    \bar{p}^{\text{dupli}, \gamma}_y
    &= \begin{cases}
        \bar{p}^{ \text{fixed}, 0}_y &\quad \gamma = 0 \\
        \frac{1}{2} \bar{p}^{ \text{fixed}, \gamma}_y &\quad 0 < \gamma \leq R \\
        \bar{p}^{ \text{fixed}, \gamma}_y &\quad R < \gamma \leq P \\
        \frac{1}{2} \bar{p}^{ \text{fixed}, \gamma - P}_y &\quad P < \gamma \leq P + R \\
    \end{cases} \\
    m^{\text{dupli}, \mu_* \gamma}
    &= \begin{cases}
        m^{\text{fixed}, \mu_* 0} &\quad \gamma = 0 \\
        m^{\text{fixed}, \mu_* \gamma} &\quad 0 < \gamma \leq P \numberthis \\
        m^{\text{fixed}, \mu_*, \gamma - P} &\quad P < \gamma \leq P + R \\
    \end{cases} \\
    p^{\text{dupli}, \gamma}_y
    &= \begin{cases}
        p^{\text{fixed}, 0}_y &\quad \gamma = 0 \\
        \frac{1}{2} p^{\text{fixed}, \gamma}_y &\quad 0 < \gamma \leq R \\
        p^{\text{fixed}, \gamma}_y &\quad R < \gamma \leq P \\
        \frac{1}{2} p^{\text{fixed}, \gamma - P}_y &\quad P < \gamma \leq P + R \\
    \end{cases}
\end{align*}
\begin{align*}
    \text{along with} \quad p_{\mathbf{h}} \left( \gamma \right)
    &= \begin{cases}
        p_{\mathbf{h}}^{\text{given}} \left( 0 \right) &\quad \gamma = 0 \\
        \frac{1}{2} p_{\mathbf{h}}^{\text{given}} \left( \gamma \right) &\quad 0 < \gamma \leq R \\
        p_{\mathbf{h}}^{\text{given}} \left( \gamma \right) &\quad R < \gamma \leq P \\
        \frac{1}{2} p_{\mathbf{h}}^{\text{given}} \left( \gamma - P \right) &\quad P < \gamma \leq P + R,
    \end{cases}
\end{align*}}%
are a fixed point of the same saddle-point equations with $P + R \in \left\{ P, ..., 2 P \right\}$ hidden units (See Appendix \ref{app:fixed-point} for a detailed derivation {and Fig. \ref{fig:hierarchy_diagram} for a illustrative example}). In other words, we duplicate some of the weights solving Eqs. (\ref{eq:saddle-point}) to construct a fixed point for a wider network. In that sense, wide DAMs contain the fixed points of narrower DAMs. In particular (see Section \ref{sec:saddle-point}), this property also holds for the stationary points of both the standard loss (Eq. \ref{eq:loss}) and the effective loss (Eq. \ref{eq:effective_loss}).

The saddle-point equations (Eq. \ref{eq:saddle-point}) with duplicated order parameters (Eqs. \ref{eq:fixed_point}) are invariant to the permutation of any hidden unit $\gamma \in \left\{ 1, ..., R \right\}$ and its duplicate $\gamma + P$. This kind of symmetry can be spontaneously broken if it leads to a higher free entropy (see Eqs. \ref{eq:free_entropy} and \ref{eq:var_free_entropy}). However, symmetry-breaking transitions can be prohibitively slow when the symmetric state is stable to local perturbations \cite{zdeborova2016statistical}. Interestingly, the DAM introduced in \cite{krotov2016dense} quickly undergoes many successive permutation symmetry-breaking bifurcations during training \cite{boukacem2024waddington}. This observation and additional empirical evidence suggest that the symmetric states are unstable, or in other words that they are saddles, which was verified analytically for DAMs with only two data points to memorize \cite{boukacem2024waddington}. In Appendix \ref{app:fixed-point}, we prove that, if $\beta$ is large enough, Eqs. (\ref{eq:fixed_point}) is a saddle, which is a major step toward explaining why permutation symmetry breaking is relatively fast in DAMs trained with a large number of data points. We call this result \textit{the saddle-point hierarchy principle}.

\section{Empirical results}
\label{sec:empirical_results}
In this Section, we use our theoretical results to improve training, and we show empirically that our DAM learns interpretable solutions to both supervised and unsupervised classification problems. Unless explicitly stated otherwise, we perform our numerical experiments on the MNIST dataset of handwritten digits \cite{lecun1998gradient}. The code and hyperparameter values used in our numerical experiments are available in the following public repository \cite{theriault2025saddlesoftware}.

\subsection{Learning by minimizing the effective loss}
\label{sec:learning_eff_loss}
\begin{figure}
    \centering
    \includegraphics[width=0.495\linewidth]{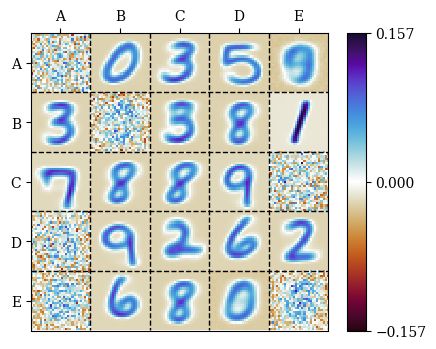}
    \includegraphics[width=0.495\linewidth]{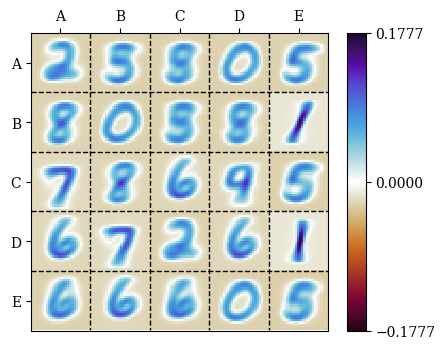}
    \caption{$25$ of the $P = 1000$ memories $\mathbf{w}^\mu$ learned by two instances of our dense associative memory (DAM) model with different values of $\beta$. Both networks are trained on the MNIST dataset of handwritten digits \cite{lecun1998gradient} using constrained stochastic gradient descent (SGD) of the negative log-likelihood loss (Eq. \ref{eq:loss}). The left-panel model has $\beta = 18$, and the right-panel one $\beta = 6$. DAMs with $18 > \beta > 6$ learn memories that interpolate between these two pictures. The hidden units are indexed using pairs of letters from A to E.}
    \label{fig:suboptimal_memories}
\end{figure}

As the number of hidden units $P$ increases, standard maximum likelihood estimation with fixed inverse temperature $\beta$ becomes progressively less apt to train our DAM to its full potential. At high $P$ and $\beta$, many memories stay stuck in noisy states that do not contribute to classification, which is wasteful (see Fig. \ref{fig:suboptimal_memories}). Using a lower $\beta$ helps the memories converge, but also reduces their resolution and diversity, possibly because it also lowers the DAM's capacity \cite{lucibello2024exponential}. This change is far from being only cosmetic. In fact, it comes with a gradual reduction in classification accuracy (see Table \ref{tab:accuracy_table}).

Optimization algorithms with a parameter analogous to $\beta$ often converge to better solutions when this parameter is increased from a small value during optimization. For example, annealing schedules \cite{kirkpatrick1983optimization, rose1990statistical} can be incorporated into minimization algorithms to help them find deeper local minima than they could otherwise. This observation suggests that our DAM would learn better memories if $\beta$ increased during training. It is tempting to do so by SGD of Eq. (\ref{eq:loss}) with respect to $\beta$, but it makes $\beta$ increase so quickly that many memories still stay stuck in noisy states.

We find that optimizing the effective loss (Eq. \ref{eq:effective_loss}) with a suitable $\varsigma$ slows down the evolution of $\beta$ enough to let the DAM learn much cleaner memories (see Fig. \ref{fig:DAM_memories}, top panel) and achieve much better classification accuracy (see Table \ref{tab:accuracy_table}) than with standard training (Eq. \ref{eq:loss}). Looking back at Section \ref{sec:saddle-point}, we propose to interpret $\varsigma$ as a regularization parameter that helps the DAM take noise from the data into account during training. There is no obvious theoretically motivated way to find the best value of $\varsigma$ for a generic dataset, so we choose it by hand. Despite this limitation, we believe that the simplicity and interpretability of our method still make it an interesting alternative to annealing schedules.

When we use this training method, we compute the DAM classification accuracy from the effective predictions
\begin{gather*}
    \label{eq:effective_preds}
    \mathcal{P}_{\beta, \varsigma} \left( y \big| \mathbf{x} ; \mathbf{w}, \mathbf{p} \right) = \frac{\mathcal{P}_{\beta, \varsigma} \left( \mathbf{x}, y \big| \mathbf{w}, \mathbf{p} \right)}{\mathcal{P}_{\beta, \varsigma} \left( \mathbf{x} \big| \mathbf{w}, \mathbf{p} \right)} \numberthis \\
    \text{where} \quad \mathcal{P}_{\beta, \varsigma} \left( \mathbf{x} \big| \mathbf{w}, \mathbf{p} \right) = \sum_{y = 0}^C \mathcal{P}_{\beta, \varsigma} \left( \mathbf{x}, y \big| \mathbf{w}, \mathbf{p} \right),
\end{gather*}
instead of the true predictions $\Prob_{\beta} \left( y \big| \mathbf{x} ; \mathbf{w}, \mathbf{p} \right) = \Prob_{\beta} \left( \mathbf{x}, y \big| \mathbf{w}, \mathbf{p} \right) / \Prob_{\beta} \left( \mathbf{x} \big| \mathbf{w}, \mathbf{p} \right)$.
This approach allows us to calculate both the accuracy and the loss through a single evaluation of $\mathcal{P}_{\beta, \varsigma} \left( y \big| \mathbf{x} ; \mathbf{w}, \mathbf{p} \right)$, which is more efficient than computing $\mathcal{P}_{\beta, \varsigma} \left( \mathbf{x}, y \big| \mathbf{w}, \mathbf{p} \right)$ and $\Prob_{\beta} \left( \mathbf{x}, y \big| \mathbf{w}, \mathbf{p} \right)$ separately when monitoring the progress of training.

\begin{table}
    \centering
    \begin{tabular}{ |c|c| }
        \hline
        Inverse temperature $\beta$ & Classification accuracy \\
        \hline\hline
        6 & 79\% \\
        \hline
        10 & 85\% \\
        \hline
        14 & 89\% \\
        \hline
        18 & 91\% \\
        \hline
        Trained with $\varsigma = 0.25$ & 96\% \\
        \hline
    \end{tabular}
    \caption{DAM classification accuracy (rounded down to two significant figures) for $P = 1000$ and various values of $\beta$. The last line is for $\beta$ trained by SGD of the effective loss (Eq. \ref{eq:effective_loss}) with $\varsigma = 0.25$.}
    \label{tab:accuracy_table}
\end{table}

\subsection{Dense associative memory is interpretable, even in unsupervised classification}
\label{sec:interpretability}
\begin{figure}[!ht]
    \centering
    \includegraphics[width=0.495\linewidth]{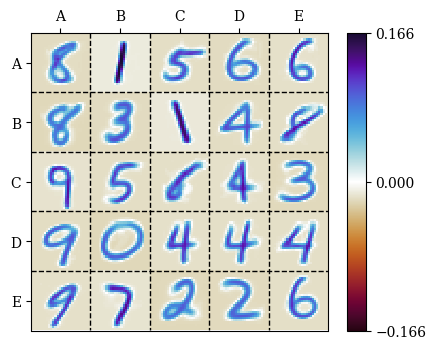}
    \includegraphics[width=\linewidth]{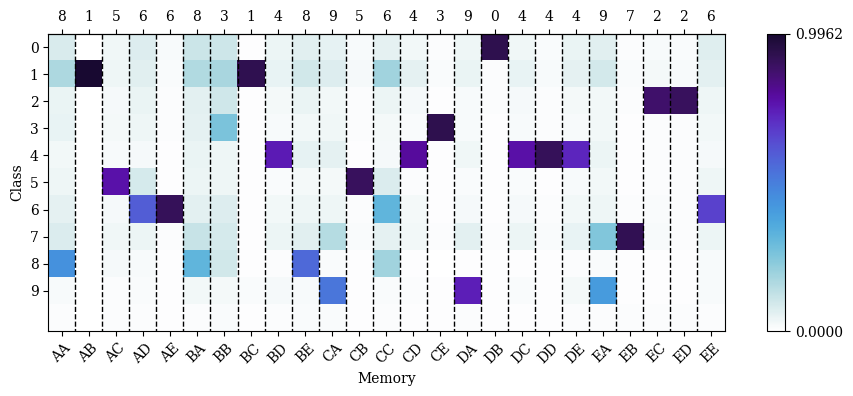}
    \caption{In the top panel, $25$ of the $P = 1000$ memories $\mathbf{w}^\mu$ learned by an instance of our dense associative memory (DAM) model trained on the MNIST dataset of handwritten digits \cite{lecun1998gradient} using constrained stochastic gradient descent (SGD) of the effective loss (Eq. \ref{eq:loss}) with $\varsigma = 0.25$. In the bottom panel, the corresponding rescaled class weights $\mathbf{p}^\mu / p_{\mathbf{h}} \left( \mu \right)$, where $p_{\mathbf{h}} \left( \gamma \right) = \frac{1}{P + 1}$ for all $0 \leq \gamma \leq P$. The hidden units are indexed using pairs of letters from A to E, and the column-wise maxima of the class weights are the classes of the memories with the corresponding letter indices. Rescaled class weights learned with $p_{\mathbf{h}} \left( \gamma \right) \neq \frac{1}{P + 1}$ are qualitatively similar to the ones shown in this figure. Approximately 98\% of test digits fed to the DAM are given the class of the memory that resembles them the most. For example, a digit that looks like the memory \#AA is given the class 8.}
    \label{fig:DAM_memories}
\end{figure}

Now that we understand how to train our DAM reasonably well, we investigate one of the most interesting properties of the solutions that it learns: their interpretability. As advertised in the Introduction, we will explain how to further improve training in the following Section. We already mentioned that the regularization parameter $\varsigma$ of the effective loss is interpretable (see Section \ref{sec:learning_eff_loss} and Eq. \ref{eq:effective_loss}). Here, we point out that the learned weights are as well. In fact, each learned $\mathbf{p}^\mu / p_{\mathbf{h}} \left( \mu \right)$ can be interpreted as a soft label for the corresponding $\mathbf{w}^\mu$ (see Fig. \ref{fig:DAM_memories}). This property was also observed by \cite{boukacem2024waddington} in K \& H's DAM for pattern classification \cite{krotov2016dense}.

At test time, we observe that our DAM classifies approximately 98\% of the test data points into the class $y = \argmax_{y^\prime} \left\{ p^\mu_{y^\prime} \right\}$ of the memory $\mathbf{w}^\mu$ to which they are the most similar. In other words, a 1-nearest neighbor classifier \cite{fix1989discriminatory} conditioned on the memories $\mathbf{w}^\mu$ and their soft labels $\mathbf{p}^\mu / p_{\mathbf{h}} \left( \mu \right)$ approximates the classification of our model with 98\% fidelity. This behavior is reminiscent of K \& H's DAM \cite{krotov2016dense}, where only a few memories participate in the classification of each data point. The 1-nearest neighbor classifier approximates DAM classification more faithfully for correctly classified data points (approximately 99\% fidelity) than incorrectly classified data points (approximately 70\% fidelity).

\begin{figure}
    \centering
    \includegraphics[width=0.495\linewidth]{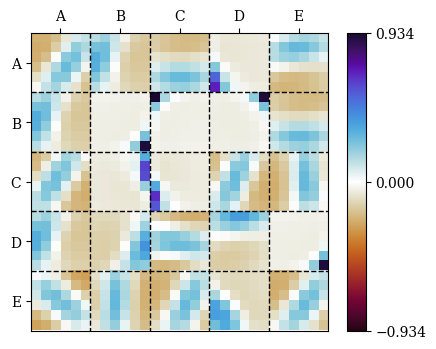}
    \includegraphics[width=\linewidth]{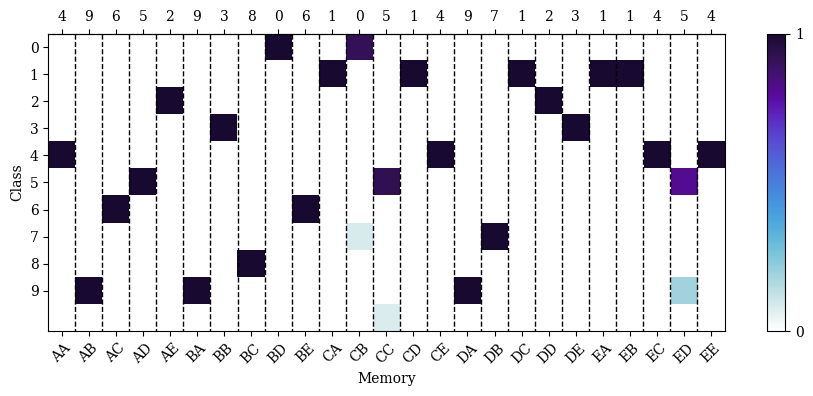}
    \caption{In the top panel, $25$ of the $P = 100$ memories $\mathbf{w}^\mu$ learned by an instance of our dense associative memory (DAM) model trained in an unsupervised way (Eq. \ref{eq:unsupervised_loss}) on $6 \times 6$ patches of the MNIST dataset of handwritten digits \cite{lecun1998gradient} while assuming $C = 10$ latent classes and $\varsigma = 0.6$. In the bottom panel, the corresponding rescaled class weights $\mathbf{p}^\mu / p_{\mathbf{h}} \left( \mu \right)$, where $p_{\mathbf{h}} \left( \gamma \right) = \frac{1}{P + 1}$ for all $0 \leq \gamma \leq P$. The hidden units are indexed using pairs of letters from A to E, and the column-wise maxima of the class weights are the classes of the memories with the corresponding letter indices.}
    \label{fig:unsupervised_DAM_memories}
\end{figure}

We now show that the notion of interpretability described in this Section extends beyond supervised learning. Given a dataset of $P^*$ unlabeled patterns $\{\mathbf{x}^{* \mu}\}_{\mu=1}^{P^*}$, we find that we can train our DAM for unsupervised classification by replacing the soft labels $q^{* \mu}_y$ in the effective loss (Eq. \ref{eq:effective_loss}) with the softened DAM predictions $\left( 1 - \varepsilon \right) \mathcal{P}_{\beta, \varsigma} \left( y | \mathbf{x}^{* \mu} ; \mathbf{w}, \mathbf{p} \right) + \varepsilon \frac{1}{C + 1}$, where $\varepsilon \in \left[ 0, 1 \right]$ and $\mathcal{P}_{\beta, \varsigma} \left( y | \mathbf{x}^{* \mu} ; \mathbf{w}, \mathbf{p} \right)$ is defined in Eq. (\ref{eq:effective_preds}). In this scenario, we are minimizing
\begin{equation}
    \label{eq:unsupervised_loss}
    \mathcal{L}_{\text{unsup}} \left( \mathbf{w}, \mathbf{p} \right) = -\frac{1}{P^*} \sum_{\mu = 1}^{P^*} \sum_{y = 0}^C \left[ \left( 1 - \varepsilon \right) \mathcal{P}_{\beta, \varsigma} \left( y | \mathbf{x}^{* \mu} ; \mathbf{w}, \mathbf{p} \right) + \varepsilon \frac{1}{C + 1} \right] \log \mathcal{P}_{\beta, \varsigma} \left( \mathbf{x}^{* \mu}, y | \mathbf{w}, \mathbf{p} \right).
\end{equation}
As before, we do so using constrained SGD (see Appendix \ref{app:weight_normalization}) with the initialization and learning rate described in Appendix \ref{app:initialization}. Equivalently, we can also view this algorithm as minimizing the combined loss
\begin{gather*}
    \label{eq:total_loss}
    \mathcal{L}_{\text{total}} \left( \mathbf{w}, \mathbf{p} \right) = \mathcal{L}_{\text{margin}} \left( \mathbf{w}, \mathbf{p} \right) + \lambda \mathcal{L}_{\text{cond}} \left( \mathbf{w}, \mathbf{p} \right), \numberthis \\
    \text{where} \quad \mathcal{L}_{\text{margin}} \left( \mathbf{w}, \mathbf{p} \right) = -\frac{1}{P^*} \sum_{\mu = 1}^{P^*} \log \mathcal{P}_{\beta, \varsigma} \left( \mathbf{x}^{* \mu} | \mathbf{w}, \mathbf{p} \right) \\
    \text{and} \quad \mathcal{L}_{\text{cond}} \left( \mathbf{w}, \mathbf{p} \right) = -\frac{1}{P^*} \sum_{\mu = 1}^{P^*} \sum_{y = 0}^C \mathcal{P}_{\beta, \varsigma} \left( y | \mathbf{x}^{* \mu} ; \mathbf{w}, \mathbf{p} \right) \log \mathcal{P}_{\beta, \varsigma} \left( y | \mathbf{x}^{* \mu} ; \mathbf{w}, \mathbf{p} \right),
\end{gather*}
where $\mathcal{P}_{\beta, \varsigma} \left( \mathbf{x}^{* \mu} | \mathbf{w}, \mathbf{p} \right) = \sum_{y = 0}^C \mathcal{P}_{\beta, \varsigma} \left( \mathbf{x}^{* \mu}, y | \mathbf{w}, \mathbf{p} \right)$ (see Eq. \ref{eq:effective_preds}). $\mathcal{L}_{\text{cond}} \left( \mathbf{w}, \mathbf{p} \right)$ is called the minimum entropy regularization term in unsupervised machine learning \cite{grandvalet2004semi}.
Intuitively, it encourages the DAM to learn different class weights $p^\gamma_y$ for each class $y$, which is not possible by minimizing only $\mathcal{L}_{\text{margin}} \left( \mathbf{w}, \mathbf{p} \right)$ because $\mathcal{P}_{\beta, \varsigma} \left( \mathbf{x}^{* \mu} | \mathbf{w}, \mathbf{p} \right)$ does not depend on $y$. Exploiting this characteristic, we train our DAM on patches of MNIST digits \cite{lecun1998gradient} and find that it learns reasonable latent classes $y = \argmax_{y^\prime} \left\{ p^\mu_{y^\prime} \right\}$ for the memories $\mathbf{w}^\mu$ (see Fig. \ref{fig:unsupervised_DAM_memories} and Figs. \ref{fig:unsupervised_DAM_memories_1}, \ref{fig:unsupervised_DAM_memories_2}, \ref{fig:unsupervised_DAM_memories_3} and \ref{fig:unsupervised_DAM_memories_4} of Appendix \ref{app:unsupervised_weights}). This approach could potentially be useful for feature extraction \cite{khalid2014survey}.

\subsection{Fast training with splitting steepest descent}
\label{sec:splitting}
We now explain how to further improve the training method of Section \ref{sec:learning_eff_loss}. More specifically, we use the saddle-point hierarchy derived in \ref{sec:hierarchy} to accelerate training.

\begin{figure}
    \centering
    \includegraphics[width=\linewidth]{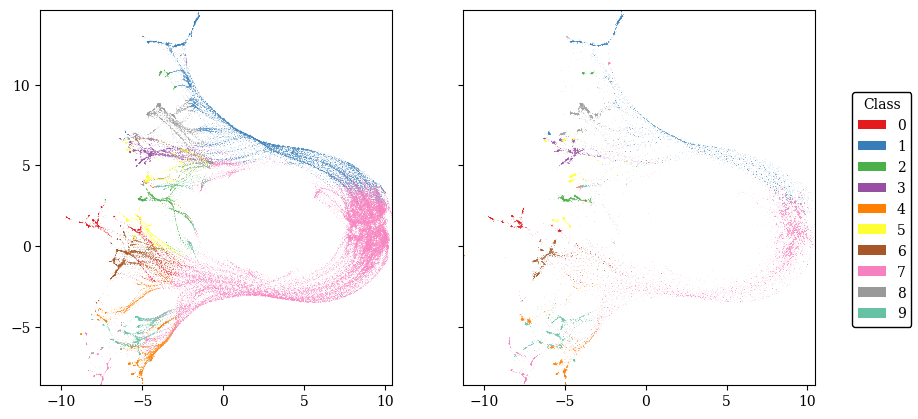}
    \caption{Overlaps $m^{\mu_* \mu} \left( \mathbf{x}^*, \mathbf{w} \right) = \sum_{i = 1}^N x^{* \mu_*}_i w^\mu_i$ between the first 1000 digits $\mathbf{x}^{* \mu_*} = \left\{ x^{* \mu_*}_i \right\}_{i = 1}^N$ of the MNIST training set \cite{lecun1998gradient} and the memories $\mathbf{w}^\mu = \left\{ w^\mu_i \right\}_{i = 1}^N$ of our dense associative memory (DAM) model while it is learning them. Each point is one of the high-dimensional magnetization vectors $m^{\cdot \mu} \left( \mathbf{x}^*, \mathbf{w} \right) = \left\{ m^{\mu_* \mu} \left( \mathbf{x}^*, \mathbf{w} \right) \right\}_{\mu_* = 1}^{P^*}$ projected onto a two-dimensional plane using the UMAP algorithm \cite{McInnes2018UMAP}. On the left, the $m^{\cdot \mu} \left( \mathbf{x}^*, \mathbf{w} \right)$ found during training without splitting steepest descent. On the right, the $m^{\cdot \mu} \left( \mathbf{x}^*, \mathbf{w} \right)$ found with splitting steepest descent (Alg. \ref{alg:splitting_descent}). In both cases, UMAP is trained on the $m^{\cdot \mu} \left( \mathbf{x}^*, \mathbf{w} \right)$ found without splitting steepest descent. The classes $y = \argmax_{y^\prime} \left\{ p^\mu_{y^\prime} \right\}$ of the memories $\mathbf{w}^\mu$ are {color coded} in the legend. 7 and 1 are the most numerous classes, so they have the top two largest entries in $\sum_{\gamma = 0}^P p^\mu_{y} = p_{\mathbf{q}} \left( y \right)$ (see Appendix \ref{app:initialization}), which is why the memories are classified as either 7 or 1 at the beginning of training.
    }
    \label{fig:umap_tree}
\end{figure}

Machine learning models that experience permutation symmetry breaking as described in Section \ref{sec:hierarchy}---such as K \& H's DAM \cite{boukacem2024waddington}, RBMs with binary units \cite{hou2019minimal, theriault2025modeling} and Gaussian mixtures \cite{kappen1993using, KAPPEN1995deterministic, rose1990statistical}
---have characteristic tree-shaped learning curves. Intuitively, permutation symmetry-breaking transitions are points where model parameters differentiate from each other, so they correspond to the bifurcations of the tree. In the left panel of Fig. (\ref{fig:umap_tree}), we show that the learning dynamics of our DAM also follows a tree of permutation symmetry-breaking transitions.

The saddle-point hierarchy principle introduced in Section \ref{sec:hierarchy} suggests the following idea to accelerate learning: train a relatively narrow DAM for cheap, then repeatedly duplicate (or ``split'') some of the hidden units, escape the corresponding saddle point and continue training. Intuitively, we expect to save a lot of computing resources by using relatively few hidden units at the start of training, when the learning dynamics follows a few branches close to the root of the learning dynamics tree. The splitting steepest descent algorithm introduced in \cite{wu2019splitting} formalizes this idea in an efficient and theoretically motivated way. We implement a variant of this algorithm that takes the constraint $\mathbf{w}^\mu \in S^{N - 1}$ into account, but is otherwise very similar to the original version (see Alg. \ref{alg:splitting_descent}), with fast splitting implemented as in \cite{wang2019energy}. The constraint $\mathbf{w}^\mu \in S^{N - 1}$ only matters in steps 5 and 11, which are also arguably the most conceptually difficult parts of Alg. (\ref{alg:splitting_descent}), so we explain them in detail in Appendix \ref{app:splitting}.
Fig. (\ref{fig:umap_tree}) shows that the learning dynamics tree of splitting steepest descent has a more sparsely populated trunk than that of SGD without splitting (see Fig. \ref{fig:umap_tree}). In other words, by using a relatively small number $P$ of memories at the beginning of training, we reduce the total number of values $P \times T$ that they take over a period of time $T$, which is consistent with our intuition that it can save computational resources. The situation could have been different if using fewer memories slowed down training.
\begin{algorithm}[H]
    \caption{Splitting steepest descent \cite{wu2019splitting, wang2019energy}}
    \label{alg:splitting_descent}
    \begin{algorithmic}[1]
        \State Preallocate space for a DAM with $P_{\text{max}}$ hidden units and the corresponding weights $\mathbf{w}$ and $\mathbf{p}$
        \State Initialize the weights $\mathbf{w}^\mu$ and $\mathbf{p}^\mu$ connected to the $P_{\text{cur}}$ first hidden units $\mu \in \left\{ 1, ..., P_{\text{cur}} \right\}$, as well as $\mathbf{p}^0$
        \State min $L \left( \mathbf{w}, \mathbf{p} \right)$ with SGD
        \While{$P_{\text{cur}} < P_{\text{max}}$}
        \State Identify a subset $\boldsymbol{\mu}_{\text{copy}} \subseteq \left\{ 1, ..., P_{\text{cur}} \right\}$ of $R \leq P_{\text{max}} - P_{\text{cur}}$ hidden units to split, \Return if empty
        \State Let $\boldsymbol{\mu}_{\text{paste}} = \left\{ P_{\text{cur}} + 1, ..., P_{\text{cur}} + R \right\}$
        \State Build weights $\mathbf{w}^{\boldsymbol{\mu}_{\text{paste}}} = \mathbf{w}^{\boldsymbol{\mu}_{\text{copy}}}$ for $\boldsymbol{\mu}_{\text{paste}}$
        \State Rescale $\mathbf{p}^{\boldsymbol{\mu}_{\text{copy}}} \gets \mathbf{p}^{\boldsymbol{\mu}_{\text{copy}}} / 2$ and $p_{\mathbf{h}} \left( {\boldsymbol{\mu}_{\text{split}}} \right) \gets p_{\mathbf{h}} \left( \boldsymbol{\mu}_{\text{split}} \right) / 2$
        \State Build weights $\mathbf{p}^{\boldsymbol{\mu}_{\text{paste}}} = \mathbf{p}^{\boldsymbol{\mu}_{\text{copy}}}$ and $p_{\mathbf{h}} \left( \boldsymbol{\mu}_{\text{paste}} \right) = p_{\mathbf{h}} \left( \boldsymbol{\mu}_{\text{copy}} \right)$ for $\boldsymbol{\mu}_{\text{paste}}$
        \State Update $P_{\text{cur}} \gets P_{\text{cur}} + R$
        \State Escape the saddle point by $2^{\text{nd}}$ order descent of $L \left( \mathbf{w}, \mathbf{p} \right)$ w.r.t. $\mathbf{w}$
        \State min $L \left( \mathbf{w}, \mathbf{p} \right)$ with SGD
        \EndWhile
        \State \Return \Comment{\colorbox{lightgray}{See Appendix \ref{app:splitting} and \cite{wu2019splitting, wang2019energy} for details about steps 5 and 11}}
    \end{algorithmic}
\end{algorithm}
We now establish a methodology that we will use to compare the training times of DAMs trained on MNIST \cite{lecun1998gradient} {and Fashion-MNIST \cite{xiao2017fashion}} using the effective loss (Eq. \ref{eq:effective_loss}) with and without splitting steepest descent. It is more interesting and meaningful to compare the training times of NNs with similar performance. Therefore, we pick hyperparameters such that DAMs trained with and without splitting have similar classification accuracy. For that purpose, we find the general region of hyperparameter space where DAMs trained without splitting steepest descent have the best classification accuracy. The accuracy does not change much in this region, so we pick generic hyperparameters inside. Next, we manually tune the hyperparameters of splitting steepest descent so that the resulting DAMs have comparable accuracy. The accuracy obtained for Fashion-MNIST is generically slightly higher with splitting steepest descent than
without it. In contrast, splitting steepest descent does not significantly affect the accuracy obtained for MNIST.

Once the hyperparameters are set, we collect statistics of the accuracy and training time. We run our experiment on a CPU and manually set the seed of pseudorandom number generation to make training deterministic and reproducible. Two DAMs with the same hyperparameters and the same seed are guaranteed to be trained using the same number of computer operations and to have the same classification accuracy after training. However, background processes and other external factors can change between two runs with the same seed, which adds noise to the bare training time that we want to measure. As such, we approximate the bare training time as the minimum over many runs with the same seed. We then calculate the average and standard deviation of the classification accuracy and bare training time over multiple seeds.

\begin{figure}
    \centering
    \hspace{7pt}\includegraphics[width=0.85\linewidth]{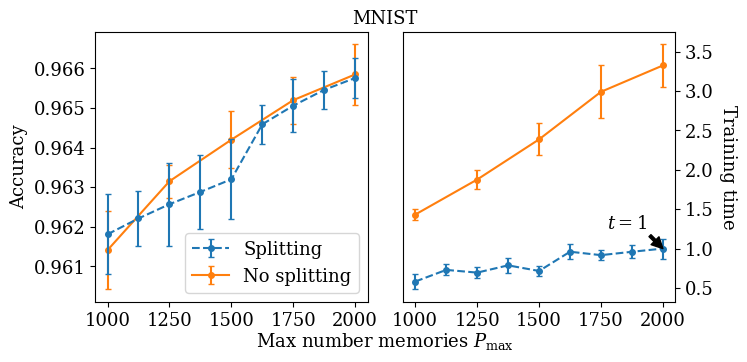}
    \includegraphics[width=0.85\linewidth]{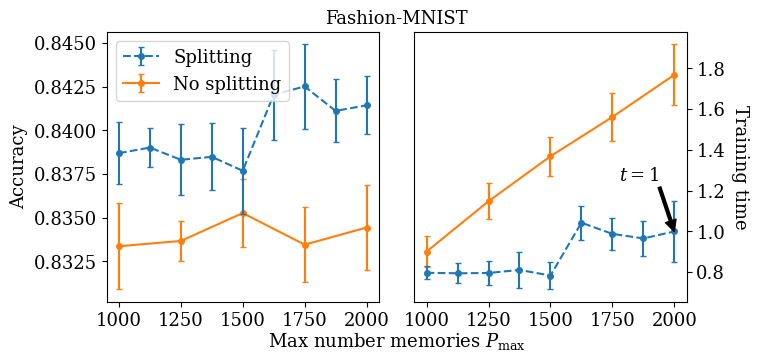}
    \caption{The classification accuracy and training time of dense associative memory (DAM) networks trained on MNIST \cite{lecun1998gradient} and Fashion-MNIST \cite{xiao2017fashion} with and without splitting as a function of the maximum number of hidden unit $P_{\text{max}}$. The round markers and their corresponding error bars show the means and standard deviations of the measurements made at each $P_{\text{max}}$. Statistics are collected from $10$ different random seeds. In the left panels, we verify that the DAMs with and without splitting have a similar accuracy. In the right panels, we compare their training times $t$. To facilitate this comparison, we normalize the $t$-axis such that the data point indicated by the arrow is at $t = 1$.}
    \label{fig:performance}
\end{figure}

In Fig. (\ref{fig:performance}), we compare DAM training time and accuracy with and without splitting steepest descent, using round markers and error bars to show their respective means and standard deviations as a function of the maximum number of hidden units $P_{\text{max}}$. Splitting stops at $P_{\text{max}}$ hidden units, and DAMs without splitting have $P_{\text{max}}$ hidden units from the start. All hidden units are split in each while-loop iteration of Alg. (\ref{alg:splitting_descent}), except for the last iteration, where exactly $P_{\text{max}} - P_{\text{cur}}$ hidden units are split. As wanted, DAMs have similar accuracy with and without splitting (left panels). Moreover, the reasonably small error bars of the training time indicate that the residual noise of the background processes is controlled and that the seed has a limited impact on the training speed (right panels). Splitting steepest descent has a significant speed advantage that scales very advantageously with $P_{\text{max}}$ (right panels). Without splitting, the training time is proportional to $P_{\text{max}}$. With splitting, it appears to be piecewise constant. The clearest sign that it increases with $P_{\text{max}}$ is the jump between $P_{\text{max}} = 1500$ and $P_{\text{max}} = 1625$, where the number of while-loop iterations in Alg. (\ref{alg:splitting_descent}) increases by 1. In other words, the run time of Alg. (\ref{alg:splitting_descent}) is better explained as proportional to $\left\lceil \log P_{\text{max}} \right\rceil$ than $P_{\text{max}}$ in the range of $P_{\text{max}}$ of our numerical experiment, which is a dramatic improvement over the run time without splitting. The benefit of the algorithm becomes evident primarily when the accuracy improves consistently across a wide range of hidden unit counts, i.e. when training a substantially large network is beneficial. This pattern, for example, does not apply to Fashion MNIST (see Fig. \ref{fig:performance}, bottom plot), where the accuracy gain from training a larger network is marginal. As a result, the need for strategies that accelerate training becomes less critical, although such acceleration still occurs.

\section{Conclusion}
In this work, we study a Dense Associative Memory model based on the framework of Boltzmann machines that is capable of learning interpretable solutions to both supervised and unsupervised problems.

We derive two sets of equations that respectively characterize the stationary points of DAMs trained on real data using maximum likelihood estimation and the fixed points of DAMs trained on synthetic data in the teacher-student setting.
Guided by their similarity, we then establish that the maximum likelihood equations are a special case of the teacher-student equations.
Building on this equivalence, we introduce an effective loss function that makes training significantly more stable thanks to a regularization parameter that mimics the noise present in the data generation process.

We show that the stationary points of the effective loss of DAMs with a given number of hidden units are saddle points of larger DAMs, and we considerably accelerate training using a custom implementation of the splitting steepest descent network-growing algorithm \cite{wu2019splitting} inspired by this \textit{saddle-point hierarchy principle}. In our numerical experiments, the training time of splitting steepest descent is approximately logarithmic in the number of hidden units, which is a huge improvement over the linear training time that we observe without splitting steepest descent. Further research is needed to clearly determine the asymptotic functional form of the training time.

In Section \ref{sec:learning_eff_loss}, we briefly discuss how our DAM is interpretable. The ingredients needed to make an interpretable neural network are not well understood, and we believe that our DAM offers a promising opportunity to explore that area. Studying the teacher-student setting with teacher inverse temperature $\beta^*$ less than order $N$ is another interesting research direction that we leave to future work. Beyond the framework of our model, \cite{rose1993constrained, rose1998deterministic} have designed a technique that trains Gaussian mixture models by replicating mixture components. In contrast with our work, it uses a handpicked cooling schedule and breaks permutation symmetries with random noise. By design, splitting steepest descent finds the directions in the loss landscape that allow NNs to escape saddle points with permutation symmetries as quickly as possible, but finding these directions (as described in Appendix \ref{app:splitting} and \cite{wu2019splitting, wang2019energy}) is more expensive than generating random noise. It would be interesting to study this tradeoff and identify the regimes where each method is preferable over the other. More generally, one could investigate how permutation symmetry breaking takes place in energy models related to DAMs, such as attentional BMs \cite{ota2023attention}, and study its relationship with the permutation symmetry-breaking \cite{hou2019minimal, theriault2025modeling} and dynamical transitions \cite{decelle2017spectral, decelle2018thermodynamics, bachtis2025cascade} found in other types of RBMs, whose careful analysis was also used to improve training \cite{bereux2025fast}. Recent works \cite{boukacem2024waddington, grishechkin2025mathematical} have made connections between DAM learning dynamics and cellular differentiation. Similarly, it could be possible to investigate whether there are similarities between cellular division and splitting steepest descent in our model. Finally, it would be interesting to see if splitting steepest descent and our proposed regularization technique can be used to improve the training of transformers and generative diffusion.

\section*{Data availability statement}
The code for the numerical experiments presented in this work is available in the following public repository \cite{theriault2025saddlesoftware}.

\section*{Acknowledgements}
This work was partially supported by project SERICS (PE00000014) under the MUR National Recovery and Resilience Plan
funded by the European Union - NextGenerationEU. The work was also supported by the project PRIN22TANTARI "Statistical Mechanics of Learning Machines: from algorithmic and information-theoretical limits to new biologically inspired paradigms" 20229T9EAT – CUP J53D23003640001. DT also acknowledges GNFM-Indam.

RT acknowledges the computational resources of the Center for High Performance Computing (CHPC) at the Scuola Normale Superiore of Pisa, as well as Carlo Lucibello and Matteo Negri for insightful discussions.

\section*{Conflicts of interest}
The authors have no conflicts of interest to declare.

\section*{Author contributions}
\textbf{Robin Thériault:} Conceptualization, Methodology, Software, Formal analysis, Investigation, Data curation, Writing -- original draft, Visualization. \textbf{Daniele Tantari:} Conceptualization, Validation, Writing -- original draft, Writing -- review and editing, Supervision, Project administration, Funding acquisition

\appendix
\section{Derivation of the model}
\label{app:model}

Consider the RBM Hamiltonian
\begin{align*}
    -\beta H \left[ \mathbf{x}, \mathbf{q}, \mathbf{h} ; \mathbf{J} \right] = \beta \sum_{\mu = 1}^P h_\mu \sum_{i = 1}^N w^\mu_i x_i + \beta \sum_{\mu = 1}^P h_\mu \sum_{y = 1}^C u^\mu_y q_y + \beta \sum_{\mu = 1}^P h_\mu b^\mu.
\end{align*}
where $\mathbf{J} = \left\{ \mathbf{w}, \mathbf{u}, \mathbf{b} \right\}$ is the set of all RBM parameters, i.e. $\mathbf{w} = \left\{ w^\mu_i \right\}_{1 \leq i \leq N}^{1 \leq \mu \leq P}$, $\mathbf{u} = \left\{ u^\mu_i \right\}_{1 \leq i \leq C}^{1 \leq \mu \leq P}$ and $\mathbf{b} = \left\{ b^\mu \right\}_{\mu = 1}^P$. The prior densities $\Prob_0 \left( \mathbf{h} \right)$ and $\Prob_0 \left( \mathbf{q} \right)$ are nonzero only when $\mathbf{h} \in \left\{ \mathbf{e}_{\gamma} \right\}_{\gamma = 0}^P$ and $\mathbf{q} \in \left\{ \mathbf{e}_y \right\}_{y = 0}^C$, in which case we have
\begin{align*}
    -\beta H \left[ \mathbf{x}, \mathbf{e}_y, \mathbf{e}_{\gamma}; \mathbf{J} \right]
    &= \begin{cases}
        0 &\quad \gamma = 0 \\
        \beta \sum_{i = 1}^N w^\gamma_i x_i + a^\gamma_y &\quad 0 < \gamma \leq P
    \end{cases} \\
    \text{where} \quad a^\mu_y
    &= \begin{cases}
        \beta b^\mu &\quad y = 0 \\
        \beta u^\mu_y + \beta b^\mu &\quad 0 < y \leq C
    \end{cases}
\end{align*}
For convenience, we write the corresponding Gibbs distribution $\Prob_\beta \left( \mathbf{x}, \mathbf{q} = \mathbf{e}_y, \mathbf{h} = \mathbf{e}_\gamma | \mathbf{J} \right)$ as $\Prob_\beta \left( \mathbf{x}, y, \mu | \mathbf{J} \right)$, and the priors $\Prob_0 \left( \mathbf{q} = \mathbf{e}_y \right)$ and $\Prob_0 \left( \mathbf{h} = \mathbf{e}_\gamma \right)$ as $\pi_{\mathbf{q}} \left( y \right)$ and $\pi_{\mathbf{h}} \left( \gamma \right)$, respectively. Given a uniform prior on $\mathbf{x}$, we find
\begin{align*}
    \Prob_{\beta} \left( \mathbf{x}, y, \mu | \mathbf{J} \right) = \frac{1}{Z_\beta \left( \mathbf{J} \right)} \pi_{\mathbf{q}} \left( y \right) \begin{cases}
        \pi_{\mathbf{h}} \left( 0 \right) &\quad \gamma = 0 \\
        \pi_{\mathbf{h}} \left( \gamma \right) \exp \left( a^\gamma_y \right) \exp \left( \beta \sum_{i = 1}^N w^\gamma_i x_i \right) &\quad 0 < \gamma \leq P,
    \end{cases}
\end{align*}
where $Z_\beta \left( \mathbf{J} \right)$ is an unknown normalization constant. The data $\mathbf{x}$ belonging to each cluster $\mu > 0$ follows a von Mises-Fisher (vMF) distribution $\Prob_{\beta} \left( \mathbf{x} | \mu, \mathbf{J} \right) \propto \exp \left( \beta \sum_{i = 1}^N w^\mu_i x_i \right)$ centered on $\mathbf{w}^\mu = \left\{ w^\mu_i \right\}_{i = 1}^N$ (see Appendix \ref{app:vmf_integration}). We assume that $\sqrt{\sum_{i = 1}^N \left( w^\mu_i \right)^2} = 1$ so that the corresponding cluster centroids $\mathbf{w}^\mu$ lie on the hypersphere $S^{N - 1}$ like the data $\mathbf{x} = \left\{ x_i \right\}_{i = 1}^N$. We marginalize $\Prob_{\beta} \left( \mathbf{x}, y, \mu | \mathbf{J} \right)$ over the hidden units $\mu$ and get
\begin{align*}
    \Prob_{\beta} \left( \mathbf{x}, y | \mathbf{J} \right) &= \frac{1}{Z_\beta \left( \mathbf{J} \right)} \pi_{\mathbf{q}} \left( y \right) \left[ \sum_{\mu = 1}^P \pi_{\mathbf{h}} \left( \mu \right) \exp \left( a^\mu_y \right) \exp \left( \beta \sum_{i = 1}^N w^\mu_i x_i \right) + \pi_{\mathbf{h}} \left( 0 \right) \right].
\end{align*}
We will now find the normalization constant $Z_\beta \left( \mathbf{J} \right)$. Marginalizing $\Prob_\beta \left( \mathbf{x}, y | \mathbf{J} \right)$ over $\mathbf{x}$, we obtain
\begin{align*}
    \Prob_{\beta} \left( y | \mathbf{J} \right) &= \int_{S^{N - 1}} d\mathbf{x} \Prob_{\beta} \left( \mathbf{x}, y | \mathbf{J} \right) \\
    &= \frac{1}{Z_\beta \left( \mathbf{J} \right)} \pi_{\mathbf{q}} \left( y \right) \int_{S^{N - 1}} d\mathbf{x} \left[ \sum_{\mu = 1}^P \pi_{\mathbf{h}} \left( \mu \right) \exp \left( a^\mu_y \right) \exp \left( \beta \sum_{i = 1}^N w^\mu_i x_i \right) + \pi_{\mathbf{h}} \left( 0 \right) \right] \\
    &= \frac{1}{Z_\beta \left( \mathbf{J} \right)} \pi_{\mathbf{q}} \left( y \right) \Omega_N \left( \beta \right) \sum_{\mu = 1}^P \pi_{\mathbf{h}} \left( \mu \right) \exp \left( a^\mu_y \right) + \Omega_N \left( 0 \right) \pi_{\mathbf{h}} \left( 0 \right) \\
    &= \frac{1}{Z_\beta \left( \mathbf{J} \right)} \pi_{\mathbf{q}} \left( y \right) \Omega_N \left( \beta \right) \left[ \sum_{\mu = 1}^P \pi_{\mathbf{h}} \left( \mu \right) \exp \left( a^\mu_y \right) + \pi_{\mathbf{h}} \left( 0 \right) \exp \left( a^0 \right) \right],
\end{align*}
where $a^0 = \log \left[ \Omega_N \left( 0 \right) / \Omega_N \left( \beta \right) \right]$. Normalization of $\Prob_\beta \left( y | \mathbf{J} \right)$ requires that
\begin{align*}
    \Prob_{\beta} \left( y | \mathbf{J} \right) &= \frac{\pi_{\mathbf{q}} \left( y \right) \left[ \sum_{\mu = 1}^P \pi_{\mathbf{h}} \left( \mu \right) \exp \left( a^\mu_y \right) + \pi_{\mathbf{h}} \left( 0 \right) \exp \left( a^0 \right) \right]}{\sum_{y^\prime = 0}^C \pi_{\mathbf{q}} \left( y^\prime \right) \left[ \sum_{\nu = 1}^P \pi_{\mathbf{h}} \left( \nu \right) \exp \left( a^{\nu}_{y^\prime} \right) + \pi_{\mathbf{h}} \left( 0 \right) \exp \left( a^0 \right) \right]},
\end{align*}
so we deduce that $Z_\beta \left( \mathbf{J} \right) = \Omega_N \left( \beta \right) \sum_{y = 0}^C \pi_{\mathbf{q}} \left( y \right) \left[ \sum_{\mu = 1}^P \pi_{\mathbf{h}} \left( \mu \right) \exp \left( a^\mu_y \right) + \pi_{\mathbf{h}} \left( 0 \right) \exp \left( a^0 \right) \right]$. We define
\begin{gather*}
    p^\gamma_y
    = \begin{cases}
        \frac{\pi_{\mathbf{q}} \left( y \right) \pi_{\mathbf{h}} \left( 0 \right) \exp \left( a^0 \right)}{\sum_{y^\prime = 0}^C \pi_{\mathbf{q}} \left( y^\prime \right) \left[ \sum_{\nu = 1}^P \pi_{\mathbf{h}} \left( \nu \right) \exp \left( a^{\nu}_{y^\prime} \right) + \pi_{\mathbf{h}} \left( 0 \right) \exp \left( a^0 \right) \right]} \quad \gamma = 0 \\
        \frac{\pi_{\mathbf{q}} \left( y \right) \pi_{\mathbf{h}} \left( \mu \right) \exp \left( a^\mu_y \right)}{\sum_{y^\prime = 0}^C \pi_{\mathbf{q}} \left( y^\prime \right) \left[ \sum_{\nu = 1}^P \pi_{\mathbf{h}} \left( \nu \right) \exp \left( a^{\nu}_{y^\prime} \right) + \pi_{\mathbf{h}} \left( 0 \right) \exp \left( a^0 \right) \right]} \quad 0 < \gamma \leq P,
    \end{cases}
\end{gather*}
from which we obtain
\begin{align*}
    \Prob_{\beta} \left( \mathbf{x}, y | \mathbf{J} \right) &= \sum_{\mu = 1}^P p^\mu_y \frac{\exp \left( \beta \sum_{i = 1}^N w^\mu_i x_i \right)}{\Omega_N \left( \beta \right)} + p^0_y \frac{1}{\Omega_N \left( 0 \right)}.
\end{align*}
The conditional distributions $\Prob_{\beta} \left( \mathbf{x} | y, \mathbf{J} \right)$ and their marginal $\Prob_{\beta} \left( \mathbf{x} | \mathbf{J} \right) = \sum_{y = 0}^C \Prob_{\beta} \left( \mathbf{x} | y, \mathbf{J} \right) \Prob_\beta \left( y | \mathbf{J} \right)$ are von Mises-Fisher mixtures \cite{banerjee2005clustering} with weights $\frac{p^\gamma_y}{\sum_{\nu = 0}^P p^\nu_y}$ and $\sum_{y = 0}^C p^\gamma_y$, respectively. As explained in Section \ref{sec:model}, we constrain the marginals $\sum_{y = 0}^C p^\gamma_y = \Prob_\beta \left( \gamma | \mathbf{J} \right)$ and $\sum_{\gamma = 0}^P p^\gamma_y = \Prob_\beta \left( y | \mathbf{J} \right)$ to be equal to fixed distributions $p_{\mathbf{h}} \left( \gamma \right)$ and $p_{\mathbf{q}} \left( y \right)$, respectively. Formally, this means that $\mathbf{p} = \left\{ p^\gamma_y \right\}_{0 \leq y \leq C}^{0 \leq \gamma \leq C}$ belongs to the transportation polytope with sum constraints $\sum_{\gamma = 0}^P p^\gamma_y = p_{\mathbf{q}} \left( y \right)$ and $\sum_{y = 0}^C p^\gamma_y = p_{\mathbf{h}} \left( \gamma \right)$ \cite{deloera2013combinatorics}.

\section{Integration of the von Mises-Fisher density}
\label{app:vmf_integration}
The von Mises-Fisher (vMF) distribution \cite{mardia1999directional} is an isotropic Gaussian distribution restricted to the $N - 1$ dimensional unit sphere $S^{N - 1}$. It takes the form
\begin{align*}
    p \left( \mathbf{x} \right) = \Omega \left( r \right)^{-1} \exp \left( \sum_{i = 1}^N r_i x_i \right),
\end{align*}
where $r = \sqrt{\sum_{i} \left[ r_i \right]^2}$ is called the concentration parameter. When $r = 0$, it reduces to the uniform distribution on the unit sphere, whose surface surface area $\frac{2 \pi^{N/2}}{\Gamma \left( N/2 \right)}$ is thus also the normalization constant $\Omega_N \left( 0 \right)$. When $r > 0$, we define the mean direction $\hat{\mathbf{r}} = \left\{ r_i / r \right\}_{i = 1}^N$ and find the normalization constant to be
\begin{align*}
    \label{eq:vmf_integration_constant}
    \Omega_N \left( r \right) &= \int_{S^{N - 1}} d \mathbf{x} \exp \left( \sum_{i = 1}^N r_i x_i \right) \numberthis \\
    &= \int_{S^{N - 1}} d \mathbf{x} \exp \left( r \sum_{i = 1}^N \hat{r}_i x_i \right) \\
    &= \Omega_{N-1} \left( 0 \right) \int_{-1}^{1} d u \left( 1 - u^2 \right)^{(N - 3)/2} \exp \left( r u \right) \\
    &= \Omega_N \left( 0 \right) \left( \frac{r}{2} \right)^{1 - N / 2} \Gamma \left( \frac{N}{2} \right) I_{N / 2 - 1} \left( r \right),
\end{align*}
where $I_n \left( x \right)$ is the modified Bessel function of the first kind of order $n$. The third line of (\ref{eq:vmf_integration_constant}) comes from the change of variables $u = \sum_{i = 1}^N \hat{r}_i x_i$, and the fourth line is a consequence of Poisson's Bessel function integral \cite{NIST:DLMF_0}. In the limit of large $N$ with $\rho = \frac{r}{N - 2} \approx \frac{r}{N}$, we find
\begin{gather*}
    \begin{aligned}
        \int_{S^{N - 1}} d \mathbf{x} \exp \left( \sum_{i = 1}^N r_i x_i \right)
        &\approx \frac{\Omega_N \left( 0 \right)}{\sqrt{2 \pi \left( N/2 - 1 \right)}} \left( \frac{N}{2} - 1 \right)^{1 - N / 2} \left( 1 + \left( 2 \rho \right)^2 \right)^{-1/4} \\
        &\quad \Gamma \left( \frac{N}{2} \right) \exp \left[ \left( \frac{N}{2} - 1 \right) \left( \eta \left( 2 \rho \right) + 1 \right) \right]
    \end{aligned} \\
    \text{where} \ \eta \left( x \right) = \left( 1 + x^2 \right)^{1/2} - 1 - \log \left[ 1 + \left( 1 + x^2 \right)^{1/2} \right] + \log 2,
\end{gather*}
by exploiting the large $N$ asymptotic expansion of $I_{N/2 - 1} \left( \left[ N/2 - 1 \right] \cdot 2\rho \right)$ found in \cite{Fröhlich1981transition, NIST:DLMF_1}. We use Stirling's approximation
\begin{align*}
    \Gamma \left( \frac{N}{2} \right)
    &\approx \sqrt{2 \pi \left( N/2 - 1 \right)} \left( \frac{N}{2} - 1 \right)^{N / 2 - 1} \exp \left( -\frac{N}{2} + 1 \right)
\end{align*}
to simplify it to
\begin{gather*}
    \label{eq:thermodynamic_approx_to_vmf}
    \begin{aligned}
        \int_{S^{N - 1}} d \mathbf{x} \exp \left( \sum_{i = 1}^N r_i x_i \right) &\approx \Omega_N \left( 0 \right) \left( 1 + \left( 2 \rho \right)^2 \right)^{-1/4} \exp \left[ \left( \frac{N}{2} - 1 \right) \eta \left( 2 \rho \right) \right]
    \end{aligned} \numberthis \\
    \text{where} \ \eta \left( x \right) = \left( 1 + x^2 \right)^{1/2} - 1 - \log \left[ 1 + \left( 1 + x^2 \right)^{1/2} \right] + \log 2.
\end{gather*}

\section{Normalization of the weights}
\label{app:weight_normalization}
In order to enforce $\sqrt{\sum_{i = 1}^N \left( w^\mu_i \right)^2} = 1$ at each SGD step, we project $\mathbf{w}^\mu$ and the gradient of the loss with respect to $\mathbf{w}^\mu$ onto the unit sphere $S^{N - 1}$ and the tangent space of $\mathbf{w}^\mu$, respectively. We divide $w^\mu_i$ by its norm to project it onto $S^{N - 1}$, and we multiply the gradient by $\delta_{j k} - w^\mu_j w^\mu_k$ to project it onto the tangent space. Projecting the gradient onto the tangent space is a mathematically sound way to obtain the gradient of a function restricted to a manifold embedded in $\mathbb{R}^N$ \cite{barilari2023lecture}.

The tasks of deriving the saddle-point equations for $\bar{p}^\gamma_y$ (Eqs. \ref{eq:saddle-point}), finding the stationarity condition of the loss with respect to the class weights $p^\gamma_y$ (Eqs. \ref{eq:stationarity}) and efficiently training $p^\gamma_y$ all amount to solving the extremization problem
\begin{align*}
    \label{eq:extremization}
    \Extr_{\mathbf{p}, \omega, \lambda} \left\{ f \left( \mathbf{p} \right) + \sum_{y = 0}^C \lambda_y \left( \sum_{\gamma = 0}^P p^\gamma_y - p_{\mathbf{q}} \left( y \right) \right) + \sum_{\gamma = 0}^P \omega^{\gamma} \left( \sum_{y = 0}^C p^\gamma_y - p_{\mathbf{h}} \left( \gamma \right) \right) \right\}, \numberthis
\end{align*}
where $\lambda_y$ and $\omega^{\gamma}$ are Lagrange multipliers that enforce the constraints $\sum_\gamma p^\gamma_y = p_{\mathbf{q}} \left( y \right)$ and $\sum_y p^\gamma_y = p_{\mathbf{h}} \left( \gamma \right)$. In the former task, $f \left( \mathbf{p} \right)$ is the free entropy (Eq. \ref{eq:var_free_entropy}) at fixed $\mathbf{m}$ and $\mathbf{\hat{m}}$. In the latter two, it is the negative log-likelihood loss (Eq. \ref{eq:loss}) at a given $\mathbf{w}$. To find the saddle-point equations and the stationarity conditions of the loss, we derive an implicit solution of Eq. (\ref{eq:extremization}) that is useful in analytical calculations. On the other hand, to train $p^\gamma_y$, we design an algorithm to quickly compute a numerical solution of Eq. (\ref{eq:extremization}). We start by noting that, since the extrema of Eq. (\ref{eq:extremization}) are the points where the gradient vanishes, they take the form
\begin{align*}
    \label{eq:nonlinear_system}
    \partial_{p^\gamma_y} f \left( \mathbf{p} \right) &= \lambda_y + \omega^{\gamma} \\
    \sum_{\gamma = 0}^P p^\gamma_y &= p_{\mathbf{q}} \left( y \right) \numberthis \\
    \sum_{y = 0}^C p^\gamma_y &= p_{\mathbf{h}} \left( \gamma \right).
\end{align*}
Define $\bar{p}^{\gamma}_y = \partial_{p^\gamma_y} \exp \left( f \left( \mathbf{p} \right) \right) = p^\gamma_y \partial_{p^\gamma_y} f \left( \mathbf{p} \right)$, then
\begin{align*}
    \frac{\bar{p}^{\gamma}_y}{p^\gamma_y} &= \lambda_y + \omega^{\gamma},
\end{align*}
along with the previously established row and column sum constraints on $\mathbf{p}$. Rearranging terms, we get
\begin{align*}
    p^\gamma_y &= \frac{1}{\lambda_y + \omega^\gamma} \bar{p}^{\gamma}_y.
\end{align*}
Using the row and column constraints $\sum_{\gamma = 0}^P p^\gamma_y = p_{\mathbf{q}} \left( y \right)$ and $\sum_{y = 0}^C p^\gamma_y = p_{\mathbf{h}} \left( \gamma \right)$, we find that the Lagrange multipliers $\lambda_y$ and $\omega^\gamma$ solve the nonlinear equations
\begin{align*}
    \label{eq:nonlinear}
    \lambda_y &= \frac{1}{p_{\mathbf{q}} \left( y \right)} \sum_{\gamma = 0}^P \frac{\lambda_y}{\lambda_y + \omega^\gamma} \bar{p}^{\gamma}_y \\
    \omega^\gamma &= \frac{1}{p_{\mathbf{h}} \left( \gamma \right)} \sum_{y = 0}^C \frac{\omega^\gamma}{\lambda_y + \omega^\gamma} \bar{p}^{\gamma}_y, \numberthis
\end{align*}
For conciseness, we define $\zeta^\gamma_y \left( \mathbf{\bar{p}} ; p_{\mathbf{h}} \right) = \lambda_y \left( \mathbf{\bar{p}} ; p_{\mathbf{h}} \right) + \omega^\gamma \left( \mathbf{\bar{p}} ; p_{\mathbf{h}} \right)$, where
$\lambda_y \left( \mathbf{\bar{p}} ; p_{\mathbf{h}} \right)$ and $\omega^\gamma \left( \mathbf{\bar{p}} ; p_{\mathbf{h}} \right)$ are the $\lambda_y$ and $\omega^\gamma$ solving Eqs. (\ref{eq:nonlinear}) at given $P$ and $p_{\mathbf{h}}$. Using these definitions, we find the implicit solution $p^\gamma_y = \bar{p}^{\gamma}_y / \zeta^\gamma_y \left( \mathbf{\bar{p}} ; p_{\mathbf{h}} \right)$. As wanted, this equation is useful in analytical calculations involving the saddle-point equations and the stationarity conditions of the loss. However, it is also quite slow to solve numerically for a given $\mathbf{\bar{p}}$, as reported in multiple studies \cite{uribe1966information, hewings1980exchanging, MCNEIL1985note}. Therefore, although we can train $p^\gamma_y$ by iterating $p^\gamma_y = \bar{p}^{\gamma}_y / \zeta^\gamma_y \left( \mathbf{\bar{p}} ; p_{\mathbf{h}} \right)$, it is not a very efficient method.

In order to efficiently train $p^\gamma_y$, we devise a faster way to solve Eqs. (\ref{eq:nonlinear_system}) than through Eq. (\ref{eq:nonlinear}). Exponentiating both sides of the first line of Eqs. (\ref{eq:nonlinear_system}), we find
\begin{align*}
    \exp \left( \eta \partial_{p^\gamma_y} f \left( \mathbf{p} \right) \right) \exp \left( -\eta \lambda_y \right) \exp \left( -\eta \omega^\gamma \right) &= 1 \\
    \exp \left( -\eta \omega^\gamma \right) p^\gamma_y \exp \left( \eta \partial_{p^\gamma_y} f \left( \mathbf{p} \right) \right) \exp \left( -\eta \lambda_y \right) &= p^\gamma_y,
\end{align*}
where $\eta$ is an arbitrary scalar that will play the role of a learning rate. Fix $k^\gamma_y = p^\gamma_y \exp \left( \eta \partial_{p^\gamma_y} f \left( \mathbf{p}  \right) \right)$. By the Sinkhorn-Knopp theorem \cite{sinkhorn1964relationship, knopp1967concerning}, there is a rescaled matrix of the form $p^{\prime \gamma}_y = D^\gamma_L \left( \mathbf{k} \right) k^\gamma_y D^y_R \left( \mathbf{k} \right)$ that satisfies the same constraints $p^{\prime \gamma}_y \geq 0$, $\sum_{\gamma = 0}^P p^{\prime \gamma}_y = p_{\mathbf{q}} \left( y \right)$ and $\sum_{y = 0}^C p^{\prime \gamma}_y = p_{\mathbf{h}} \left( \gamma \right)$ as $p^\gamma_y$ if some technical conditions are satisfied \cite{idel2016review}. Moreover, if $p^{\prime \gamma}_y$ exists, then it is unique \cite{MENON1969spectrum, Hershkowitz1988classifications}, and we can quickly compute suitable scaling factors $D^\gamma_L \left( \mathbf{k} \right)$ and $D^\gamma_R \left( \mathbf{k} \right)$ for $\mathbf{k} = \left\{ k^\gamma_y \right\}^{0 \leq \gamma \leq P}_{0 \leq y \leq C}$ using the Sinkhorn-Knopp algorithm first proposed in \cite{deming1940least} and theoretically justified in subsequent works \cite{sinkhorn1964relationship, knopp1967concerning}. See \cite{idel2016review} for a concise review of all these points. $p^{\prime \gamma}_y$ is generally not equal to $p^\gamma_y$. However, we observe that the iteration
\begin{align*}
    \label{eq:sinkhorn_iteration}
    k^\gamma_y \left( t \right) &= p^\gamma_y \left( t \right) \exp \left( \eta \partial_{p^\gamma_y \left( t \right)} f \left( \mathbf{p} \left( t \right) \right) \right) \\
    p^\gamma_y \left( t + 1 \right) &= D^\gamma_L \left( \mathbf{k} \left( t \right) \right) k^\gamma_y \left( t \right) D^y_R \left( \mathbf{k} \left( t \right) \right) \numberthis
\end{align*}
converges at small $\eta$ (for example $\sim 0.1 / P$), which means that it must converge to a solution of Eqs. (\ref{eq:nonlinear_system}). Since the Sinkhorn-Knopp algorithm is fast, it is significantly more efficient to iterate Eqs. (\ref{eq:sinkhorn_iteration}) than $p^\gamma_y = \bar{p}^{\gamma}_y / \zeta^\gamma_y \left( \mathbf{\bar{p}} ; p_{\mathbf{h}} \right)$ to solve Eqs. (\ref{eq:nonlinear_system}).

In practice, we train $p^\gamma_y$ using a stochastic variant of Eqs. (\ref{eq:sinkhorn_iteration}) where the gradient is estimated over small batches of data and smoothed with a momentum hyperparameter.
Furthermore, we multiply the gradient by $\delta_{y y^\prime} - p^\gamma_y / p_{\mathbf{h}} \left( \gamma \right)$ to reduce the size of the components of $\exp \left( \eta \partial_{p^\gamma_y \left( t \right)} f \left( \mathbf{p} \left( t \right) \right) \right)$ that move $p^\gamma_y$ away from the constraints $\sum_{\gamma = 0}^P p^\gamma_y = p_{\mathbf{q}} \left( y \right)$ and $\sum_{y = 0}^C p^\gamma_y = p_{\mathbf{h}} \left( \gamma \right)$.
This step is a simple approximation of the gradient projection proposed in \cite{douik2019manifold}.

\section{Initialization and learning rate}
\label{app:initialization}
To make all our figures, except Fig. (\ref{fig:umap_tree}), we initialize the memories $\mathbf{w}^\mu$ uniformly at random on the unit hypersphere $S^{N - 1}$ \cite{muller1959note}. To make Fig. \ref{fig:umap_tree}, we instead use the algorithm of \cite{pinzon2023fast} to sample the initial memories $\mathbf{w}^\mu$ from a vMF distribution (see Appendix \ref{app:vmf_integration}) with mean direction $\tilde{\mathbf{x}}^* = \bar{\mathbf{x}}^* / \| \bar{\mathbf{x}}^* \|$, where $\bar{\mathbf{x}}^* = \frac{1}{N} \sum_{\mu_* = 1}^{P^*} \mathbf{x}^{* \mu_*}$ is the mean of the patterns $\mathbf{x}^{* \mu_*}$ and $\| \bar{\mathbf{x}}^* \| = \sqrt{\sum_{i = 1}^N \left[ \bar{x}^*_i \right]^2}$. By construction, the mean direction $\Tilde{\mathbf{x}}^*$ is the ordered solution of Eq. (\ref{eq:saddle-point}) with the most permutation symmetries, or in other words the root of the learning dynamics tree shown in Fig. \ref{fig:umap_tree}, so initializing the memories around it helps reveal the tree structure.

We use a learning rate of $\eta \sim 0.1$ to train the memories $\mathbf{w}^\mu$. Based on our experience, $\mathbf{p}$ trains well when its own learning rate is approximately $\left( 1 + \frac{p_{\mathbf{h}} \left( 0 \right)}{1 - p_{\mathbf{h}} \left( 0 \right)} \right) P$ times smaller. Without this rescaling, the multiplicative update factor $\exp \left( \eta \partial_{p^\gamma_y} L \left( \mathbf{w}, \mathbf{p} \right) \right)$ described in Appendix \ref{app:weight_normalization} can be relatively large even when $\eta$ is relatively small, making it ill behaved. Equivalently, we can also train $g^\gamma_y = \left( 1 + \frac{p_{\mathbf{h}} \left( 0 \right)}{1 - p_{\mathbf{h}} \left( 0 \right)} \right) P \mathbf{p}$ with the same learning rate $\eta$ as $\mathbf{w}$. We adopt this approach in our code available at \cite{theriault2025saddlesoftware}.

Let $P_0 = \frac{p_{\mathbf{h}} \left( 0 \right)}{1 - p_{\mathbf{h}} \left( 0 \right)} P$ so that $g^\gamma_y = \left( P + P_0 \right) p^\gamma_y$. In terms of $g^{\text{fixed}, \gamma}_y = \left( P + P_0 \right) p^{\text{fixed}, \gamma}_y$, the duplicated parameters (Eq. \ref{eq:fixed_point}) of the saddle-point hierarchy principle are
{\allowdisplaybreaks
\begin{align*}
    \bar{x}^{\text{dupli}, \mu}_i
    &= \begin{cases}
        \bar{x}^{\text{fixed}, \mu}_i &\quad 0 < \mu \leq P \\
        \bar{x}^{\text{fixed}, \mu - P}_i &\quad P < \mu \leq P + R \\
    \end{cases} \\
    \bar{g}^{\text{dupli}, \gamma}_y
    &= \frac{P + R}{P} \begin{cases}
        \bar{g}^{ \text{fixed}, 0}_y &\quad \gamma = 0 \\
        \frac{1}{2} \bar{g}^{ \text{fixed}, \gamma}_y &\quad 0 < \gamma \leq R \\
        \bar{g}^{ \text{fixed}, \gamma}_y &\quad R < \gamma \leq P \\
        \frac{1}{2} \bar{g}^{ \text{fixed}, \gamma - P}_y &\quad P < \gamma \leq P + R \\
    \end{cases} \\
    m^{\text{dupli}, \mu_* \gamma}
    &= \begin{cases}
        m^{\text{fixed}, \mu_* 0} &\quad \gamma = 0 \\
        m^{\text{fixed}, \mu_* \gamma} &\quad 0 < \gamma \leq P \\
        m^{\text{fixed}, \mu_*, \gamma - P} &\quad P < \gamma \leq P + R \\
    \end{cases} \\
    g^{\text{dupli}, \gamma}_y
    &= \frac{P + R}{P} \begin{cases}
        g^{\text{fixed}, 0}_y &\quad \gamma = 0 \\
        \frac{1}{2} g^{\text{fixed}, \gamma}_y &\quad 0 < \gamma \leq R \\
        g^{\text{fixed}, \gamma}_y &\quad R < \gamma \leq P \\
        \frac{1}{2} g^{\text{fixed}, \gamma - P}_y &\quad P < \gamma \leq P + R \\
    \end{cases}
\end{align*}
\begin{align*}
    \text{along with} \quad g_{\mathbf{h}} \left( \gamma \right)
    &= \frac{P + R}{P} \begin{cases}
        g_{\mathbf{h}}^{\text{given}} \left( 0 \right) &\quad \gamma = 0 \\
        \frac{1}{2} g_{\mathbf{h}}^{\text{given}} \left( \gamma \right) &\quad 0 < \gamma \leq R \\
        g_{\mathbf{h}}^{\text{given}} \left( \gamma \right) &\quad R < \gamma \leq P \\
        \frac{1}{2} g_{\mathbf{h}}^{\text{given}} \left( \gamma - P \right) &\quad P < \gamma \leq P + R,
    \end{cases}
\end{align*}}%
where $g_{\mathbf{h}}^{\text{given}} \left( \gamma \right) = \left( P + P_0 \right) p_{\mathbf{h}}^{\text{given}} \left( \gamma \right)$ and $g_{\mathbf{h}} \left( \gamma \right) = \left( P + P_0 \right) p_{\mathbf{h}} \left( \gamma \right)$. The newly introduced scaling factor of $\frac{P + R}{P}$ must be taken into account in our implementation of splitting steepest descent (Alg. \ref{alg:splitting_descent}), which gives Alg. (\ref{alg:rescaled_splitting_descent}), shown below.
\begin{algorithm}[H]
    \caption{Rescaled splitting steepest descent \cite{wu2019splitting, wang2019energy}}
    \label{alg:rescaled_splitting_descent}
    \begin{algorithmic}[1]
        \State Preallocate space for a DAM with $P_{\text{max}}$ hidden units and the corresponding weights $\mathbf{w}$ and $\mathbf{g}$
        \State Initialize the weights $\mathbf{w}^\mu$ and $\mathbf{g}^\mu$ connected to the $P_{\text{cur}}$ first hidden units $\mu \in \left\{ 1, ..., P_{\text{cur}} \right\}$, as well as $\mathbf{g}^0$
        \State min $L \left( \mathbf{w}, \mathbf{g} \right)$ with SGD
        \While{$P_{\text{cur}} < P_{\text{max}}$}
        \State Identify a subset $\boldsymbol{\mu}_{\text{copy}} \subseteq \left\{ 1, ..., P_{\text{cur}} \right\}$ of $R \leq P_{\text{max}} - P_{\text{cur}}$ hidden units to split, \Return if empty
        \State Let $\boldsymbol{\mu}_{\text{paste}} = \left\{ P_{\text{cur}} + 1, ..., P_{\text{cur}} + R \right\}$ and $\boldsymbol{\mu}_{\text{dupli}} = \left\{ 1, ..., P_{\text{cur}} + R \right\}$
        \State Build weights $\mathbf{w}^{\boldsymbol{\mu}_{\text{paste}}} = \mathbf{w}^{\boldsymbol{\mu}_{\text{copy}}}$ for $\boldsymbol{\mu}_{\text{paste}}$
        \State Rescale $\mathbf{g}^{\boldsymbol{\mu}_{\text{copy}}} \gets \mathbf{g}^{\boldsymbol{\mu}_{\text{copy}}} / 2$ and $g_{\mathbf{h}} \left( \boldsymbol{\mu}_{\text{split}} \right) \gets g_{\mathbf{h}} \left( \boldsymbol{\mu}_{\text{split}} \right) / 2$
        \State Build weights $\mathbf{g}^{\boldsymbol{\mu}_{\text{paste}}} = \mathbf{g}^{\boldsymbol{\mu}_{\text{copy}}}$ and $g \left( \boldsymbol{\mu}_{\text{paste}} \right) = g \left( \boldsymbol{\mu}_{\text{copy}} \right)$ for $\boldsymbol{\mu}_{\text{paste}}$
        \State Rescale $\mathbf{g}^{\boldsymbol{\mu}_{\text{dupli}}} \gets \frac{P_{\text{cur}} + R}{P_{\text{cur}}} \mathbf{g}^{\boldsymbol{\mu}_{\text{dupli}}}$ and $g_{\mathbf{h}} \left( \boldsymbol{\mu}_{\text{dupli}} \right) \gets \frac{P_{\text{cur}} + R}{P_{\text{cur}}} g_{\mathbf{h}} \left( \boldsymbol{\mu}_{\text{dupli}} \right)$
        \State Update $P_{\text{cur}} \gets P_{\text{cur}} + R$
        \State Escape the saddle point by $2^{\text{nd}}$ order descent of $L \left( \mathbf{w}, \mathbf{g} \right)$ w.r.t. $\mathbf{w}$
        \State min $L \left( \mathbf{w}, \mathbf{g} \right)$ with SGD
        \EndWhile
        \State \Return \Comment{\colorbox{lightgray}{See Appendix \ref{app:splitting} and \cite{wu2019splitting, wang2019energy} for details about steps 5 and 12}}
    \end{algorithmic}
\end{algorithm}
We initialize the weights $\mathbf{g}^\gamma$ and the hidden unit distribution $g_{\mathbf{h}} \left( \gamma \right)$ according to
\begin{gather*}
    g^\gamma_y
    = \begin{cases}
        P_0 p_{\mathbf{q}} \left( y \right) &\quad \gamma = 0 \\
        p_{\mathbf{q}} \left( y \right) &\quad \gamma > 0
    \end{cases} \\
    \text{and} \quad g_{\mathbf{h}} \left( \gamma \right)
    = \begin{cases}
        \frac{P_0}{P + P_0} &\quad \gamma = 0 \\
        \frac{1}{P + P_0} &\quad \gamma > 0.
    \end{cases}
\end{gather*}
When we train our DAM without splitting steepest descent, we set $P_0 = 1$. As discussed in \cite{BARVINOK2010WHAT, good1963maximum}, the resulting weights are, in some sense, the most ``typical'' for the constraints $g^\gamma_y \geq 0$, $\sum_{\gamma = 0}^P g^\gamma_y = \left( P + 1 \right) p_{\mathbf{q}} \left( y \right)$ and $\sum_{y = 0}^C g^\gamma_y = \left( P + 1 \right) p_{\mathbf{h}} \left( \gamma \right) = 1$ (see Section \ref{sec:model}). With splitting steepest descent, we set $P_0 = P_{\text{cur}}/P_{\text{final}}$ to ensure that $g_{\mathbf{h}} \left( 0 \right) = \sum_{y = 0}^C g^0_y$ is approximately the same size as the other entries of $g_{\mathbf{h}} \left( \gamma \right)$ after training. For simplicity, we set $p_{\mathbf{q}} \left( y \right)$ to the proportions of classes in the data during supervised training. For unsupervised training (see Section \ref{sec:interpretability}), we break the permutation symmetry between the different classes by using different values for all the entries of $p_{\mathbf{q}} \left( y \right)$. Otherwise, class weights with unbroken permutation symmetries stay stuck at their initial conditions.
\pagebreak

\section{Important symbols introduced in the Model Section}
\label{app:model_summary}
The following table summarizes the important symbols introduced in Section \ref{sec:model} and their meaning.
\begin{table}[h!]
    \centering
    \begin{tabular}{ |c|c| }
        \hline
        Symbol & Meaning \\
        \hline\hline
        $\mathbf{x} = \left\{ x_i \right\}_{i = 1}^N$ & Data layer \\
        \hline
        $\mathbf{h} = \left\{ h_\mu \right\}_{\mu = 1}^P$ & Hidden layer \\
        \hline
        $\mathbf{q} = \left\{ q_y \right\}_{y = 1}^C$ & Class layer \\
        \hline
        $N$ & \# data units \\
        \hline
        $P$ & \# hidden units (and clusters) \\
        \hline
        $C$ & \# class units (and classes) \\
        \hline
        $\mathbf{J}$ & Set of all trainable weights \\
        \hline
        $\mathbf{w} = \left\{ w_i^\mu \right\}_{1 \leq i \leq N}^{1 \leq \mu \leq P}$ & Weight matrix of the data layer \\
        \hline
        $\mathbf{p} = \left\{ p^\gamma_y \right\}_{0 \leq y \leq C}^{0 \leq \gamma \leq P}$ & Weight matrix of the class layer \\
        \hline
        $p_{\mathbf{q}} \left( y \right)$ & Marginal distribution  $\sum_{\gamma = 0}^P p^\gamma_y $ \\
        \hline
        $p_{\mathbf{h}} \left( \gamma \right)$ & Marginal distribution $\sum_{y = 0}^C p^\gamma_y $ \\
        \hline
        $\beta$ & Inverse temperature \\
        \hline
        $\Omega_N \left( \beta \right)$ & Normalization constant of the vMF distribution \\
        \hline
        $\mathbf{x}^{* \mu} = \left\{ x^{* \mu}_i \right\}_{i = 1}^N$ & Pattern from a given dataset \\
        \hline
        $\mathbf{q}^{* \mu} = \left\{ q^{* \mu}_y \right\}_{y = 0}^C$ & Soft label of the pattern \\
        \hline
        $P^*$ & \# hidden units (and clusters) of the teacher \\
        \hline
        $\mathbf{w}^* = \left\{ w_i^{* \mu} \right\}_{1 \leq i \leq N}^{1 \leq \mu \leq P}$ & Weight matrix of the data layer of the teacher DAM \\
        \hline
        $\mathbf{p}^* = \left\{ p^{* \gamma}_y \right\}_{0 \leq y \leq C}^{0 \leq \gamma \leq P}$ & Weight matrix of the class layer of the teacher DAM \\
        \hline
        $\mathbf{g}^*$ & Rescaled weight matrix $\mathbf{g}^* = P^* \mathbf{p}^*$ of the class layer of the teacher DAM \\
        \hline
        $p_{\mathbf{q}}^* \left( y \right)$ & Marginal distribution  $\sum_{\gamma = 0}^P p^{* \gamma}_y $ \\
        \hline
        $p_{\mathbf{h}}^* \left( \gamma \right)$ & Marginal distribution  $\sum_{y = 0}^C p^{* \gamma}_y $ \\
        \hline
        $\beta^*$ & Inverse temperature of the teacher \\
        \hline
        $\mathcal{D} = \left\{ \mathbf{x}^c, y^c \right\}_{c = 1}^M$ & Set of data points $\mathbf{x}^c$ and labels $y^c$ generated by the teacher \\
        \hline
        $M$ & Number of data points generated by the teacher \\
        \hline
        $\alpha = M/N$ & Load of data generated by the teacher \\
        \hline
    \end{tabular}
    \caption{Summary of the important symbols introduced in Section \ref{sec:model} and their meaning.}
    \label{tab:notation_table}
\end{table}

\section{Stationarity conditions of the loss}
\label{app:stationarity}
Since we constrain the memories $\mathbf{w}^\mu$ to have unit norm (see Section \ref{sec:model}), the method of Lagrange multipliers tells us any set of memories $\mathbf{w}$ that minimizes Eq. (\ref{eq:loss}) must solve the extremization problem
\begin{align*}
    \Extr_{\mathbf{w}, \varphi} \left\{ L \left( \mathbf{w}, \mathbf{p} \right) + \frac{1}{2} \sum_{\mu = 1}^P \varphi^\mu \left( \sum_{i = 1}^N \left[ w^\mu_i \right]^2 - 1 \right) \right\}.
\end{align*}
The extrema are the points where the gradient vanishes, so they take the form
\begin{gather*}
    \partial_{w^\mu_i} L \left( \mathbf{w}, \mathbf{p} \right) + \varphi^\mu w^\mu_i = 0 \\
    \sum_{i = 1}^N \left[ w^\mu_i \right]^2 = 1
\end{gather*}
for all $1 \leq \mu \leq P$. We calculate $\partial_{w^\mu_i} L \left( \mathbf{w}, \mathbf{p} \right)$ and write the solution in the more explicit form
\begin{align*}
   \bar{w}^\mu_i &= \sum_{\mu_* = 1}^{P^*} x^{* \mu_*}_i \sum_{y = 0}^C q^{* \mu_*}_y \sigma_\mu \left( \beta m^{\mu_*} \left( \mathbf{x}^*, \mathbf{w} \right) + \log \left[ \mathbf{p}_y \right] \right) \\
    w^\mu_i &= \frac{\bar{w}^\mu_i}{\sqrt{\sum_{j = 1}^N \left[\bar{w}^\mu_j \right]^2}},
\end{align*}
for all $1 \leq \mu \leq P$, where $\sigma_\gamma \left( x^{\gamma_*} \right) = \frac{\exp \left( x^{\gamma_* \gamma} \right)}{\sum_{\nu = 0}^P \exp \left( x^{\gamma_* \nu} \right)}$, $m^{\mu_* \mu} \left( \mathbf{x}^*, \mathbf{w} \right) = \sum_{i = 1}^N x^{* \mu_*}_i w^\mu_i$ for $1 \leq \mu \leq P$ and $m^{\mu_* 0} \left( \mathbf{x}, \mathbf{w} \right) = \frac{1}{\beta} \log \left[ \Omega_N \left( \beta \right) / \Omega_N \left( 0 \right) \right]$.

Similarly, any set of class weights $\mathbf{p}$ that minimizes Eq. (\ref{eq:loss}) must solve the extremization problem
\begin{align*}
    \Extr_{\mathbf{p}, \omega, \lambda} \left\{ L \left( \mathbf{w}, \mathbf{p} \right) + \sum_{y = 0}^C \lambda_y \left( \sum_{\gamma = 0}^P p^\gamma_y - p_{\mathbf{q}} \left( y \right) \right) + \sum_{\gamma = 0}^P \omega^{\gamma} \left( \sum_{y = 0}^C p^\gamma_y - p_{\mathbf{h}} \left( \gamma \right) \right) \right\}.
\end{align*}
In Appendix \ref{app:weight_normalization}, we show that its solution is
\begin{align*}
    \bar{p}^\gamma_y &= \sum_{\mu_* = 1}^{P^*} q^{* \mu_*}_y \sigma_\gamma \left( \beta m^{\mu_*} \left( \mathbf{x}, \mathbf{w} \right) + \log \left[ \mathbf{p}_y \right] \right) \\
    p^\gamma_y &= \frac{\bar{p}^\gamma_y}{\zeta^\gamma_y \left( \mathbf{\bar{p}} ; p_{\mathbf{h}} \right)}
\end{align*}
for all $0 \leq \gamma \leq P$, where the normalization constant $\zeta^\gamma_y \left( \mathbf{\bar{p}} ; p_{\mathbf{h}} \right)$ is defined using Eqs. (\ref{eq:nonlinear}) of Appendix \ref{app:weight_normalization}. Combining these equations with the ones for $\mathbf{w}$, we find the stationarity conditions
\begin{align*}
   \bar{w}^\mu_i &= \sum_{\mu_* = 1}^{P^*} x^{* \mu_*}_i \sum_{y = 0}^C q^{* \mu_*}_y \sigma_\mu \left( \beta m^{\mu_*} \left( \mathbf{x}^*, \mathbf{w} \right) + \log \left[ \mathbf{p}_y \right] \right) \\
    \bar{p}^\gamma_y &= \sum_{\mu_* = 1}^{P^*} q^{* \mu_*}_y \sigma_\gamma \left( \beta m^{\mu_*} \left( \mathbf{x}^*, \mathbf{w} \right) + \log \left[ \mathbf{p}_y \right] \right) \\
    w^\mu_i &= \frac{\bar{w}^\mu_i}{\sqrt{\sum_{j = 1}^N \left[\bar{w}^\mu_j \right]^2}} \\
    p^\gamma_y &= \frac{\bar{p}^\gamma_y}{\zeta^\gamma_y \left( \mathbf{\bar{p}} ; p_{\mathbf{h}} \right)}
\end{align*}
for all $1 \leq \mu \leq P$ and $0 \leq \gamma \leq P$.

\section{Replicated partition function and free entropy}
\label{app:partition_function}
Suppose a student DAM model is trained using a dataset $\mathcal{D} = \left\{ \mathbf{x}^c, y^c \right\}_{c = 1}^M$ of $M$ i.i.d. examples $\mathbf{x}^c$ with labels $y^c$. By Bayes' theorem, the student weights $\mathbf{w}$ and $\mathbf{p}$ follow the distribution
\begin{align*}
    \Prob_\beta \left( \mathbf{w}, \mathbf{p} | \mathcal{D} \right) &= \mathcal{Z} \left( \mathcal{D} \right)^{-1} \Prob \left( \mathbf{w}, \mathbf{p} \right) \prod_{c = 1}^M \Prob_\beta \left( \mathbf{x}^c, y^c \big| \mathbf{w}, \mathbf{p} \right),
\end{align*}
where $\mathcal{Z} \left( \mathcal{D} \right) = \mathbb{E}_{\mathbf{w}, \mathbf{p}} \left[ \prod_{c = 1}^M \Prob_\beta \left( \mathbf{x}^c, y^c | \mathbf{w}, \mathbf{p} \right) \right]$. Assuming the examples are sampled from a teacher DAM with weights $\mathbf{w}^*$ and $\mathbf{p}^*$, the average replicated partition takes the form
\begin{align*}
    \left\langle \mathcal{Z}^L \right\rangle &= \sum_{\left\{ y^c \right\}_{c = 1}^M} \int
    \left[ \prod_{c = 1}^M d \mathbf{x}^c \right] \ \mathbb{E}_{\mathbf{w}^*, \mathbf{p}^*} \left[ \prod_{c = 1}^M \Prob_\beta \left( \mathbf{x}^c, y^c | \mathbf{w}^*, \mathbf{p}^* \right) \right] \mathcal{Z} \left( \mathcal{D} \right)^L \\
    &= \sum_{\left\{ y^c \right\}_{c = 1}^M} \int
    \left[ \prod_{c = 1}^M d \mathbf{x}^c \right] \ \mathbb{E}_{\mathbf{w}^*, \mathbf{p}^*} \left[ \prod_{c = 1}^M \Prob_\beta \left( \mathbf{x}^c, y^c | \mathbf{w}^*, \mathbf{p}^* \right) \right] \prod_{a = 1}^L \mathbb{E}_{\mathbf{w}^a, \mathbf{p}^a} \left[ \prod_{c = 1}^M \Prob_\beta \left( \mathbf{x}^c, y^c | \mathbf{w}^a, \mathbf{p}^a \right) \right] \\
    &= \mathbb{E}_{\mathbf{w}^*, \mathbf{w}} \mathbb{E}_{\mathbf{p}^*, \mathbf{p}} \left[ \prod_{c = 1}^M \sum_{y^c = 0}^C \int_{S^{N - 1}} d \mathbf{x}^c \Prob_\beta \left( \mathbf{x}^c, y^c | \mathbf{w}^*, \mathbf{p}^* \right) \prod_{a = 1}^L \Prob_\beta \left( \mathbf{x}^c, y^c | \mathbf{w}^a, \mathbf{p}^a \right) \right] \\
    &= \mathbb{E}_{\mathbf{w}^*, \mathbf{w}} \mathbb{E}_{\mathbf{p}^*, \mathbf{p}} \left[ \left( \sum_{y = 0}^C \int_{S^{N - 1}} d \mathbf{x} \Prob_\beta \left( \mathbf{x}, y | \mathbf{w}^*, \mathbf{p}^* \right) \prod_{a = 1}^L \Prob_\beta \left( \mathbf{x}, y | \mathbf{w}^a, \mathbf{p}^a \right) \right)^M \right],
\end{align*}
where we redefined $\mathbf{w} = \left\{ \mathbf{w}^a \right\}_{a = 1}^L$ and $\mathbf{p} = \left\{ \mathbf{p}^a \right\}_{a = 1}^L$. In order to simplify the argument of $\left( \cdot \right)^M$, it is convenient to write Eq. (\ref{eq:direct_distribution}) in the form
\begin{gather*}
    \label{eq:alt_direct_distribution}
    \Prob_\beta \left( \mathbf{x}, y | \mathbf{w}^a, \mathbf{p}^a \right) = \sum_{\gamma = 0}^P \Omega_N \left( \beta \left[ 1 - \delta_{\gamma 0} \right] \right)^{-1} p^{a \gamma}_y \exp \left( \beta \left[ 1 - \delta_{\gamma 0} \right] \sum_{i = 1}^N w^{a \gamma}_i x_i \right), \numberthis
\end{gather*}
where the value of $\mathbf{w}^{a 0}$ is arbitrary. Until the end of this Section, all sums over the hidden units will include the index $0$ unless explicitly indicated otherwise. Define $I_y \left( \mathbf{x} \right) = \Prob_\beta \left( \mathbf{x}, y | \mathbf{w}^*, \mathbf{p}^* \right) \prod_{a = 1}^L \Prob_\beta \left( \mathbf{x}, y | \mathbf{w}^a, \mathbf{p}^a \right)$, then
\begin{align*}
    I_y \left( \mathbf{x} \right) &= \left[ \sum_{\gamma_* = 0}^{P^*} \Omega_N \left( \beta^* \left[ 1 - \delta_{\gamma_* 0} \right] \right)^{-1} p^{* \gamma_*}_y \exp \left( \beta^* \left[ 1 - \delta_{\gamma_* 0} \right] \sum_{i = 1}^N w^{* \gamma_*}_i x_i \right) \right] \\
    &\quad \prod_{a = 1}^L \left[ \sum_{\gamma = 0}^P \Omega_N \left( \beta \left[ 1 - \delta_{\gamma 0} \right] \right)^{-1} p^{a \gamma}_y \exp \left( \beta \left[ 1 - \delta_{\gamma 0} \right] \sum_{i = 1}^N w^{a \gamma}_i x_i \right) \right] \\
    &= \sum_{\gamma_* \gamma_1 ... \gamma_L} \Omega_N \left( \beta^* \left[ 1 - \delta_{\gamma_* 0} \right] \right)^{-1} \left[ \prod_{a} \Omega_N \left( \beta \left[ 1 - \delta_{\gamma_a 0} \right] \right) \right]^{-1} \\
    &\quad p^{* \gamma_*}_y \left[ \prod_a p^{a \gamma_a}_y \right] \exp \left( \beta^* \left[ 1 - \delta_{\gamma_* 0} \right] \sum_{i} w^{* \gamma_*}_i x_i + \beta \sum_{a} \left[ 1 - \delta_{\gamma_a 0} \right] \sum_{i} w^{a \gamma_a}_i x_i \right).
\end{align*}
We will now evaluate the integral $\int_{S^{N - 1}} d \mathbf{x} \ I_y \left( \mathbf{x} \right)$. Using Eq. (\ref{eq:thermodynamic_approx_to_vmf}) of Appendix \ref{app:vmf_integration} with $\rho = \rho_{\gamma_* \gamma} := \frac{1}{N} \sqrt{\sum_i \left[ \beta^{*} \left[ 1 - \delta_{\gamma_* 0} \right] w^{* \gamma_*}_i + \beta \sum_{a} \left[ 1 - \delta_{\gamma_a 0} \right] w^{a \gamma_a}_i \right]^2}$, we get
\begin{align*}
    &\int_{S^{N - 1}} d \mathbf{x} \exp \left( \beta^* \left[ 1 - \delta_{\gamma_* 0} \right] \sum_{i} w^{* \gamma_*}_i x_i + \beta \sum_{a} \left[ 1 - \delta_{\gamma_a 0} \right] \sum_{i} w^{a \gamma_a}_i x_i \right) \\
    &\approx \Omega_N \left( 0 \right) \left( 1 + \left( 2 \rho_{\gamma_* \gamma} \right)^2 \right)^{-1/4} \exp \left[ \left( \frac{N}{2} - 1 \right) \eta \left( 2 \rho_{\gamma_* \gamma} \right) \right],
\end{align*}
in the limit of large $N$. Since $\sqrt{\sum_i \left( w^{* \gamma_*}_i \right)^2} = 1$, the square of $\rho_{\gamma_* \gamma}$ simplifies to
\begin{align*}
    \left( \rho_{\gamma_* \gamma} \right)^2 &= \left( \frac{1}{N} \sqrt{\sum_i \left[ \beta^{*} \left[ 1 - \delta_{\gamma_* 0} \right] w^{* \gamma_*}_i + \beta \sum_{a} \left[ 1 - \delta_{\gamma_a 0} \right] w^{a \gamma_a}_i \right]^2} \right)^2 \\
    &= \upsilon^2 \left[ 1 - \delta_{\gamma_* 0} \right] + 2 \frac{\beta}{N} \upsilon \left[ 1 - \delta_{\gamma_* 0} \right] \sum_a \left[ 1 - \delta_{\gamma_a 0} \right] \sum_i w^{* \gamma_*}_i w^{a \gamma_a}_i \\
    &\quad+ \frac{\beta^2}{N^2} \sum_{a, b} \left[ 1 - \delta_{\gamma_a 0} \right] \left[ 1 - \delta_{\gamma_b 0} \right] \sum_i w^{a \gamma_a}_i w^{b \gamma_b}_i,
\end{align*}
where $\upsilon = \frac{\beta^*}{N}$.  To leading order in $\frac{\beta}{N}$ we find
\begin{align*}
    \left( 1 + \left( 2 \rho_{\gamma_* \gamma} \right)^2 \right)^{-1/4} &\approx \left( 1 + \left( 2 \upsilon \left[ 1 - \delta_{\gamma_* 0} \right] \right)^2 \right)^{-1/4} \\
    \text{and} \quad \left( \frac{N}{2} - 1 \right) \eta \left( 2 \rho_{\gamma_* \gamma_a} \right) &\approx \left( \frac{N}{2} - 1 \right) \eta \left( 2 \upsilon \left[ 1 - \delta_{\gamma_* 0} \right] \right) + \beta_{\text{eff}} \left[ 1 - \delta_{\gamma_* 0} \right] \sum_a \left[ 1 - \delta_{\gamma_a 0} \right] \sum_i w^{* \gamma_*}_i w^{a \gamma_a}_i \\
    &\quad+ \frac{\left[ \beta \xi_{\gamma_*} \right]^2}{2 N} \sum_{a, b} \left[ 1 - \delta_{\gamma_a 0} \right] \left[ 1 - \delta_{\gamma_b 0} \right] \sum_i w^{a \gamma_a}_i w^{b \gamma_b}_i,
\end{align*}
where $\beta_{\text{eff}} = \frac{2 \upsilon}{\sqrt{\left[ 2 \upsilon \right]^2 + 1} + 1} \beta$ and $\xi_{\gamma_*} = \delta_{\gamma_* 0} + \sqrt{\frac{2}{\sqrt{\left[ 2 \upsilon \right]^2 + 1} + 1}} \left[ 1 - \delta_{\gamma_* 0} \right]$. Assuming that $\upsilon \gg 1/N$ and $\beta \ll N$, we drop the last term. Finally, we use Eq. (\ref{eq:thermodynamic_approx_to_vmf}) backwards with $\rho = \upsilon \left[ 1 - \delta_{\gamma_* 0} \right]$ and obtain 
\begin{align*}
    &\Omega_N \left( \beta^* \left[ 1 - \delta_{\gamma_* 0} \right] \right)^{-1} \int_{S^{N - 1}} d \mathbf{x} \exp \left( \beta^* \left[ 1 - \delta_{\gamma_* 0} \right] \sum_{i} w^{* \gamma_*}_i x_i + \beta \sum_{a} \left[ 1 - \delta_{\gamma_a 0} \right] \sum_{i} w^{a \gamma_a}_i x_i \right) \\
    &\approx \exp \left( \beta_{\text{eff}} \left[ 1 - \delta_{\gamma_* 0} \right] \sum_a \left[ 1 - \delta_{\gamma_a 0} \right] \sum_i w^{* \gamma_*}_i w^{a \gamma_a}_i \right),
\end{align*}
from which we conclude that
\begin{align*}
    \sum_{y = 0}^C \int_{S^{N - 1}} d \mathbf{x} \ I_y \left( \mathbf{x} \right) &\approx \sum_{\gamma_* \gamma_1 ... \gamma_L} \sum_{y} p^{* \gamma_*}_y \left[ \prod_{a} p^{a \gamma_a}_y \right] \left[ \prod_{a} \Omega_N \left( \beta \left[ 1 - \delta_{\gamma_a 0} \right] \right) \right]^{-1} \\
    &\quad \exp \left( \beta_{\text{eff}} \left[ 1 - \delta_{\gamma_* 0} \right] \sum_a \left[ 1 - \delta_{\gamma_a 0} \right] \sum_i w^{* \gamma_*}_i w^{a \gamma_a}_i \right).
\end{align*}
We define $\alpha = M / N$, introduce the order parameter
\begin{equation}
\label{eq:alt_order_parameters}
    m^{a \gamma_* \nu} \hspace{-7pt} \quad \text{for} \ \sum_i w^{* \gamma_*}_i w^{a \nu}_i,
\end{equation}
and insert into $\left\langle \mathcal{Z}^L \right\rangle$ the identity operator
\begin{align*}
    1 &= \int_{\mathbb{R}} \prod_{\gamma_*, \nu ; a} d m^{a \gamma_* \nu}
    \delta \left( m^{a \gamma_* \nu} - \sum_i w^{* \gamma_*}_i w^{a \nu}_i \right) \\
    &= \int_{\mathbb{R}} \prod_{\gamma_*, \nu ; a} d m^{a \gamma_* \nu}
    \int_{i \mathbb{R}} \prod_{\gamma_*, \nu ; a} d \hat{m}^{a \gamma_* \nu} \exp \left\{ \frac{\beta_{\text{eff}} \alpha}{P^*} \sum_{\gamma_*, \nu ; a} \hat{m}^{a \gamma_* \nu} \left( \sum_i w^{* \gamma_*}_i w^{a \nu}_i - m^{a \gamma_* \nu} \right) \right\}
\end{align*}
so that we can rewrite it as
\begin{gather*}
\label{eq:replicated_partition_function}
\begin{aligned}
    \left\langle \mathcal{Z}^L \right\rangle &= \int \prod_{\gamma_*, \nu ; a} d \hat{m}^{a \gamma_* \nu} d m^{a \gamma_* \nu} \mathbb{E}_{\mathbf{w}^*, \mathbf{w}} \exp \left\{ N H_{S} \left( \mathbf{w}, \mathbf{w}^* ; \mathbf{\hat{m}} \right) \right\}
    \exp \left\{ -N H_Q \left( \mathbf{m}, \mathbf{\hat{m}} \right) \right\} \\
    &\quad \mathbb{E}_{\mathbf{p}^*, \mathbf{p}} \exp \left\{ \alpha N \log \left[ \sum_{\gamma_* \gamma_1 ... \gamma_L} \sum_{y} p^{* \gamma_*}_y \left[ \prod_{a} p^{a \gamma_a}_y \right] \exp \left\{ H_E \left( \gamma, \gamma_* ; \mathbf{m} \right) \right\} \right] \right\},
\end{aligned} \numberthis \\[12pt]
\begin{aligned}
    \text{where} \quad H_{S} \left( \mathbf{w}, \mathbf{w}^* ; \mathbf{\hat{m}} \right) &= \frac{\beta_{\text{eff}} \alpha}{P^*} \sum_{\gamma_*, \nu ; a} \hat{m}^{a \gamma_* \nu} \sum_i w^{* \gamma_*}_i w^{a \nu}_i \\
    H_Q \left( \mathbf{m}, \mathbf{\hat{m}} \right) &= \frac{\beta_{\text{eff}} \alpha}{P^*} \sum_{\gamma_*, \nu ; a} \hat{m}^{a \gamma_* \nu} m^{a \gamma_* \nu}, \\
    H_E \left( \gamma, \gamma_* ; \mathbf{m} \right) &= -\sum_a \log \left[ \Omega_N \left( \beta \left[ 1 - \delta_{\gamma_a 0} \right] \right) \right] + \beta_{\text{eff}} \left[ 1 - \delta_{\gamma_* 0} \right] \sum_a \left[ 1 - \delta_{\gamma_a 0} \right] m^{a \gamma_* \gamma_a}.
\end{aligned}
\end{gather*}
Using Eq. (\ref{eq:thermodynamic_approx_to_vmf}) of Appendix \ref{app:vmf_integration}, the exponential of $H_{S}$ integrates to
\begin{align*}
    &\mathbb{E}_{\mathbf{w}} \exp \left\{ N H_{S} \left( \mathbf{w}, \mathbf{w}^* ; \mathbf{\hat{m}} \right) \right\} \\
    &= \Omega_N \left( 0 \right)^{-P L} \int_{S^{N - 1}} d \mathbf{w} \exp \left( \frac{\beta_{\text{eff}} \alpha}{P^*} \sum_{\gamma_*, \nu ; a} \hat{m}^{a \gamma_* \nu} \sum_i w^{* \gamma_*}_i w^{a \nu}_i \right) \\
    &\approx \prod_{\nu ; a} \exp \left[ -\frac{1}{4} \log \left( 1 + \left[ \frac{2 \beta_{\text{eff}} \alpha}{P^*} \right]^2 \sum_i \left[ \sum_{\gamma_*} \hat{m}^{a \gamma_* \nu} w^{* \gamma_*}_i \right]^2 \right) \right] \\
    &\quad \exp \left[ \left( \frac{N}{2} - 1 \right) \eta \left( \frac{2 \beta_{\text{eff}} \alpha}{P^*} \sqrt{\sum_i \left[ \sum_{\gamma_*} \hat{m}^{a \gamma_* \nu} w^{* \gamma_*}_i \right]^2} \right) \right].
\end{align*}
In order to continue the calculations, we make the replica-symmetric ansatz
\begin{gather*}
    m^{a \gamma_* \nu} = m^{\gamma_* \nu} \hspace{-7pt} \quad \text{and} \ \hat{m}^{a \gamma_* \nu} = \hat{m}^{\gamma_* \nu} \ \text{for all} \ a ; \gamma_*, \nu \\
    p^{a \nu}_y = p^{\nu}_y \ \text{for all} \ a ; \nu
\end{gather*}
so that we can use the replica trick to simplify
\begin{align*}
    &\frac{1}{L} \log \left[ \sum_{\gamma_* \gamma_1 ... \gamma_L} \sum_{y} p^{* \gamma_*}_y \left[ \prod_{a} p^{a \gamma_a}_y \right] \exp \left\{ H_E \left( \gamma, \gamma_* ; \mathbf{m} \right) \right\} \right] \\
    &= \frac{1}{L} \log \left[ \sum_{\gamma_*} \sum_{y} p^{* \gamma_*}_y \sum_{\gamma_1 ... \gamma_L} \left[ \prod_{a} p^{a \gamma_a}_y \right] \exp \left\{ H_E \left( \gamma, \gamma_* ; \mathbf{m} \right) \right\} \right] \\
    &\approx \sum_{\gamma_*} \sum_{y} p^{* \gamma_*}_y \log \Bigg[ \sum_{\gamma} p^\gamma_y \Omega_N \left( \beta \left[ 1 - \delta_{\gamma 0} \right] \right)^{-1} \exp \left( \beta_{\text{eff}} \left[ 1 - \delta_{\gamma_* 0} \right] \left[ 1 - \delta_{\gamma 0} \right] m^{\gamma_* \gamma} \right) \Bigg],
\end{align*}
as we take $L$ to zero. Similarly, we get
\begin{align*}
    \frac{1}{L} \log \left[ \mathbb{E}_{\mathbf{w}} \exp \left\{ N H_{S} \left( \mathbf{w}, \mathbf{w}^* ; \mathbf{\hat{m}} \right) \right\} \right] &= -\frac{1}{4} \log \left( 1 + \left[ \frac{2 \beta_{\text{eff}} \alpha}{P^*} \right]^2 \sum_i \left[ \sum_{\gamma_*} \hat{m}^{\gamma_* \nu} w^{* \gamma_*}_i \right]^2 \right) \\
    &\quad+ \left( \frac{N}{2} - 1 \right) \eta \left( \frac{2 \beta_{\text{eff}} \alpha}{P^*} \sqrt{\sum_i \left[ \sum_{\gamma_*} \hat{m}^{\gamma_* \nu} w^{* \gamma_*}_i \right]^2} \right).
\end{align*}
From now on, we assume for simplicity that $\sum_y p^{* 0}_y = p_{\mathbf{h}}^* \left( 0 \right) = 0$ and $\sum_y p^{* \mu_*}_y = p_{\mathbf{h}}^* \left( \mu_* \right) = 1/P^*$ for all $\mu_* > 0$. Defining $g^{* \mu_*}_y = P^* p^{* \mu_*}_y = \Prob_\beta \left( y | \mu_*, \mathbf{J} \right)$ for all $\mu_* > 0$, $\mathbf{g}^* = \left\{ g^{* \gamma_*}_y \right\}_{0 \leq y \leq C}^{1 \leq \mu_* \leq P^*}$ and $\varrho = \frac{\alpha}{P^*} = \frac{M}{P^* N}$, the free entropy $f$ then takes the form
\begin{align*}
    \label{eq:var_free_entropy}
    &f \approx \Extr_{\mathbf{m}, \mathbf{\hat{m}}, \mathbf{p}} \left\{ f \left( \mathbf{m}, \mathbf{\hat{m}}, \mathbf{p} \right) \right\} \quad \text{such that} \quad \sum_{\gamma = 0}^P p^\gamma_y = p_{\mathbf{q}} \left( y \right) \quad \text{and} \quad \sum_{y = 0}^C p^\gamma_y = p_{\mathbf{h}} \left( \gamma \right) \numberthis \\
    &\text{with} \quad f \left( \mathbf{m}, \mathbf{\hat{m}}, \mathbf{p} \right) = -\beta_{\text{eff}} \varrho \sum_{\gamma_*, \gamma = 0}^{P^*, P} \hat{m}^{\gamma_* \gamma} m^{\gamma_* \gamma} + \frac{1}{2} \mathbb{E}_{\mathbf{w}^*} \left[ \sum_{\gamma = 0}^P \eta \left( 2\beta_{\text{eff}} \varrho \sqrt{\sum_{i = 1}^N \left[ \sum_{\gamma_* = 0}^{P^*} \hat{m}^{\gamma_* \gamma} w^{* \gamma_*}_i \right]^2} \right) \right] \\
    &\quad+ \varrho \sum_{\mu_* = 1}^{P^*} \sum_{y = 0}^C \mathbb{E}_{\mathbf{g}^*} \left[ g^{* \mu_*}_y \right] \log \left[ \sum_{\gamma = 0}^P p^\gamma_y \Omega_N \left( \beta \left[ 1 - \delta_{\gamma 0} \right] \right)^{-1} \exp \left( \beta_{\text{eff}} \left[ 1 - \delta_{\gamma 0} \right] m^{\mu_* \gamma} \right) \right]
\end{align*}
where we approximated the expectation over $\mathbf{p}$ as a (constrained) extremization problem using Laplace's method. By inspection, the order parameters $m^{0 \gamma}$ and $\hat{m}^{0 \gamma}$ always vanish, so we ignore them in the incoming derivation of the saddle-point equations.

\section{Saddle-point equations}
\label{app:saddle-point}
In this Appendix, we adopt the convention $1 \leq \mu_* \leq P^*$, $0 \leq \gamma \leq P$ and $1 \leq \mu \leq P$. By the Lagrange multiplier theorem, any set of class weights $\mathbf{p}$ that extremizes Eq. (\ref{eq:var_free_entropy}) must solve the extremization problem
\begin{align*}
    \Extr_{\mathbf{p}, \omega, \lambda} \left\{ f \left( \mathbf{m}, \mathbf{\hat{m}}, \mathbf{p} \right) + \sum_{y = 0}^C \lambda_y \left( \sum_{\gamma = 0}^P p^\gamma_y - p_{\mathbf{q}} \left( y \right) \right) + \sum_{\gamma = 0}^P \omega^{\gamma} \left( \sum_{y = 0}^C p^\gamma_y - p_{\mathbf{h}} \left( \gamma \right) \right) \right\}.
\end{align*}
In Appendix \ref{app:weight_normalization}, we show that its solution is
\begin{align*}
    \bar{p}^\gamma_y &= \sum_{\mu_* = 1}^{P^*} \mathbb{E}_{\mathbf{g}^*} \left[ g^{* \mu_*}_y \right] \sigma_\gamma \left( \beta_{\text{eff}} \left[ 1 - \delta_{\gamma 0} \right] m^{\mu_*} - \log \left[ \Omega_N \left( \beta \left[ 1 - \delta_{\gamma 0} \right] \right) \right] + \log \left[ \mathbf{p}_y \right] \right) \\
    p^\gamma_y &= \frac{\bar{p}^\gamma_y}{\zeta^\gamma_y \left( \mathbf{\bar{p}} ; p_{\mathbf{h}} \right)},
\end{align*}
for all $0 \leq \gamma \leq P$, where $\zeta^\gamma_y \left( \mathbf{\bar{p}} ; p_{\mathbf{h}} \right)$ is defined using Eqs. (\ref{eq:nonlinear}) of Appendix \ref{app:weight_normalization} and $\sigma_{\gamma} \left( x^{\mu_*} \right) = \frac{\exp \left( x^{\mu_* \gamma} \right)}{\sum_{\kappa = 0}^P \exp \left( x^{\mu_* \kappa} \right)}$ is the softmax function. We extremize Eq. (\ref{eq:var_free_entropy}) with respect to the remaining parameters by solving for the gradient equal to zero. $\partial_{m^{\mu_* \mu}} f \left( \mathbf{m}, \mathbf{\hat{m}}, \mathbf{p} \right) = 0$ immediately yields
\begin{align*}
    \label{eq:prelim_saddle_point}
    \hat{m}^{\mu_* \gamma} &= \left[ 1 - \delta_{\gamma 0} \right] \sum_{y = 0}^C \mathbb{E}_{\mathbf{g}^*} \left[ g^{* \mu_*}_y \right] \frac{p^\gamma_y \Omega_N \left( \beta \left[ 1 - \delta_{\gamma 0} \right] \right)^{-1} \exp \left( \beta_{\text{eff}} \left[ 1 - \delta_{\gamma 0} \right] m^{\mu_* \gamma} \right)}{\sum_{\kappa = 0}^P p^\kappa_y \Omega_N \left( \beta \left[ 1 - \delta_{\kappa 0} \right] \right)^{-1} \exp \left( \beta_{\text{eff}} \left[ 1 - \delta_{\kappa 0} \right] m^{\mu_* \kappa} \right)}. \numberthis
\end{align*}
On the other hand, $\partial_{\hat{m}^{\mu_* \gamma}} f \left( \mathbf{m}, \mathbf{\hat{m}}, \mathbf{p} \right) = 0$ gives an equation that depends on the choice of prior for the teacher memories $\mathbf{w}^{* \mu_*}$. We first investigate the case where the teacher memories $\mathbf{w}^{* \mu_*}$ and the class weights $\mathbf{g}^*$ are distributed uniformly at random over the sets to which they are constrained (see Section \ref{sec:teacher-student}). In this scenario, we have $\mathbb{E}_{\mathbf{g}^{*}} \left[ g^{* \mu_*}_y \right] = p_{\mathbf{q}}^* \left( y \right)$. Moreover, assuming that $P^* \ll N$, random vectors on the unit sphere are orthonormal with high probability, which means that
\begin{align*}
    \sum_{i = 1}^N \left[ \sum_{\mu_* = 1}^{P^*} \hat{m}^{\mu_* \gamma} w^{* \mu_*}_i \right]^2 &= \sum_{i = 1}^N \sum_{\mu_*, \nu_* = 1}^{P^*} \hat{m}^{\mu_* \gamma} \hat{m}^{\nu_* \gamma} w^{* \mu_*}_i w^{* \nu_*}_i \\
    &= \sum_{\mu_* = 1}^{P^*} \left[ \hat{m}^{\mu_* \gamma} \right]^2 + \sum_{\mu_* \neq \nu_*} \hat{m}^{\mu_* \gamma} \hat{m}^{\nu_* \gamma} \sum_{i = 1}^N w^{* \mu_*}_i w^{* \nu_*}_i \\
    &\approx \sum_{\mu_* = 1}^{P^*} \left[ \hat{m}^{\mu_* \gamma} \right]^2.
\end{align*}
Therefore, $\partial_{\hat{m}^{\mu_* \gamma}} f \left( \mathbf{m}, \mathbf{\hat{m}}, \mathbf{p} \right) = 0$ gives
\begin{align*}
    0 &= \partial_{\hat{m}^{\mu_* \gamma}} f \left( \mathbf{m}, \mathbf{\hat{m}}, \mathbf{p} \right) \\
    0 &= -\beta_{\text{eff}} \varrho m^{\mu_* \gamma} + \frac{1}{2} \partial_{\hat{m}^{\mu_* \gamma}} \sum_{\kappa = 0}^P \eta \left( 2 \beta_{\text{eff}} \varrho \sqrt{\sum_{\nu_* = 1}^{P^*} \left[ \hat{m}^{\nu_* \kappa} \right]^2} \right) \\
    m^{\mu_* \gamma} &= \varsigma \left( 2 \beta_{\text{eff}} \varrho \sqrt{\sum_{\nu_* = 1}^{P^*} \left[ \hat{m}^{\nu_* \gamma} \right]^2} \right) \frac{\hat{m}^{\mu_* \gamma}}{\sqrt{\sum_{\nu_* = 1}^{P^*} \left[ \hat{m}^{\nu_* \gamma} \right]^2}},
\end{align*}
where $\varsigma \left( x \right) = \frac{x}{\sqrt{x^2 + 1} + 1}$. $\hat{m}^{\mu_* \gamma}$ vanishes (see Eqs. \ref{eq:prelim_saddle_point}) and $m^{\mu_* \gamma}$ is arbitrary (see Eqs. \ref{eq:alt_direct_distribution} and \ref{eq:alt_order_parameters}) when $\gamma = 0$. Therefore, we may update only $\hat{m}^{\mu_* \mu}$ and $m^{\mu_* \mu}$ with $1 \leq \mu \leq P$ when solving for $\nabla f = 0$ by fixed-point iteration. Defining $m^{\mu_* 0} = \frac{1}{\beta_{\text{eff}}} \log \left[ \Omega_N \left( \beta \right) / \Omega_N \left( 0 \right) \right]$, we then obtain the saddle-point equations
\begin{align*}
    \hat{m}^{\mu_* \mu} &= \sum_{y = 0}^C p_{\mathbf{q}}^* \left( y \right) \sigma_\mu \left( \beta_{\text{eff}} m^{\mu_*} + \log \left[ \mathbf{p}_y \right] \right) \\
    m^{\mu_* \mu} &= \varsigma \left( 2 \beta_{\text{eff}} \varrho \sqrt{\sum_{\nu_* = 1}^{P^*} \left[ \hat{m}^{\nu_* \mu} \right]^2} \right) \frac{\hat{m}^{\mu_* \mu}}{\sqrt{\sum_{\nu_* = 1}^{P^*} \left[ \hat{m}^{\nu_* \mu} \right]^2}},
\end{align*}
for all $1 \leq \mu \leq P$, where we simplified the argument of the softmax function $\sigma_{\mu} \left( x^{\mu_*} \right) = \frac{\exp \left( x^{\mu_* \mu} \right)}{\sum_{\kappa = 0}^P \exp \left( x^{\mu_* \kappa} \right)}$ using $m^{\mu_* 0} = \frac{1}{\beta_{\text{eff}}} \log \left[ \Omega_N \left( \beta \right) / \Omega_N \left( 0 \right) \right]$. Putting the equations for $m^{\mu_* \mu}$ and $p^\gamma_y$ together, we find
\begin{align*}
    \hat{m}^{\mu_* \mu} &= \sum_{y = 0}^C p_{\mathbf{q}}^* \left( y \right) \sigma_\mu \left( \beta_{\text{eff}} m^{\mu_*} + \log \left[ \mathbf{p}_y \right] \right) \\
    \bar{p}^\gamma_y &= p_{\mathbf{q}}^* \left( y \right) \sum_{\mu = 1}^{P^*} \sigma_\gamma \left( \beta_{\text{eff}} m^{\mu_*} + \log \left[ \mathbf{p}_y \right] \right) \\
    m^{\mu_* \mu} &= \varsigma \left( 2 \beta_{\text{eff}} \varrho \sqrt{\sum_{\nu_* = 1}^{P^*} \left[ \hat{m}^{\nu_* \mu} \right]^2} \right) \frac{\hat{m}^{\mu_* \mu}}{\sqrt{\sum_{\nu_* = 1}^{P^*} \left[ \hat{m}^{\nu_* \mu} \right]^2}} \\
    p^\gamma_y &= \frac{\bar{p}^\gamma_y}{\zeta^\gamma_y \left( \mathbf{\bar{p}} ; p_{\mathbf{h}} \right)}
\end{align*}
for all $1 \leq \mu \leq P$ and $0 \leq \gamma \leq P$. If we instead clamp $\mathbf{w}^{* \mu_*}$ and $\mathbf{g}^{* \mu_*}$ to some fixed patterns $\mathbf{x}^{* \mu_*}$ and their soft labels $\mathbf{q}^{* \mu_*}$, respectively, then the solutions of Eq. (\ref{eq:var_free_entropy}) take the form
\begin{align*}
    \hat{m}^{\mu_* \mu} &= \sum_{y = 0}^C q^{* \mu_*}_y \sigma_\mu \left( \beta_{\text{eff}} m^{\mu_*} + \log \left[ \mathbf{p}_y \right] \right) \\
    \bar{p}^\gamma_y &= \sum_{\mu_* = 1}^{P^*} q^{* \mu_*}_y \sigma_\gamma \left( \beta_{\text{eff}} m^{\mu_*} + \log \left[ \mathbf{p}_y \right] \right) \\
    m^{\mu_* \mu} &= \varsigma \left( 2 \beta_{\text{eff}} \varrho \sqrt{\sum_{i = 1}^N \left[ \sum_{\nu_* = 1}^{P^*} \hat{m}^{\nu_* \mu} x^{* \nu_*}_i \right]^2} \right) \frac{\sum_{i = 1}^N x^{* \nu_*}_i \sum_{\nu_* = 1}^{P^*} \hat{m}^{\nu_* \mu} x^{* \nu_*}_i}{\sqrt{\sum_{i = 1}^N \left[ \sum_{\nu_* = 1}^{P^*} \hat{m}^{\nu_* \mu} x^{* \nu_*}_i \right]^2}} \\
    p^\gamma_y &= \frac{\bar{p}^\gamma_y}{\zeta^\gamma_y \left( \mathbf{\bar{p}} ; p_{\mathbf{h}} \right)}.
\end{align*}
Defining $\bar{x}^\mu_i = \sum_{\mu_*} \hat{m}^{\mu_* \mu} x^{* \mu_*}_i$, we thus find
\begin{align*}
    \bar{x}^\mu_i &= \sum_{\mu_* = 1}^{P^*} x^{* \mu_*}_i \sum_{y = 0}^C q^{* \mu_*}_y \sigma_{\mu} \left( \beta_{\text{eff}} m^{\mu_*} + \log \left[ \mathbf{p}_y \right] \right) \\
    \bar{p}^\gamma_y &= \sum_{\mu_* = 1}^{P^*} q^{* \mu_*}_y \sigma_\gamma \left( \beta_{\text{eff}} m^{\mu_*} + \log \left[ \mathbf{p}_y \right] \right) \\
    m^{\mu_* \mu} &= \varsigma \left( 2 \beta_{\text{eff}} \varrho \sqrt{\sum_{i = 1}^N \left[ \bar{x}^\mu_i \right]^2} \right) \frac{\sum_{i = 1}^N x^{* \mu_*}_i \bar{x}^\mu_i}{\sqrt{\sum_{i = 1}^N \left[ \bar{x}^\mu_i \right]^2}} \\
    p^\gamma_y &= \frac{\bar{p}^\gamma_y}{\zeta^\gamma_y \left( \mathbf{\bar{p}} ; p_{\mathbf{h}} \right)},
\end{align*}
for all $1 \leq \mu \leq P$ and $0 \leq \gamma \leq P$.

\section{Saddle-point hierarchy}
\label{app:fixed-point}
Suppose that the set of parameters $\bar{x}^{\text{fixed}, \mu}_i$, $\bar{p}^{\text{fixed}, \gamma}_y$, $m^{\text{fixed}, \mu_* \gamma}$, $p^{\text{fixed}, \gamma}_y$ with hidden unit prior $p_{\mathbf{h}}^{\text{given}} \left( \gamma \right)$ is a fixed point of Eqs. (\ref{eq:saddle-point}) with $P$ hidden units. Substitute into the same saddle-point equations with $P + R \in \left\{ P, ..., 2 P \right\}$ hidden units the duplicated order parameters
\begin{align*}
    \bar{x}^{\text{dupli}, \mu}_i
    &= \begin{cases}
        \bar{x}^{\text{fixed}, \mu}_i &\quad 0 < \mu \leq P \\
        \bar{x}^{\text{fixed}, \mu - P}_i &\quad P < \mu \leq P + R \\
    \end{cases} \\
    \bar{p}^{\text{dupli}, \gamma}_y
    &= \begin{cases}
        \bar{p}^{ \text{fixed}, 0}_y &\quad \gamma = 0 \\
        \frac{1}{2} \bar{p}^{ \text{fixed}, \gamma}_y &\quad 0 < \gamma \leq R \\
        \bar{p}^{\text{fixed}, \gamma}_y &\quad R < \gamma \leq P \\
        \frac{1}{2} \bar{p}^{ \text{fixed}, \gamma - P}_y &\quad P < \gamma \leq P + R \\
    \end{cases} \\
    m^{\text{dupli}, \mu_* \gamma}
    &= \begin{cases}
        m^{\text{fixed}, \mu_* 0} &\quad \gamma = 0 \\
        m^{\text{fixed}, \mu_* \gamma} &\quad 0 < \gamma \leq P \\
        m^{\text{fixed}, \mu_*, \gamma - P} &\quad P < \gamma \leq P + R \\
    \end{cases} \\
    p^{\text{dupli}, \gamma}_y
    &= \begin{cases}
        p^{\text{fixed}, 0}_y &\quad \gamma = 0 \\
        \frac{1}{2} p^{\text{fixed}, \gamma}_y &\quad 0 < \gamma \leq R \\
        p^{\text{fixed}, \gamma}_y &\quad R < \gamma \leq P \\
        \frac{1}{2} p^{\text{fixed}, \gamma - P}_y &\quad P < \gamma \leq P + R \\
    \end{cases}
\end{align*}
\begin{align*}
    \text{along with} \quad p_{\mathbf{h}} \left( \gamma \right)
    &= \begin{cases}
        p_{\mathbf{h}}^{\text{given}} \left( 0 \right) &\quad \gamma = 0 \\
        \frac{1}{2} p_{\mathbf{h}}^{\text{given}} \left( \gamma \right) &\quad 0 < \gamma \leq R \\
        p_{\mathbf{h}}^{\text{given}} \left( \gamma \right) &\quad R < \gamma \leq P \\
        \frac{1}{2} p_{\mathbf{h}}^{\text{given}} \left( \gamma - P \right) &\quad P < \gamma \leq P + R,
    \end{cases}
\end{align*}
where the hidden units $\gamma \in \left\{ P + 1, ..., P + R \right\}$ and their corresponding order parameters are duplicates, or copies, of $\gamma \in \left\{ 1, ..., R \right\}$. $\gamma = 0$ can also be duplicated, but the result is less interesting. By definition (see Eqs. \ref{eq:nonlinear}), $\zeta^\gamma_y \left( \mathbf{\bar{p}}^{\text{dupli}} ; p_{\mathbf{h}} \right) = \lambda_y \left( \mathbf{\bar{p}}^{\text{dupli}} ; p_{\mathbf{h}} \right) + \omega^\gamma \left( \mathbf{\bar{p}}^{\text{dupli}} ; p_{\mathbf{h}} \right)$, where $\lambda_y \left( \mathbf{\bar{p}}^{\text{dupli}} ; p_{\mathbf{h}} \right)$ and $\omega^\gamma \left( \mathbf{\bar{p}}^{\text{dupli}} ; p_{\mathbf{h}} \right)$ are the $\lambda_y$ and $\omega^\gamma$ solving
\begin{align*}
    \omega^\gamma &= \frac{1}{p_{\mathbf{h}} \left( \gamma \right)} \sum_{y = 0}^C \frac{\omega^\gamma}{\lambda_y + \omega^\gamma} \bar{p}^{\text{dupli}, \gamma}_y \\
    &= \frac{1}{p_{\mathbf{h}}^{\text{given}} \left( \gamma \right)} \sum_{y = 0}^C \frac{\omega^\gamma}{\lambda_y + \omega^\gamma} \bar{p}^{\text{fixed}, \gamma}_y \\
    \lambda_y &= \frac{1}{p_{\mathbf{q}} \left( y \right)} \sum_{\gamma = 0}^{P + R} \frac{\lambda_y}{\lambda_y + \omega^\gamma} \bar{p}^{ \text{dupli}, \gamma}_y \\
    &= \frac{1}{p_{\mathbf{q}} \left( y \right)} \sum_{\gamma = 0}^{P} \frac{\lambda_y}{\lambda_y + \omega^\gamma} \bar{p}^{ \text{fixed}, \gamma}_y.
 \end{align*}
Therefore, $\zeta^\gamma_y \left( \mathbf{\bar{p}}^{ \text{dupli}} ; p_{\mathbf{h}} \right) = \zeta^\gamma_y \left( \mathbf{\bar{p}}^{ \text{fixed}} ; p_{\mathbf{h}}^{\text{given}} \right)$, and the saddle-point equations simplify to
\begin{align*}
    \label{eq:replicated_saddle-point}
    \bar{x}^{\text{fixed}, \mu}_i &= \frac{1}{2} \left( 1 + \mathbb{I} \left( \mu > R \right) \right) \sum_{\mu_* = 1}^{P^*} x^{* \mu_*}_i \sum_{y = 0}^C q^{* \mu_*}_y \sigma_{\mu} \left( \beta_{\text{eff}} m^{\text{fixed}, \mu_*} + \log \left[ \mathbf{p}^{\text{fixed}}_y \right] \right) \\
    \bar{p}^{ \text{fixed}, \gamma}_y &= \sum_{\mu_* = 1}^{P^*} q^{* \mu_*}_y \sigma_{\gamma} \left( \beta_{\text{eff}} m^{\text{fixed}, \mu_*} + \log \left[ \mathbf{p}^{\text{fixed}}_y \right] \right) \\
    m^{\text{fixed}, \mu_* \mu} &= \varsigma \left( 2 \beta_{\text{eff}} \varrho \sqrt{\sum_{i = 1}^N \left[ \mathbf{\bar{x}}^{* \text{fixed}, \mu}_i \right]^2} \right) \frac{\sum_{i = 1}^N x^{* \mu_*}_i \mathbf{\bar{x}}^{* \text{fixed}, \mu}_i}{\sqrt{\sum_{i = 1}^N \left[ \mathbf{\bar{x}}^{* \text{fixed}, \mu}_i \right]^2}} \numberthis \\
    p^{\text{fixed}, \gamma}_y &= \frac{\bar{p}^{ \text{fixed}, \gamma}_y}{\zeta^\gamma_y \left( \mathbf{\bar{p}}^{ \text{fixed}} ; p_{\mathbf{h}}^{\text{given}} \right)},
\end{align*}
where $\mathbb{I} \left( \mu > R \right)$ is the indicator function equal to $1$ when $\mu > R$ and $0$ otherwise. Assume that $\varrho \rightarrow \infty$ so that $\varsigma \left( 2 \beta_{\text{eff}} \varrho \sqrt{\sum_{i = 1}^N \left[ \bar{x}^{\mu}_i \right]^2} \right) \rightarrow \mathbb{I} \left( \sqrt{\sum_{i = 1}^N \left[ \bar{x}^{\mu}_i \right]^2} > 0 \right)$, then the saddle-point equations are the same no matter how the prefactor of $\frac{1}{2} \left( 1 + \mathbb{I} \left( \mu > R \right) \right)$ affects the norm of $\bar{x}^{\mu}_i$, and Eqs. (\ref{eq:fixed_point}) are a fixed point of the saddle-point equations with $P + R$ hidden units. In particular, Eq. (\ref{eq:fixed_point}) is a stationary point of the loss (Eq. \ref{eq:loss}) when $\beta_{\text{eff}} = \beta$.

For the rest of this Appendix, all sums are understood as having the same bounds as Eqs. (\ref{eq:replicated_saddle-point}) and (\ref{eq:saddle-point}). The stability of any fixed point of the form $\mathbf{x} = \mathbf{F} (\mathbf{x})$, such as those of Eqs. (\ref{eq:saddle-point}), can be evaluated using the Jacobian matrix $\mathbf{J}$ of $\mathbf{F}$. If all eigenvalues $\lambda$ of the Jacobian satisfy $| \lambda | < 1$, then the fixed point is stable. Conversely, if the Jacobian has an eigenvalue $\lambda$ with $| \lambda | > 1$, then the fixed point is unstable. In particular, if the quadratic form $\mathbf{v}^T \mathbf{J} \mathbf{v}$ is larger than $1$ for some $\mathbf{v}$ with $\| \mathbf{v} \| = 1$, then the fixed point is unstable. For the rest of this Appendix, we evaluate the stability of Eqs. (\ref{eq:saddle-point}) with duplicated order parameters. Keeping $\mathbf{\bar{p}}$ and $\mathbf{p}$ fixed, the Jacobian of the saddle-point equation for $\mathbf{\bar{x}}$ is
\begin{align*}
    \partial_{\bar{x}^{\nu}_k} \bar{x}^{\mu}_j
    &= \sum_{\mu_*} x^{* \mu_*}_j \partial_{\bar{x}^{\nu}_k} m^{\mu_* \nu} \sum_{y} q^{* \mu_*}_y \partial_{m^{\mu_* \nu}} \sigma_{\mu} \left( \beta_{\text{eff}} m^{\mu_*} + \log \left[ \mathbf{p}_y \right] \right) \\
    &= \beta_{\text{eff}} \sum_{\mu_*} x^{* \mu_*}_j \partial_{\bar{x}^{\nu}_k} m^{\mu_* \nu} \sum_{y} q^{* \mu_*}_y \sigma_{\mu} \left( \beta_{\text{eff}} m^{\mu_*} + \log \left[ \mathbf{p}_y \right] \right) \left( \delta_{\mu \nu} - \sigma_{\nu} \left( \beta_{\text{eff}} m^{\mu_*} + \log \left[ \mathbf{p}_y \right] \right) \right) \\
    &= \beta_{\text{eff}} \sum_{\mu_*} x^{* \mu_*}_j \frac{1}{\sqrt{\sum_i \left[ \bar{x}^{\nu}_i \right]^2}} \left( x^{* \mu_*}_k - \left[ \sum_{i} x^{* \mu_*}_i \tilde{x}^{\nu}_i \right] \tilde{x}^{\nu}_k \right) \\
    &\quad \sum_{y} q^{* \mu_*}_y \sigma_{\mu} \left( \beta_{\text{eff}} m^{\mu_*} + \log \left[ \mathbf{p}_y \right] \right) \left( \delta_{\mu \nu} - \sigma_{\nu} \left( \beta_{\text{eff}} m^{\mu_*} + \log \left[ \mathbf{p}_y \right] \right) \right),
\end{align*}
where $\tilde{x}^\nu_j = \frac{\bar{x}^{\nu}_j}{\sqrt{\sum_i \left[ \bar{x}^{\nu}_i \right]^2}}$. Suppose that $\bar{x}^{\mu}_i$, $\bar{p}^\gamma_y$, $m^{\mu_* \gamma}$ and $p^{ \gamma}_y$ are the duplicated parameters of Eq. (\ref{eq:fixed_point}) with $0 < \mu, \nu \leq R$ or $P < \mu, \nu \leq P + R$, then
\begin{gather*}
    \begin{aligned}
    \partial_{\bar{x}^{\nu}_k} \bar{x}^{\mu}_j &= \beta_{\text{eff}} \sum_{\mu_*} x^{* \mu_*}_j \frac{1}{\sqrt{\sum_i \left[ \bar{x}^{\nu}_i \right]^2}} \left( x^{* \mu_*}_k - \left[ \sum_{i} x^{* \mu_*}_i \tilde{x}^{\nu}_i \right] \tilde{x}^{\nu}_k \right) \\
    &\quad \sum_{y} q^{* \mu_*}_y \frac{1}{2} \sigma_{\theta \left( \mu \right) } \left( \beta_{\text{eff}} m^{\text{fixed}, \mu_*} + \log \left[ \mathbf{p}^{\text{fixed}}_y \right] \right) \left( \delta_{\mu \nu} - \frac{1}{2} \sigma_{\theta \left( \nu \right)} \left( \beta_{\text{eff}} m^{\text{fixed}, \mu_*} + \log \left[ \mathbf{p}^{\text{fixed}}_y \right] \right) \right),
    \end{aligned} \\
    \text{where} \quad \theta \left( \mu \right) = \begin{cases}
        \mu &\quad 0 < \mu \leq R \\
        \mu - P &\quad 0 < \mu \leq P + R.
    \end{cases}
\end{gather*}
If $\beta_{\text{eff}}$ is relatively large (for instance of order $\sqrt{N}$), then the softmax $\sigma_{\theta \left( \mu \right)}$ splits the indices $\mu_*$ into a cover $\mathcal{S}$ of sets $\mathcal{S} \left( \mu \right)$ such that $\sigma_{\theta \left( \mu \right)} \left( \beta_{\text{eff}} m^{\text{fixed}, \mu_*} + \log \left[ \mathbf{p}_y \right] \right) \approx \mathbb{I} \left( \mu_* \in \mathcal{S} \left( \mu \right) \right)$. Therefore, we have
\begin{align*}
    \partial_{\bar{x}^{\nu}_k} \bar{x}^{\mu}_j &\approx \beta_{\text{eff}} \sum_{\mu_*} x^{* \mu_*}_j \frac{1}{\sqrt{\sum_i \left[ \bar{x}^{\nu}_i \right]^2}} \left( x^{* \mu_*}_k - \left[ \sum_{i} x^{* \mu_*}_i \tilde{x}^{\nu}_i \right] \tilde{x}^{\nu}_k \right) \\
    &\quad \frac{1}{2} \mathbb{I} \left( \mu_* \in \mathcal{S} \left( \mu \right) \right) \left( \delta_{\mu \nu} - \frac{1}{2} \mathbb{I} \left( \mu_* \in \mathcal{S} \left( \nu \right) \right) \right).
\end{align*}
The subcovers $\mathcal{S} \left( 0 < \mu \leq P \right)$ and $\mathcal{S} \left( R < \mu \leq P + R \right)$ are both partitions of the indices $\mu_*$. As such, $\left[ \partial_{\bar{x}^{\nu}_k} \bar{x}^{\mu}_j \right]_{\mu, \nu = 1}^{P + R}$ is block diagonal with $2 \times 2$ blocks coupling the indices $0 < \mu \leq R$ and $\nu = \mu + P$. Without loss of generality, we investigate the stability of the block
\begin{align*}
    &\begin{bmatrix}
        \partial_{\bar{x}^{1}_k} \bar{x}^{1}_j & \partial_{\bar{x}^{P + 1}_k} \bar{x}^{1}_j \\
        \partial_{\bar{x}^{1}_k} \bar{x}^{P + 1}_j & \partial_{\bar{x}^{P + 1}_k} \bar{x}^{P + 1}_j
    \end{bmatrix} \\
    &= \frac{1}{2} \frac{\beta_{\text{eff}}}{\sqrt{\sum_i \left[ x^{\prime 1}_i \right]^2}} \sum_{\mu_* \in \mathcal{S} (1)}
    \begin{bmatrix}
        x^{* \mu_*}_j \left( x^{* \mu_*}_k - \left[ \sum_{i} x^{* \mu_*}_i \tilde{x}^{1}_i \right] \tilde{x}^{1}_k \right) & -x^{* \mu_*}_j \left( x^{* \mu_*}_k - \left[ \sum_{i} x^{* \mu_*}_i \tilde{x}^{1}_i \right] \tilde{x}^{1}_k \right) \\
        -x^{* \mu_*}_j \left( x^{* \mu_*}_k - \left[ \sum_{i} x^{* \mu_*}_i \tilde{x}^{1}_i \right] \tilde{x}^{1}_k \right) & x^{* \mu_*}_j \left( x^{* \mu_*}_k - \left[ \sum_{i} x^{* \mu_*}_i \tilde{x}^{1}_i \right] \tilde{x}^{1}_k \right)
    \end{bmatrix}.
\end{align*}
where $x^{\prime 1}_i = \sum_{\mu_* \in \mathcal{S} \left( 1 \right)} x^{* \mu_*}_i$. Let $\mathbf{u}^{1}$ be an arbitrary vector orthogonal to $\mathbf{\Tilde{x}}^{1}$ and define
\begin{align*}
    v^{\mu}_j
    &= \frac{1}{\sqrt{2}} \begin{cases}
        u^{1}_j &\quad \mu = 1 \\
        -u^{1}_j &\quad \mu = P + 1,
    \end{cases}
\end{align*}
then we have the quadratic form
\begin{align*}
    \sum_{\nu \in \left\{ 1, P + 1 \right\}} \sum_{k = 1}^N v^{\nu}_k \ \partial_{\bar{x}^{\nu}_k} \bar{x}^{\mu}_j &=
    \frac{\beta_{\text{eff}}}{\sqrt{\sum_i \left[ x^{\prime 1}_i \right]^2}} \sum_{\mu_* \in \mathcal{S} (1)} \frac{1}{\sqrt{2}}
    \begin{bmatrix}
        x^{* \mu_*}_j \sum_k x^{* \mu_*}_k u^{1}_k \\
        -x^{* \mu_*}_j \sum_k x^{* \mu_*}_k u^{1}_k
    \end{bmatrix} \\
    \sum_{\mu, \nu \in \left\{ 1, P + 1 \right\}} \sum_{j, k = 1}^N v^{\mu}_j v^{\nu}_k \ \partial_{\bar{x}^{\nu}_k} \bar{x}^{\mu}_j &= \frac{\beta_{\text{eff}}}{\sqrt{\sum_i \left[ x^{\prime 1}_i \right]^2}} \sum_{\mu_* \in \mathcal{S} (1)} \left[ \sum_j x^{* \mu_*}_j u^{1}_j \right]^2.
\end{align*}
If $\mathbf{u}^1$ is orthogonal to all the $\mathbf{x}^{* \mu_*}$ such that $\mu_* \in \mathcal{S} \left( 1 \right)$, and in particular if $\mathcal{S} \left( 1 \right)$ contains a single pattern $\mathbf{x}^{* \mu_*} = \mathbf{\Tilde{x}}^{1}$, then $\sum_{\mu_* \in \mathcal{S} (1)} \left[ \sum_j x^{* \mu_*}_j u^{1}_j \right]^2 = 0$, so the quadratic form vanishes. In this case, $\mathbf{u}^{1}$ is a stable direction of the saddle-point equations. Otherwise, the quadratic form does not vanish, so there is a $\beta_{\text{split}}$ such that the direction $\mathbf{u}^1$ is unstable when $\beta_{\text{eff}} > \beta_{\text{split}}$.

\section{Weights learned with unsupervised training}
\label{app:unsupervised_weights}
This Appendix contains plots of DAM weights learned in an unsupervised way (see Section \ref{sec:interpretability}) and sorted in increasing $y = \argmax_{y^\prime} \left\{ p^\mu_{y^\prime} \right\}$. They are not in the main text because they take a lot of space.
\begin{figure}
    \centering
    \includegraphics[width=0.495\linewidth]{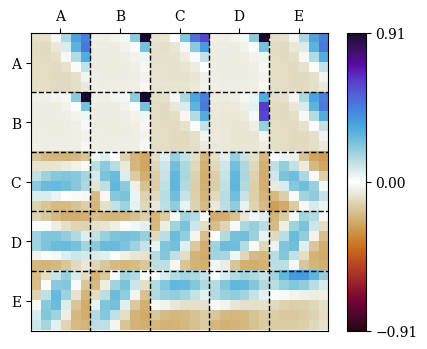}
    \includegraphics[width=\linewidth]{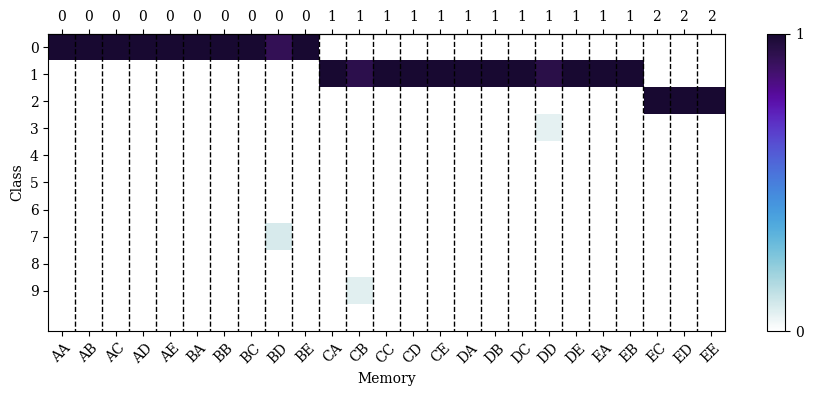}
    \caption{In the top panel, $25$ of the $P = 100$ memories $\mathbf{w}^\mu$ learned by an instance of our dense associative memory (DAM) model trained in an unsupervised way (Eq. \ref{eq:unsupervised_loss}) on $6 \times 6$ patches of the MNIST dataset of handwritten digits \cite{lecun1998gradient} while assuming $C = 10$ latent classes and $\varsigma = 0.6$. In the bottom panel, the corresponding rescaled class weights $\mathbf{p}^\mu / p_{\mathbf{h}} \left( \mu \right)$, where $p_{\mathbf{h}} \left( \gamma \right) = \frac{1}{P + 1}$ for all $0 \leq \gamma \leq P$. The hidden units are indexed using pairs of letters from A to E, and the column-wise maxima of the class weights are the classes of the memories with the corresponding letter indices. $\mathbf{w}^\mu$ and $\mathbf{p}^\mu$ are sorted in increasing $y = \argmax_{y^\prime} \left\{ p^\mu_{y^\prime} \right\}$, and this figure shows $1 \leq \mu \leq 25$.}
    \label{fig:unsupervised_DAM_memories_1}
\end{figure}

\begin{figure}
    \centering
    \includegraphics[width=0.495\linewidth]{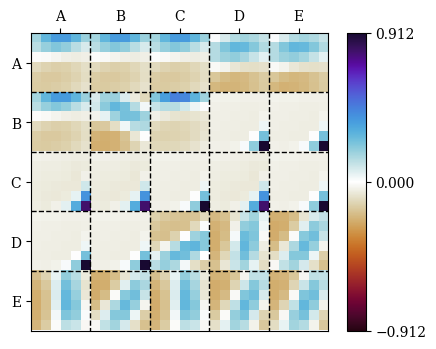}
    \includegraphics[width=\linewidth]{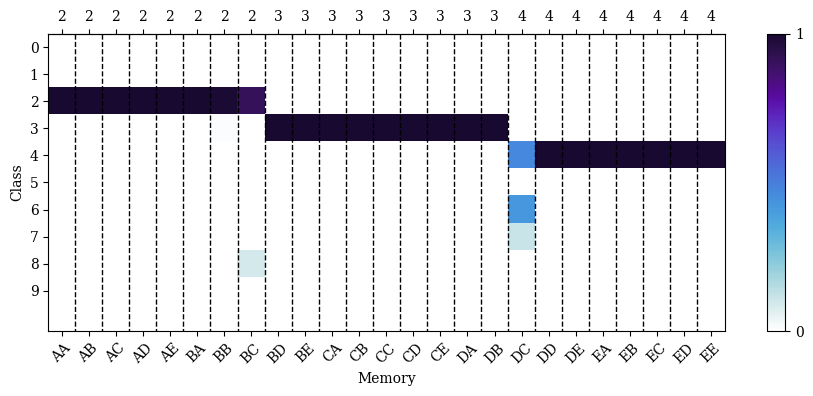}
    \caption{In the top panel, $25$ of the $P = 100$ memories $\mathbf{w}^\mu$ learned by an instance of our dense associative memory (DAM) model trained in an unsupervised way (Eq. \ref{eq:unsupervised_loss}) on $6 \times 6$ patches of the MNIST dataset of handwritten digits \cite{lecun1998gradient} while assuming $C = 10$ latent classes and $\varsigma = 0.6$. In the bottom panel, the corresponding rescaled class weights $\mathbf{p}^\mu / p_{\mathbf{h}} \left( \mu \right)$, where $p_{\mathbf{h}} \left( \gamma \right) = \frac{1}{P + 1}$ for all $0 \leq \gamma \leq P$. The hidden units are indexed using pairs of letters from A to E, and the column-wise maxima of the class weights are the classes of the memories with the corresponding letter indices. $\mathbf{w}^\mu$ and $\mathbf{p}^\mu$ are sorted in increasing $y = \argmax_{y^\prime} \left\{ p^\mu_{y^\prime} \right\}$, and this figure shows $26 \leq \mu \leq 50$.}
    \label{fig:unsupervised_DAM_memories_2}
\end{figure}

\begin{figure}
    \centering
    \includegraphics[width=0.495\linewidth]{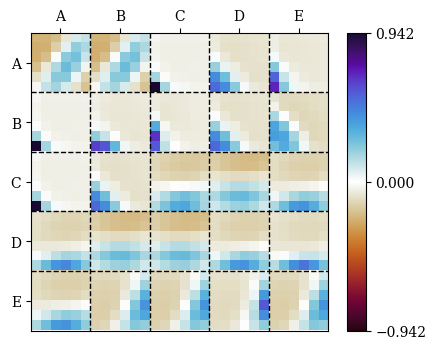}
    \includegraphics[width=\linewidth]{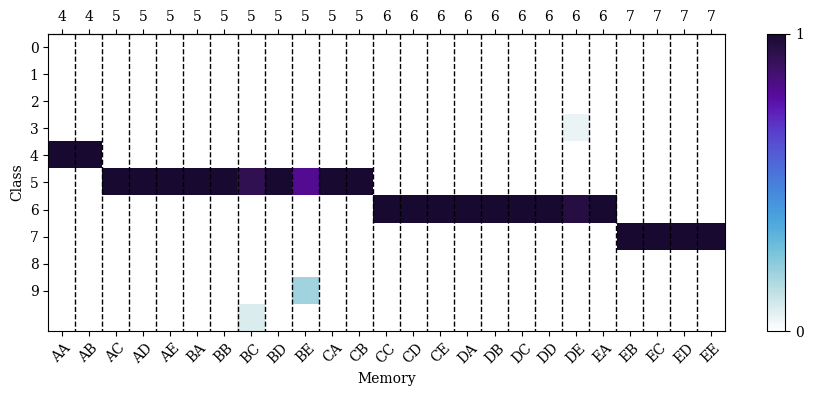}
    \caption{In the top panel, $25$ of the $P = 100$ memories $\mathbf{w}^\mu$ learned by an instance of our dense associative memory (DAM) model trained in an unsupervised way (Eq. \ref{eq:unsupervised_loss}) on $6 \times 6$ patches of the MNIST dataset of handwritten digits \cite{lecun1998gradient} while assuming $C = 10$ latent classes and $\varsigma = 0.6$. In the bottom panel, the corresponding rescaled class weights $\mathbf{p}^\mu / p_{\mathbf{h}} \left( \mu \right)$, where $p_{\mathbf{h}} \left( \gamma \right) = \frac{1}{P + 1}$ for all $0 \leq \gamma \leq P$. The hidden units are indexed using pairs of letters from A to E, and the column-wise maxima of the class weights are the classes of the memories with the corresponding letter indices. $\mathbf{w}^\mu$ and $\mathbf{p}^\mu$ are sorted in increasing $y = \argmax_{y^\prime} \left\{ p^\mu_{y^\prime} \right\}$, and this figure shows $51 \leq \mu \leq 75$.}
    \label{fig:unsupervised_DAM_memories_3}
\end{figure}

\begin{figure}
    \centering
    \includegraphics[width=0.495\linewidth]{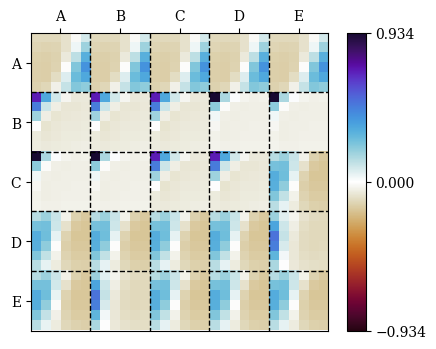}
    \includegraphics[width=\linewidth]{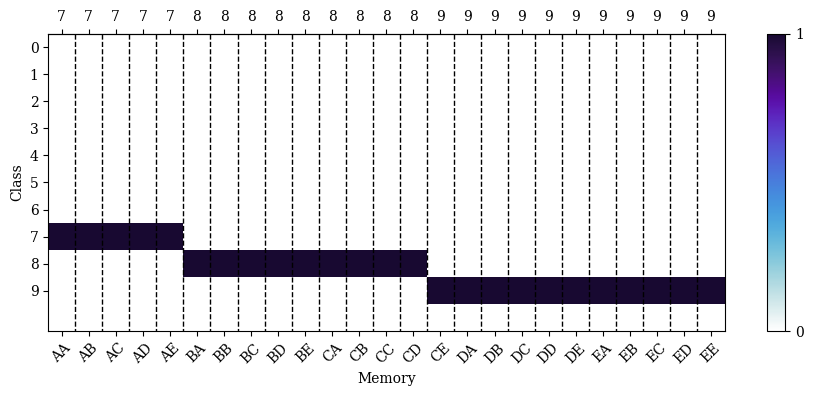}
    \caption{In the top panel, $25$ of the $P = 100$ memories $\mathbf{w}^\mu$ learned by an instance of our dense associative memory (DAM) model trained in an unsupervised way (Eq. \ref{eq:unsupervised_loss}) on $6 \times 6$ patches of the MNIST dataset of handwritten digits \cite{lecun1998gradient} while assuming $C = 10$ latent classes and $\varsigma = 0.6$. In the bottom panel, the corresponding rescaled class weights $\mathbf{p}^\mu / p_{\mathbf{h}} \left( \mu \right)$, where $p_{\mathbf{h}} \left( \gamma \right) = \frac{1}{P + 1}$ for all $0 \leq \gamma \leq P$. The hidden units are indexed using pairs of letters from A to E, and the column-wise maxima of the class weights are the classes of the memories with the corresponding letter indices. $\mathbf{w}^\mu$ and $\mathbf{p}^\mu$ are sorted in increasing $y = \argmax_{y^\prime} \left\{ p^\mu_{y^\prime} \right\}$, and this figure shows $76 \leq \mu \leq 100$.}
    \label{fig:unsupervised_DAM_memories_4}
\end{figure}
\pagebreak

\section{Splitting steepest descent}
\label{app:splitting}
Steps 5 and 11 of splitting steepest descent (Alg. \ref{alg:splitting_descent}) involve the splitting matrices $\mathcal{S}_\mu \left( \mathbf{w}, \mathbf{p} \right)$ derived in \cite{wu2019splitting}. In this appendix, we explain their role in the algorithm. Consult \cite{wu2019splitting} for a detailed explanation of their interpretation and theoretical underpinnings.

Define the thresholds $\tau_{\text{thres}} \in \left( 0, 1 \right]$ and $\lambda_{\text{thres}} \leq 0$. At step 5 of Alg. (\ref{alg:splitting_descent}), we duplicate the hidden units corresponding to the $R \leq \min \left\{ \tau_{\text{thres}} P_{\text{cur}}, P_{\text{max}} - P_{\text{cur}} \right\}$ most negative minimum eigenvalues $\lambda_{\text{min}}^\mu$ of the splitting matrices $\mathcal{S}_\mu \left( \mathbf{w}, \mathbf{p} \right)$ such that $\lambda_{\text{min}}^\mu \leq \lambda_{\text{thres}}$ \cite{wu2019splitting}. To make Fig. (\ref{fig:performance}), we pick $\tau_{\text{thres}} = 1$ and $\lambda_{\text{thres}} \approx 0$. Our splitting matrices are
\begin{align*}
    \mathcal{S}_\mu \left( \mathbf{w}, \mathbf{p} \right) = -\mathbb{E}_{\mathbf{x}^*, y^*} \left[ \frac{p^\mu_y \Bar{\nabla}_{\mathbf{w}^\mu} \Bar{\nabla}_{\mathbf{w}^\mu}^T \exp \left( \beta_{\text{eff}} \sum_{i = 1}^N w^\mu_i x_i \right)}{\sum_{\nu = 1}^P p^{\nu}_y \exp \left( \beta_{\text{eff}} \sum_{i = 1}^N w^\nu_i x_i \right) + p^0_y \frac{\Omega_N \left( \beta \right)}{\Omega_N \left( 0 \right)}} \right],
\end{align*}
where $\Bar{\nabla}_{\bm{\theta}}$ is the gradient constrained to the unit hypersphere $S^{N - 1}$ (see Appendix \ref{app:weight_normalization}) and $\beta_{\text{eff}}$ can be written explicitly in terms of $\beta$ as $\beta_{\text{eff}} = \varsigma \beta$ (see Eqs. \ref{eq:effective_loss} and \ref{eq:saddle-point}). At step 11 of Alg. (\ref{alg:splitting_descent}), we break permutation symmetries in the memories $\mathbf{w}^\mu$ by descending along the eigenvectors $\mathbf{u}^\mu \in S^{N - 1}$ corresponding to the eigenvalues $\lambda_{\text{min}}^\mu$. To be more precise, we update the memories $\mathbf{w}^\mu$ and their duplicates $\mathbf{w}^{\text{dupli}, \mu}$ according to $\mathbf{w}^\mu \gets \mathbf{w}^\mu + \delta \mathbf{u}^\mu$ and $\mathbf{w}^{\text{dupli}, \mu} \gets \mathbf{w}^{\text{dupli}, \mu} - \delta \mathbf{u}^\mu$, respectively, where $\delta$ is a relatively small learning rate \cite{wu2019splitting}. In physics terminology, the eigenvectors $\mathbf{u}^\mu$ are excitation modes that break the permutation symmetries of the memories. $N$ and $P$ are generally large, so it is prohibitively expensive to store $\mathcal{S}_\mu \left( \mathbf{w}, \mathbf{p} \right)$ explicitly for all hidden units $1 \leq \mu \leq P$. Therefore, as proposed in \cite{wang2019energy}, we find the eigenvectors $\mathbf{u}^\mu$ and their eigenvalues $\lambda_{\text{min}}^\mu$ by minimizing the Rayleigh quotients $Q_\mu \left[ \mathbf{w}, \mathbf{p} \right] : S^{N - 1} \ni \mathbf{u}^\mu \mapsto \left[ \mathbf{u}^\mu \right]^T \mathcal{S}_\mu \left( \mathbf{w}, \mathbf{p} \right) \mathbf{u}^\mu$, which can be evaluated without constructing $\mathcal{S}_\mu \left( \mathbf{w}, \mathbf{p} \right)$ explicitly. To derive $Q_\mu \left[ \mathbf{w}, \mathbf{p} \right]$, we first compute $\Bar{\nabla}_{\mathbf{w}^\mu} \Bar{\nabla}_{\mathbf{w}^\mu}^T \exp \left( \beta_{\text{eff}} \sum_i w^\mu_i x_i \right)$. Given $f \left( \bm{\theta} \right) = \sum_i \theta_i x_i$, we find
\begin{align*}
    \Bar{\nabla}_{\bm{\theta}} \Bar{\nabla}_{\bm{\theta}}^T \exp \left( \beta_{\text{eff}} f \left( \bm{\theta} \right) \right) &= \exp \left( \beta_{\text{eff}} f \left( \bm{\theta} \right) \right) \left( \beta_{\text{eff}} \Bar{\nabla}_{\bm{\theta}} \Bar{\nabla}_{\bm{\theta}}^T f \left( \bm{\theta} \right) + \beta_{\text{eff}}^2 \Bar{\nabla}_{\bm{\theta}} f \left( \bm{\theta} \right) \Bar{\nabla}_{\bm{\theta}}^T f \left( \bm{\theta} \right) \right).
\end{align*}
As mentioned at the beginning of Appendix \ref{app:weight_normalization}, $\Bar{\nabla}_{\bm{\theta}} f \left( \bm{\theta} \right)$ is the unrestricted gradient of $f \left( \bm{\theta} \right)$ projected onto the tangent space of $\bm{\theta}$. It remains to calculate $\Bar{\nabla}_{\bm{\theta}} \Bar{\nabla}_{\bm{\theta}}^T f \left( \bm{\theta} \right)$. Following \cite{barilari2023lecture}, we find it to be the iterated projected gradient
\begin{align*}
    \left[ \Bar{\nabla}_{\bm{\theta}} \Bar{\nabla}_{\bm{\theta}}^T f \left( \bm{\theta} \right) \right]_{i \ell} &= \sum_k \left( \delta_{k \ell} - \theta_k \theta_\ell \right) \partial_{\theta_i} \left( \sum_j \left( \delta_{j k} - \theta_j \theta_k \right) \partial_{\theta_j} \left[ \sum_h \theta_h x_h \right] \right) \\ &= \sum_k \left( \delta_{k \ell} - \theta_k \theta_\ell \right) \partial_{\theta_i} \left( x_k - \left[ \sum_j \theta_j x_j \right] \theta_k \right) \\
    &= -\sum_k \left( \delta_{k \ell} - \theta_k \theta_\ell \right) \left( x_i \theta_k + \left[ \sum_j \theta_j x_j \right] \delta_{i k} \right) \\
    &= -\left[ \sum_j \theta_j x_j \right] \left( \delta_{i \ell} - \theta_i \theta_\ell \right),
\end{align*}
so we obtain
\begin{align*}
    Q_\mu \left[ \mathbf{w}, \mathbf{p} \right] \left( \mathbf{u}^\mu \right) &= -\mathbb{E}_{\mathbf{x}^*, y^*} \left[ \frac{p^\mu_y \exp \left( \beta_{\text{eff}} \sum_{i = 1}^N w^\mu_i x_i \right)}{\sum_{\nu = 1}^P p^{\nu}_y \exp \left( \beta_{\text{eff}} \sum_{i = 1}^N w^\nu_i x_i \right) + p^0_y \frac{\Omega_N \left( \beta \right)}{\Omega_N \left( 0 \right)}} F \left( \mathbf{u}^\mu ; \mathbf{w}^\mu, \mathbf{x} \right) \right] \\
    \text{where} \quad F \left( \bm{\varphi} ; \bm{\theta}, \mathbf{x} \right) &= \bm{\varphi}^T \left[ \beta_{\text{eff}} \bar{\nabla}_{\bm{\theta}} \bar{\nabla}_{\bm{\theta}}^T f \left( \bm{\theta} \right) + \beta_{\text{eff}}^2 \bar{\nabla}_{\bm{\theta}} f \left( \bm{\theta} \right) \bar{\nabla}_{\bm{\theta}}^T f \left( \bm{\theta} \right) \right] \bm{\varphi} \\
    &= \beta_{\text{eff}}^2 \left( \sum_{k} \varphi_k x_k - \left[ \sum_{k} \varphi_k \theta_k \right] \left[ \sum_{j} \theta_j x_j \right]\right)^2 \\
    &\quad+ \beta_{\text{eff}} \left[ \sum_j \theta_j x_j \right] \left[ \sum_{i} \varphi_i \theta_i - 1 \right] \left[ \sum_{\ell} \varphi_\ell \theta_\ell + 1 \right].
\end{align*}
We can directly minimize $Q \left[ \mathbf{w}, \mathbf{p} \right] \left( \mathbf{u} \right) = \sum_\mu Q_\mu \left[ \mathbf{w}, \mathbf{p} \right] \left( \mathbf{u}^\mu \right)$ to find the set of all eigenvectors $\mathbf{u}^\mu$ simultaneously. However, it is more convenient to integrate the eigenvector calculations into the DAM architecture. As such, we define the modified loss
\begin{align*}
    \mathcal{L}_\epsilon \left( \mathbf{w}, \mathbf{p}, \mathbf{u} \right) &= -\log \left[ \sum_{\mu = 1}^P p^\mu_y \exp \left( \varsigma \beta \sum_{i = 1}^N w^\mu_i x_i \right) \left[ 1 + \epsilon F \left( \mathbf{u}^\mu ; \mathbf{w}^\mu, \mathbf{x} \right) \right] + p^0_y \frac{\Omega_N \left( \beta \right)}{\Omega_N \left( 0 \right)} \right] \\
    &= -\log \left[ \sum_{\mu = 1}^P p^\mu_y \exp \left( \varsigma \beta \sum_{i = 1}^N w^\mu_i x_i + \log \left[ 1 + \epsilon F \left( \mathbf{u}^\mu ; \mathbf{w}^\mu, \mathbf{x} \right) \right] \right) + p^0_y \frac{\Omega_N \left( \beta \right)}{\Omega_N \left( 0 \right)} \right],
\end{align*}
so that the four gradients used to trained the DAM can be calculated using the equations
\begin{align*}
    \nabla_{\mathbf{u}} \mathcal{Q} \left[ \mathbf{w}, \mathbf{p} \right] \left( \mathbf{u} \right) &= \lim_{\epsilon \rightarrow 0} \left\{ \frac{1}{\epsilon} \nabla_{\mathbf{u}} \mathcal{L}_{\epsilon} \left( \mathbf{w}, \mathbf{p}, \mathbf{u} \right) \right\}, \\
    \nabla_{\mathbf{w}} \mathcal{L} \left( \mathbf{w}, \mathbf{p} \right) &= \nabla_{\mathbf{w}} \mathcal{L}_0 \left( \mathbf{w}, \mathbf{p}, \mathbf{u} \right) \\
    \quad \nabla_{\mathbf{p}} \mathcal{L} \left( \mathbf{w}, \mathbf{p} \right) &= \nabla_{\mathbf{p}} \mathcal{L}_0 \left( \mathbf{w}, \mathbf{p}, \mathbf{u} \right) \\
    \text{and} \quad \nabla_{\beta} \mathcal{L} \left( \mathbf{w}, \mathbf{p} \right) &= \nabla_{\beta} \mathcal{L}_0 \left( \mathbf{w}, \mathbf{p}, \mathbf{u} \right).
\end{align*}
This technique is based on the automatic differentiation trick proposed in \cite{wang2019energy}. We numerically implement the limit in the first equation by setting $\epsilon = 0$ during loss evaluation and $\epsilon = 1$ during gradient computation. When we optimize $\mathcal{Q} \left[ \mathbf{w}, \mathbf{p} \right] \left( \mathbf{u} \right)$, the minima of the different Rayleigh quotients $Q_\mu \left[ \mathbf{w}, \mathbf{p} \right] \left( \mathbf{u}^\mu \right)$ can span multiple orders of magnitude, so we normalize the gradients $\nabla_{\mathbf{u}^\mu} Q_\mu \left[ \mathbf{w}, \mathbf{p} \right] \left( \mathbf{u}^\mu \right)$ by a running average of their magnitudes to facilitate convergence. This is inspired by the use of RMSProp in \cite{wang2019energy}. Moreover, we constrain the eigenvectors $\mathbf{u}^\mu$ and the gradients $\nabla_{\mathbf{u}^\mu} Q_\mu \left[ \mathbf{w}, \mathbf{p} \right] \left( \mathbf{u}^\mu \right)$ to the unit hypersphere as we do for $\mathbf{w}^\mu$ (see Appendix \ref{app:weight_normalization}).
To make Fig. (\ref{fig:performance}), we train the eigenvectors for $1$ epoch, during which we monitor $\min_\mu \left\{ \frac{1}{2} \left[ \mathbf{u}^\mu \right]^T \nabla_{\mathbf{u}^\mu} \mathcal{Q} \left[ \mathbf{w}, \mathbf{p} \right] \left( \mathbf{u}^\mu \right) \right\} = \min_\mu \left\{ \mathcal{Q} \left[ \mathbf{w}, \mathbf{p} \right] \left( \mathbf{u}^\mu \right) \right\} \sim \min_\mu \left\{ \lambda_{\text{min}}^\mu \right\}$ as a metric. The Rayleigh quotients converge very quickly, and further training of the eigenvectors is not necessary.

\bibliographystyle{IEEEtran}
\bibliography{bibliography}

\end{document}